\documentclass[11pt]{article}

\usepackage[final]{acl}

\usepackage{times}
\usepackage{latexsym}

\usepackage[T1]{fontenc}

\usepackage[utf8]{inputenc}

\usepackage{amsmath,amsfonts,bm}

\def\eqref#1{equation~\ref{#1}}

\def\1{\bm{1}}

\DeclareMathAlphabet{\mathsfit}{\encodingdefault}{\sfdefault}{m}{sl}
\SetMathAlphabet{\mathsfit}{bold}{\encodingdefault}{\sfdefault}{bx}{n}

\usepackage{microtype}
\usepackage[table]{xcolor}  
\usepackage{inconsolata}
\usepackage[ruled,vlined]{algorithm2e}
\usepackage{float}
\usepackage{tabularx}
\usepackage{multirow}
\usepackage{graphicx}
\usepackage{pifont}
\usepackage{amsmath,amssymb,amsfonts}
\usepackage{amsthm}
\usepackage{booktabs}

\newtheorem*{theorem*}{Theorem}

\usepackage{pifont}
\newcommand{\cmark}{\ding{51}}
\newcommand{\xmark}{\ding{55}}

\usepackage{arydshln}
\usepackage{makecell}
\usepackage{subcaption}
\usepackage{xcolor}

\definecolor{darkgreen}{RGB}{0,100,0}
\definecolor{darkred}{RGB}{180,0,0}
\definecolor{darkorange}{RGB}{210,105,30}

\newcommand{\tablestyle}[2]{\setlength{\tabcolsep}{#1}\renewcommand{\arraystretch}{#2}\centering\footnotesize}

\title{All for 1-Bit: Towards Genuine 1-Bit Post-Training Quantization for LLMs}

\author{
  \textbf{Zhixiong Zhao}\textsuperscript{1,2}\thanks{~~Equal contribution.}\thanks{~~This work was conducted during his internship at Houmo AI.} ,
  \textbf{Zukang Xu}\textsuperscript{1}\footnotemark[1] ,
  \textbf{Guangyu Sun}\textsuperscript{2} ,
  \textbf{Lifeng Liu}\textsuperscript{2}\footnotemark[3] ,
  \textbf{Dawei Yang}\textsuperscript{1}\thanks{~~Corresponding author.} \\
  \\
  \textsuperscript{1}Houmo AI\\
  \textsuperscript{2}School of Integrated Circuits, Peking University\\
  \small{\textbf{Correspondence:} \href{mailto:dawei.yang@houmo.ai}{dawei.yang@houmo.ai}}
}

\begin{document}
\maketitle

\begin{abstract}
Large language models (LLMs) have achieved remarkable progress, yet their massive storage and memory-bandwidth demands still hinder efficient deployment. Weight binarization is a promising solution, but existing binarization-based post-training quantization (PTQ) methods usually far exceed the nominal 1-bit storage target due to hidden overhead. To address this gap, we propose \textbf{A}ll \textbf{f}or \textbf{1}-Bit (\textbf{AF1}), a genuine 1-bit PTQ framework for LLMs. 
AF1 comprises two complementary components: 
(1) Null-space-Aware Binary Factorization (NABF) for improving binary reconstruction through Hessian-aware surrogate reparameterization, null-space-aware binary factorization, and scale-only global reconstruction; and 
(2) Hierarchical Shapley Allocation (HiSA) for assigning structural capacity using hierarchical Shapley sensitivity. Together, they preserve model accuracy under a strict 1.0 effective-BPW budget for target linear weights in the PTQ setting.
Experiments on LLaMA, Qwen, and Gemma families show that AF1 consistently outperforms existing binarization-based PTQ methods in perplexity and zero-shot accuracy. Compared with BF16, AF1 achieves an average \textbf{2.5$\times$} inference speedup and over \textbf{90\%} memory reduction across evaluated models, providing a practical path toward deployable genuine 1-bit compression for LLMs.
The code for reproducibility is available at \href{https://github.com/Kishon-zzx/AF1}{AF1}.
\end{abstract}
\section{Introduction}
\label{sec:intro}

In recent years, large language models (LLMs) have achieved remarkable progress across multiple domains. For example, LLMs have demonstrated strong reasoning capabilities in structured information understanding and complex question answering tasks~\citep{ye-etal-2026-tableqa,guo-etal-2026-rethinking-table,ye2026rubricguidedprocessrewardstepwise,ye2026rethinkingstepwisemodelrouting,guo-etal-2025-sqlforge}. Beyond conventional language generation, recent studies have further extended LLMs toward autonomous agents, reinforcement learning-based reasoning, and long-horizon decision-making systems~\citep{hao2026rethinking,hao2026recreate,hao2026evolve,wang2026scheduling,liu2026meta}. 
\begin{figure}
    \centering
    \includegraphics[width=0.9\linewidth]{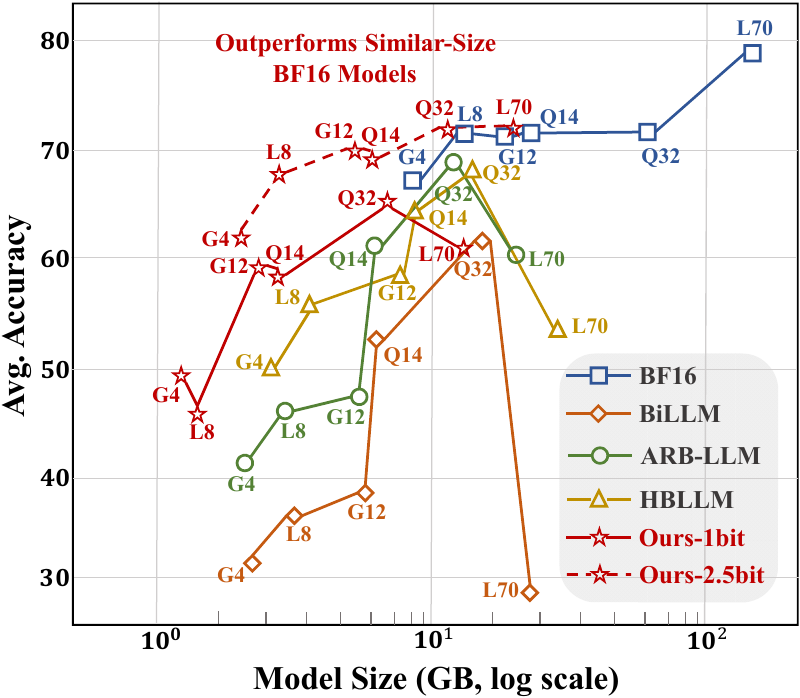}
    \caption{Accuracy--memory trade-off across BF16 models and binarization-based PTQ methods, with L, G, and Q denoting LLaMA-3, Gemma-3, and Qwen3.}
    \label{fig1}
\end{figure}
Meanwhile, foundation models have also been increasingly adapted to specialized vision-language applications and domain-specific scenarios~\citep{zhu2025pathology,zhu2026medeyes,lin2026medcausalx,zhu2026medsynapsevbridgingvisualperception,zhu2024advancing,zhu2025fmri2ges,zhu2026anatomy}. However, their performance gains are often closely coupled with the continuous expansion of the model scale. For example, representative open-source models such as LLaMA-3.1-405B~\citep{llama3herdmodels} contain 405B parameters, while the recently released Qwen3.5-397B-A17B~\citep{qwen3.5} reaches 397B total parameters with 17B activated parameters. This rapid scaling leads to substantial memory overhead during inference. For this class of models approaching the 400B-parameter scale, the model weights alone may require memory on the order of 800 GB under BF16/FP16 precision, even before accounting for additional runtime costs such as the KV cache and intermediate activations. Therefore, reducing the storage and inference costs of LLMs without compromising performance has emerged as a critical challenge for practical deployment, motivating extensive studies on efficient model compression and low-bit quantization.

Post-training quantization (PTQ) has emerged as a key paradigm for reducing the storage and inference costs of LLMs~\citep{zhao2025quark,zhao2026specquant,xu2026kbvq}, with recent studies exploring increasingly aggressive low-bit representations beyond conventional quantization~\citep{zhao2026bwla,zhao2026twla}, among which weight binarization is particularly attractive due to its potential for single-bit storage and bit-level computation. Recent binarization-based PTQ methods, including BiLLM~\citep{huang2024billmpushinglimitposttraining}, ARB-LLM~\citep{li2024arbllmalternatingrefinedbinarizations}, and HBLLM~\citep{chen2025hbllm}, mitigate the inherent performance degradation of binarization by leveraging local Hessian information to identify salient weights and allocate additional storage to them. However, as shown in Fig.~\ref{fig1}, such designs often come with non-negligible auxiliary metadata, including scaling factors, group-wise parameters, and outlier indices. As a result, the end-to-end bits per weight (BPW) can substantially exceed the nominal 1-bit target, often entering an effective 2--4-bit regime in practice. On the other hand, quantization-aware training (QAT) methods such as OneBit~\citep{xu2024onebit} can produce models with genuine 1-bit representations, but they typically rely on large-scale training data, long optimization schedules, and substantial computational resources. Therefore, achieving truly 1-bit PTQ without heavy auxiliary metadata or costly retraining remains a critical challenge for end-to-end efficient LLM inference.

To further understand the difficulty of achieving genuine 1-bit quantization under the PTQ setting, we identify two key limitations of existing methods. First, most binarization-based PTQ methods follow a single-matrix binarization paradigm, where a full-precision weight matrix is directly approximated by a scaled binary matrix, i.e., $\mathbf{W} \approx \alpha \mathbf{B}$, with $\mathbf{B} \in \{-1, +1\}^{m \times n}$. However, the limited representational capacity of a single binary matrix makes it difficult to capture the complex magnitude distributions and structural correlations of LLM weights. Unlike QAT, which can refine binary parameters through large-scale data-driven optimization, PTQ must perform discrete reconstruction with limited calibration data and optimization steps, resulting in much stricter error-control requirements. Recent studies on dataset distillation and data-efficient learning have explored reducing data requirements while preserving model performance, providing complementary perspectives for efficient model development~\citep{li2026mindmarginboundarydistilled,li2026fixedanchorsenoughdynamic,li2026randomautomaticinnerloopoptimization,Li_2024,li2025adaptive}. Second, different structural units in LLMs exhibit highly heterogeneous sensitivity to binarization errors. Uniform quantization configurations or fixed structural budgets often fail to match the non-uniform importance of layers and submodules, leading to suboptimal capacity allocation and full-model performance degradation.

To this end, we propose \textbf{A}ll \textbf{f}or \textbf{1}-Bit (\textbf{AF1}), a genuine 1-bit post-training quantization framework designed to overcome the representational bottleneck of single-matrix binarization and the heterogeneous sensitivity of LLM components. \textbf{AF1} operates through two seamlessly coupled mechanisms. First, Null-space-Aware Binary Factorization (NABF) expands 1-bit capacity via a dual-factor binary structure. It constructs a Hessian-aware surrogate space and strategically redirects discrete projection residuals into the approximate null space of the fixed factor, absorbing structural mismatch into lightweight scaling vectors with zero inference overhead. It then performs a low-cost, scale-only global reconstruction to recalibrate continuous scales while keeping binary matrices frozen. Second, Hierarchical Shapley Allocation (HiSA) optimizes the strictly limited structural budget from a global interaction perspective. It first assigns coarse layer-level budgets via progressive Shapley sampling on end-to-end NLL degradation, and then distributes these budgets to intra-layer submodules using dual-sensitivity Shapley estimates based on the directional and magnitude distortions of block outputs. Extensive experiments demonstrate that \textbf{AF1} achieves a superior accuracy--memory trade-off over existing binarization-based PTQ methods while maintaining a strict 1.0 effective-BPW storage budget for target linear weights.
Our key contributions can be summarized as follows:
\begin{itemize}
    \item We identify two key obstacles to genuine 1-bit PTQ for LLMs: the limited representational capacity of single-matrix binarization and the heterogeneous sensitivity of layers and submodules under extreme compression.
    \item We propose \textbf{A}ll \textbf{f}or \textbf{1}-Bit (\textbf{AF1}), a genuine 1-bit PTQ framework that combines \textbf{NABF} for expressive and error-suppressed binary factorization with \textbf{HiSA} for hierarchical, interaction-aware structural budget allocation.
    \item Extensive experiments on LLM families demonstrate that AF1 achieves a superior accuracy--memory trade-off over existing binarization-based PTQ methods while maintaining a strict 1.0 effective-BPW storage budget for target linear weights.
\end{itemize}
\section{Related Work}
\label{sec:related}

\paragraph{Binarization for LLMs.}

Weight binarization represents LLM weights with binary values, e.g., $\{-1,+1\}$, to reduce storage, memory bandwidth, and inference costs. Existing methods mainly compensate for severe reconstruction error. PB-LLM~\citep{shang2023pbllmpartiallybinarizedlarge} preserves salient weights with Hessian-aware compensation, while BiLLM~\citep{huang2024billmpushinglimitposttraining} uses weight distribution properties, residual approximation, and group-wise search. ARB-LLM~\citep{li2024arbllmalternatingrefinedbinarizations}, STBLLM~\citep{dong2024stbllm}, and HBLLM~\citep{chen2025hbllm} further improve fidelity through refined binarization, structured sparsity, frequency-aware grouping, or saliency-driven selection. Recent studies have also explored ternary representations to further extend the boundary of low-bit LLM compression~\citep{yan2026pt,zhao2026twla}. Despite their effectiveness, these methods require auxiliary metadata, such as salient weights, fine-grained scales, bitmaps, masks, or grouping statistics, making their effective BPW far above the nominal 1-bit target. Achieving genuine 1.0 effective-BPW linear-weight binarization without excessive reconstruction overhead remains challenging.

\paragraph{Mixed-Precision Quantization for LLMs.}

Mixed-precision quantization (MPQ) improves the accuracy--efficiency trade-off by assigning different bit-widths to components. APTQ~\citep{guan2024aptq} combines Hessian trace and attention-aware sensitivity for Transformer-block-level allocation. Recent methods like SpQR~\citep{dettmers2023spqr} preserve high-precision outliers; Atom~\citep{zhao2024atom} combines channel reordering, fine-grained grouping, and mixed precision; SliM-LLM~\citep{huang2024slim} performs salience-based group allocation; CMPQ~\citep{chen2024channel} uses activation distributions; and AMQ~\citep{lee2025amq} adopts AutoML search for weight-only MPQ. Recent low-bit quantization studies have further investigated robust activation quantization and ternary representations for large-scale models~\citep{yang2025robuq,yan2026pt}. Related quantization studies beyond LLMs have also explored adaptive search and reconstruction strategies for diffusion transformers and other large-scale models under aggressive compression settings~\citep{yang2025treeq,zhu2025qartsr,zhang2026q}. However, these methods largely rely on local sensitivity, saliency, or activation statistics, and rarely model layer--submodule interactions. Since quantization errors can propagate across layers and alter downstream distributions and error accumulation, interaction-aware hierarchical allocation remains important for LLM binarization.
\section{Preliminaries}

\begin{figure*}[ht!]
    \centering
    \includegraphics[width=\linewidth]{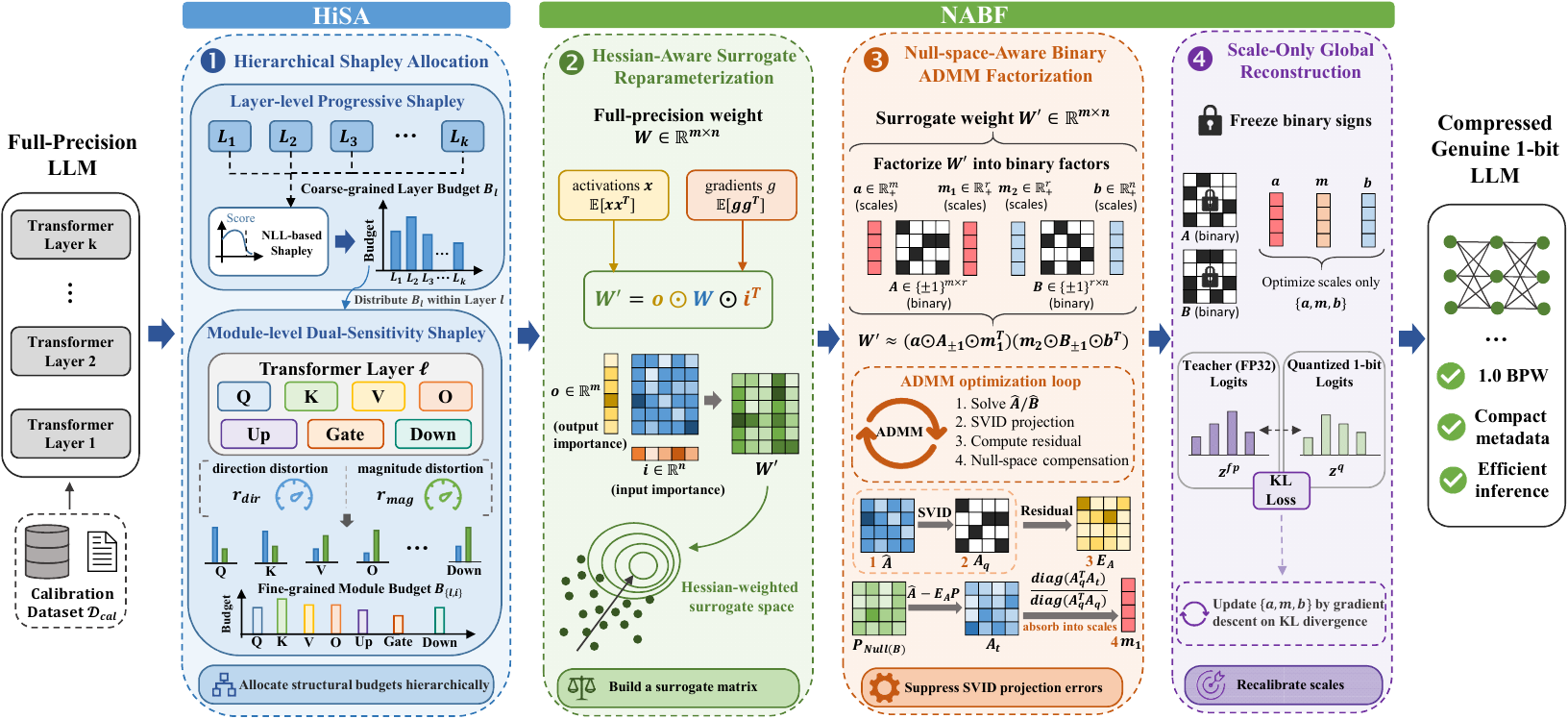}
    \caption{Overview of AF1. 
\textcolor[HTML]{24a645}{Null-space-Aware Binary Factorization (NABF)} constructs Hessian-aware surrogate weights, suppresses SVID projection errors, and performs scale-only reconstruction.
\textcolor[HTML]{1d73b6}{Hierarchical Shapley Allocation (HiSA)} allocates structural budgets from coarse layer-level capacity to fine-grained module-level dimensions.
}
    \label{fig:overview}
\end{figure*}


\paragraph{Null Space.}
Null space describes perturbation directions that are canceled by a linear mapping. Given $\mathbf{B}\in\mathbb{R}^{r\times d}$, a perturbation $\mathbf{E}\in\mathbb{R}^{n\times r}$ satisfying $\mathbf{E}\mathbf{B}=\mathbf{0}$ lies in the left null space induced by $\mathbf{B}$. Similarly, given $\mathbf{A}\in\mathbb{R}^{n\times r}$, a perturbation $\mathbf{E}\in\mathbb{R}^{r\times d}$ satisfying $\mathbf{A}\mathbf{E}=\mathbf{0}$ lies in the right null space induced by $\mathbf{A}$. Such perturbations do not affect the output of the matrix multiplication, making them useful for compensating discrete projection errors in our binary factorization.

\paragraph{Shapley Value.}
The Shapley value~\citep{winter2002shapley} measures the importance of a participant by averaging its marginal contribution over all possible coalitions. Given participants $\mathcal{N}=\{1,\dots,n\}$ and a value function $v(S)$ for coalition $S\subseteq\mathcal{N}$, the Shapley value of participant $i$ is
\begin{equation}
\resizebox{\columnwidth}{!}{$
    \displaystyle
    \phi_i =
    \sum_{S\subseteq \mathcal{N}\setminus\{i\}}
    \frac{|S|!(n-|S|-1)!}{n!}
    \left[
        v(S\cup\{i\}) - v(S)
    \right].
$}
\end{equation}
In LLMs, participants can correspond to layers, blocks, or submodules, while $v(S)$ measures the reconstruction change under a given compression state. We use them to allocate structural capacity.
\section{Method}
\label{sec:method}

\paragraph{Overview.}
The overview of \textbf{AF1} is illustrated in Fig.~\ref{fig:overview}. AF1 integrates \textbf{NABF}, which improves binary reconstruction through Hessian-aware surrogate reparameterization, null-space-aware binary ADMM factorization, and scale-only global reconstruction, with \textbf{HiSA}, which hierarchically allocates structural budgets according to global layer sensitivity and submodule-level distortion. This design enables accurate LLM compression under a strict 1.0 effective-BPW budget for target linear weights. The detailed algorithmic workflow is provided in Appendix~\ref{app:algorithm}.

\subsection{Null-space-Aware Binary Factorization}

\paragraph{Hessian-Aware Surrogate Reparameterization.}
To avoid treating all weight perturbations uniformly under mean-squared reconstruction, we construct a Hessian-aware surrogate weight space. For a linear weight matrix $\mathbf{W}$, a second-order approximation suggests that quantization errors should be weighted by the local Hessian geometry. We follow DBF~\citep{bovza2025addition} and use a K-FAC-inspired diagonal approximation~\citep{martens2015optimizing} based on input activations and output gradients. We define the input importance vector $\mathbf{i}$ and output importance vector $\mathbf{o}$ as
\begin{equation}
    i_j = \sqrt{\mathbb{E}[x_j^2]},
    \qquad
    o_u = \sqrt{\mathbb{E}[g_u^2]},
\end{equation}
where $\mathbf{x}$ denotes the input activation and $\mathbf{g}=\partial \mathcal{L}/\partial \mathbf{Y}$ denotes the output gradient. The original weight matrix is then reparameterized as
\begin{equation}
    \mathbf{W}' = \mathbf{o} \odot \mathbf{W} \odot \mathbf{i}^{\top}.
\end{equation}
This converts the Hessian-weighted reconstruction into a Frobenius objective in the surrogate space, so that subsequent binary factorization emphasizes task-sensitive input and output channels.

\paragraph{Null-space-Aware Binary ADMM Factorization.}
Given the Hessian-aware surrogate matrix $\mathbf{W}'$, we first follow DBF~\citep{bovza2025addition} and decompose it into two binary sign factors with continuous scaling vectors to enhance representation capacity (detailed proof provided in Appendix~\ref{app:double_factor_capacity}):
\begin{equation}
    \mathbf{W}' \approx
    \left(
        \mathbf{a} \odot \mathbf{A}_{\pm1} \odot \mathbf{m}_1^{\top}
    \right)
    \left(
        \mathbf{m}_2 \odot \mathbf{B}_{\pm1} \odot \mathbf{b}^{\top}
    \right),
\label{eq:double_factor_binary}
\end{equation}
where $\mathbf{A}_{\pm1}$ and $\mathbf{B}_{\pm1}$ are binary sign matrices, and $\{\mathbf{a}, \mathbf{m}_1, \mathbf{m}_2, \mathbf{b}\}$ are full-precision scaling vectors. We optimize this factorization using the framework of DBF~\citep{bovza2025addition}: fixing one factor, ADMM solves a continuous proxy, which SVID projects back to the binary-scaled domain (Appendix~\ref{app:dbf_admm}). Taking the left-factor update as an example, let $\hat{\mathbf{A}}$ denote the continuous proxy and $\mathbf{A}_q=\mathrm{SVID}(\hat{\mathbf{A}})$ denote its discrete projection. The residual introduced by discrete projection is
\begin{equation}
    \mathbf{E}_A = \hat{\mathbf{A}} - \mathbf{A}_q .
\end{equation}
This residual propagates through the fixed factor and causes non-negligible output perturbations. The core innovation of NABF is to suppress this propagated projection error through null-space-aware compensation. Taking the left-factor update as an example, the actual perturbation is not $\mathbf{E}_A$ itself but $\mathbf{E}_A\mathbf{B}$. If $\mathbf{E}_A$ lies in the approximate left null space of the fixed factor $\mathbf{B}$, then $\mathbf{E}_A\mathbf{B}\approx\mathbf{0}$, so the projection error can be largely canceled in the forward mapping~\citep{fang2024alphaedit}. Therefore, instead of only minimizing the residual norm, we guide its compensable component toward the null-space directions of the fixed factor.

To construct this subspace efficiently, we avoid directly computing the null space of $\mathbf{B}$. Following existing null-space projection strategies~\citep{wang2021training}, we use the uncentered covariance matrix $\mathbf{C}_B=\mathbf{B}\mathbf{B}^{\top}$, whose null eigenspace is equivalent to the left null space of $\mathbf{B}$ (see Appendix~\ref{app:nullspace_proof}). 
Then, we perform SVD on $\mathbf{C}_B$ as $\mathbf{C}_B=\mathbf{U}\boldsymbol{\Lambda}\mathbf{U}^{\top}$ and collect the trailing singular vectors selected by an adaptive residual-to-total energy cutoff~\citep{Jolliffe2016PrincipalCA} into $\widehat{\mathbf{U}}$, with the threshold set to $\eta=0.01$ (see Appendix~\ref{app:adaptive_nullspace_cutoff}). The approximate null-space projector is defined as $\mathbf{P}_B=\widehat{\mathbf{U}}\widehat{\mathbf{U}}^{\top}$, satisfying $\mathbf{P}_B\mathbf{B}\approx\mathbf{0}$ (see Appendix~\ref{app:projector_proof}).

This property implies that residual components projected onto this subspace have negligible effect after multiplication by the fixed factor $\mathbf{B}$. We therefore define the null-space-compensated target as
\begin{equation}
    \mathbf{A}_{\mathrm{target}}
    =
    \hat{\mathbf{A}}
    -
    \mathbf{E}_A\mathbf{P}_B .
\end{equation}
Multiplying both sides by $\mathbf{B}$ gives
\begin{equation}
\begin{aligned}
    \mathbf{A}_{\mathrm{target}}\mathbf{B}
    &=
    \left(
        \hat{\mathbf{A}}
        -
        \mathbf{E}_A\mathbf{P}_B
    \right)\mathbf{B}  \\
    &=
    \hat{\mathbf{A}}\mathbf{B}
    -
    \mathbf{E}_A\mathbf{P}_B\mathbf{B}
    \approx
    \hat{\mathbf{A}}\mathbf{B}.
\end{aligned}
\end{equation}
Thus, $\mathbf{A}_{\mathrm{target}}$ preserves the propagated behavior of the continuous proxy while providing a better target for recalibrating the discrete factor. To incorporate this compensated target back into the double-factor binary parameterization in Eq.~\ref{eq:double_factor_binary}, we absorb the compensation through the existing scaling degrees of freedom. Specifically, for the left-factor update, we fit a column-wise correction vector $\boldsymbol{\gamma}$ along the intermediate dimension:
\begin{equation}
    \boldsymbol{\gamma}^{*}
    =
    \arg\min_{\boldsymbol{\gamma}}
    \left\|
        \mathbf{A}_{\mathrm{target}}
        -
        \mathbf{A}_q\operatorname{Diag}(\boldsymbol{\gamma})
    \right\|_F^2 .
    \label{eq:gamma_lstsq}
\end{equation}
This convex least-squares problem has the closed-form solution (the proof given in Appendix~\ref{app:gamma_proof}):
\begin{equation}
    \boldsymbol{\gamma}^{*}
    =
    \frac{
        \operatorname{diag}
        \left(
            \mathbf{A}_q^{\top}\mathbf{A}_{\mathrm{target}}
        \right)
    }{
        \operatorname{diag}
        \left(
            \mathbf{A}_q^{\top}\mathbf{A}_q
        \right)
    } .
    \label{eq:gamma_solution}
\end{equation}
We fold the correction into the intermediate scaling vector, i.e., $\mathbf{m}_1 \leftarrow \mathbf{m}_1 \odot \boldsymbol{\gamma}^{*}$. This update changes neither the binary sign matrices nor the inference-time computation graph (right-factor update is symmetric), enabling NABF to absorb discrete projection errors through scale recalibration while preserving the 1-bit parameterization.

\paragraph{Scale-Only Global Reconstruction.}
After double-factor binary optimization, we freeze all binary sign matrices and perform the final correction only through the continuous scale parameters. For the $l$-th linear layer, the reconstructed weight is
\begin{equation}
\resizebox{0.86\columnwidth}{!}{$
    \displaystyle
    \hat{\mathbf{W}}^{(l)}
    =
    \left(
        \mathbf{a}^{(l)}
        \odot
        \mathbf{A}_{\pm1}^{(l)}
        \odot
        (\mathbf{m}^{(l)})^{\top}
    \right)
    \left(
        \mathbf{B}_{\pm1}^{(l)}
        \odot
        (\mathbf{b}^{(l)})^{\top}
    \right),
$}
    \label{eq:final_reconstruction}
\end{equation}
where $\mathbf{A}_{\pm1}^{(l)}$ and $\mathbf{B}_{\pm1}^{(l)}$ are fixed binary sign matrices, 
$\mathbf{m}^{(l)}=\mathbf{m}_1^{(l)}\odot\mathbf{m}_2^{(l)}$ denotes the merged intermediate scale from the two factorization scales, 
and $\mathbf{a}^{(l)}$, $\mathbf{m}^{(l)}$, and $\mathbf{b}^{(l)}$ are learnable scales. We collect learnable scales as $\mathcal{S}=\{\mathbf{a}^{(l)},\mathbf{m}^{(l)},\mathbf{b}^{(l)}\}_{l=1}^{L}$. Given a small calibration set $\mathcal{D}_{\mathrm{cal}}$, we use the full-precision model $M$ as a teacher and optimize $\mathcal{S}$ so that the compressed model $\hat{M}(\cdot;\mathcal{S})$ matches the teacher distribution~\citep{kwon2022alphatuning}. Let $\mathbf{z}_T(\mathbf{X})$ and $\mathbf{z}_Q(\mathbf{X};\mathcal{S})$ denote the teacher and compressed-model logits. The scale-only objective is
\begin{equation}
\begin{aligned}
    \min_{\mathcal{S}} \quad
    &\mathbb{E}_{\mathbf{X}\sim\mathcal{D}_{\mathrm{cal}}}
    \Big[
    D_{\mathrm{KL}}
    \big(
    \mathrm{softmax}(\mathbf{z}_T(\mathbf{X}))
    \,\Vert\, \\
    &\qquad\qquad\quad
    \mathrm{softmax}(\mathbf{z}_Q(\mathbf{X};\mathcal{S}))
    \big)
    \Big].
\end{aligned}
\label{eq:scale_only_kl}
\end{equation}
This stage recalibrates only the scale space while preserving the double-factor binary computation graph. Since binary matrices are frozen, the trainable parameters per linear layer are reduced from $O(d_{\mathrm{out}}r + rd_{\mathrm{in}})$ to $O(d_{\mathrm{out}} + r + d_{\mathrm{in}})$. Thus, NABF bridges performance gaps with minimal memory while preserving the 1-bit parameterization.

\begin{table*}[t!]
\centering
\vspace{-0.3cm}
\resizebox{\textwidth}{!}{%
\tablestyle{2pt}{1.2}
\begin{tabular}{l|c|cc:cc|cc:cc:cc|cc:cc}
\toprule
\multirow{3}{*}{\textbf{Method}} 
& \multirow{3}{*}{\makecell{\textbf{\#Bits}\\ (W)}} 
& \multicolumn{2}{c}{\textbf{LLaMA-3-8B}} 
& \multicolumn{2}{c}{\textbf{LLaMA-3-70B}} 
& \multicolumn{2}{c}{\textbf{Qwen3-8B}} 
& \multicolumn{2}{c}{\textbf{Qwen3-14B}} 
& \multicolumn{2}{c}{\textbf{Qwen3-32B}} 
& \multicolumn{2}{c}{\textbf{Gemma-3-4B-it}} 
& \multicolumn{2}{c}{\textbf{Gemma-3-12B-it}} \\
\cdashline{3-16}
& 
& 0-shot$^7$ & Wiki 
& 0-shot$^7$ & Wiki 
& 0-shot$^7$ & Wiki 
& 0-shot$^7$ & Wiki 
& 0-shot$^7$ & Wiki 
& 0-shot$^7$ & Wiki 
& 0-shot$^7$ & Wiki \\
& 
& Avg.($\uparrow$) & ($\downarrow$) 
& Avg.($\uparrow$) & ($\downarrow$) 
& Avg.($\uparrow$) & ($\downarrow$) 
& Avg.($\uparrow$) & ($\downarrow$) 
& Avg.($\uparrow$) & ($\downarrow$) 
& Avg.($\uparrow$) & ($\downarrow$) 
& Avg.($\uparrow$) & ($\downarrow$) \\
\hdashline
BF16 
& 16 
& 72.51 & 6.14 
& 79.08 & 2.86 
& 69.09 & 9.00 
& 72.47 & 8.64 
& 72.42 & 7.61 
& 67.12 & 17.39 
& 72.20 & 25.20 \\
\hdashline
RTN 
& 1.00 
& 25.82 & 2.57e6
& 24.88 & 3.77e4
& 25.28 & 7.81e6
& 25.55 & 1.37e8
& 25.41 & 3.82e6
& 25.28 & 2.12e7
& 25.30 & 7.29e7 \\

BiLLM 
& 2.88 
& 35.38 & 48.18 
& 28.32 & 137.64 
& 33.70 & 96.21 
& 53.53 & 28.66 
& 61.23 & 16.79 
& 32.33 & 134.42 
& 38.41 & 381.19 \\

STBLLM$_{\text{(4:8)}} $
& 3.50 
& 29.43 & 196.03 
& 26.55 & 963.92 
& 28.02 & 291.76 
& 40.88 & 48.80 
& 55.86 & 17.31 
& 29.31 & 202.54 
& 31.46 & 4.69e5 \\

ARB-LLM$_{\text{RC}} $
& 2.51 
& 46.21 & 28.97 
& 60.76 & 11.10 
& 49.85 & 40.90 
& 61.60 & 18.40 
& \underline{69.56} & 12.93 
& 41.83 & 95.18 
& 47.15 & 4.68e3 \\

HBLLM$_{\text{row}}$ 
& 3.25 
& \underline{55.35} & \underline{14.26} 
& 53.19 & 9.55 
& 54.76 & 16.28 
& \underline{64.12} & \underline{11.01} 
& 67.28 & \underline{10.27} 
& \underline{50.35} & 43.05 
& 57.40 & 758.45 \\
\hdashline
\rowcolor{gray!5}
AF1 
& 2.50 
& \textbf{67.35} & \textbf{8.07} 
& \textbf{72.27} & \textbf{4.52} 
& \textbf{67.95} & \textbf{9.71} 
& \textbf{70.33} & \textbf{8.87} 
& \textbf{72.49} & \textbf{8.38} 
& \textbf{62.29} & \textbf{15.39} 
& \textbf{70.44} & \textbf{15.88} \\
\rowcolor{gray!5}
AF1 
& \textbf{1.00} 
& 45.48 & 22.18 
& \underline{61.80} & \underline{8.91} 
& \underline{54.82} & \underline{16.26} 
& 59.61 & 13.32 
& 64.26 & 11.28 
& 49.86 & \underline{32.82} 
& \underline{59.02} & \underline{27.07} \\
\bottomrule
\end{tabular}%
}
\vspace{-0.3cm}
\caption{Comparison of perplexity on WikiText2 and averaged accuracy on seven zero-shot tasks. Bold and underline denote the \textbf{best} and \underline{second-best} quantized results.
Full results are in Appendix~\ref{app:detailed main results}.}
\vspace{-0.3cm}
\label{tab:main_results}
\end{table*}

\subsection{Hierarchical Shapley Allocation}
Although NABF achieves accurate operator-level reconstruction under a given BPW budget, applying a uniform BPW configuration to the full model ignores the heterogeneous sensitivity of different layers and submodules. However, conventional local sensitivity estimates can be unreliable because LLM compression errors propagate across layers and interact with each other~\citep{zhao2025impq}. We therefore propose Hierarchical Shapley Allocation (HiSA), which evaluates component importance from a cooperative-game perspective. By capturing global interactions and within-layer heterogeneity under fixed storage constraints, HiSA generates adaptive BPW configurations for NABF.

\paragraph{Layer-Level Coarse-Grained Budget Allocation via Progressive Shapley.}
We formulate layer-wise dimension assignment as a structural-capacity game, whose interaction-aware property is discussed in Appendix~\ref{app:shapley_interaction}. Following Shapley-based mixed-precision quantization~\citep{zhao2025impq}, we evaluate the structural importance of each layer by changing its intermediate dimension. Let $T=\{1,2,\dots,L\}$ denote the layers. For any subset $S\subseteq T$, layers in $S$ keep the original intermediate dimension $r_{\mathrm{orig}}$, while the remaining layers are degraded to a proxy dimension $r_{\mathrm{proxy}}$ that satisfies the 1.0 effective-BPW budget for target linear weights. Given a calibration corpus $\mathcal{D}$, we evaluate each mixed-capacity state by the average token-level negative log-likelihood:
\begin{equation}
\resizebox{0.86\columnwidth}{!}{$
    \displaystyle
v_{\mathrm{NLL}}(S)
=
\mathbb{E}_{(x,t)\sim\mathcal{D}}
\left[
-\log p(x_{t+1}\mid x_{\le t};S)
\right],
$}
\label{eq:hisa_layer_value}
\end{equation}
where a larger value indicates worse performance. Since enumerating all states is infeasible, we estimate Shapley scores by Monte Carlo permutations. For a sampled permutation $\pi=(\pi_1,\dots,\pi_L)$, we start from the full-capacity state and progressively remove layers from $S$, i.e., degrade them to $r_{\mathrm{proxy}}$. Let $S_\ell$ be the set of layers still retaining $r_{\mathrm{orig}}$ before degrading $\pi_\ell$. The marginal contribution of $\pi_\ell$ is the induced NLL increase:
\begin{equation}
\Delta v_{\pi_\ell}
=
v_{\mathrm{NLL}}(S_\ell\setminus\{\pi_\ell\})
-
v_{\mathrm{NLL}}(S_\ell).
\label{eq:hisa_layer_marginal}
\end{equation}
Averaging over $M$ sampled permutations gives
\begin{equation}
\Phi_l
=
\frac{1}{M}
\sum_{m=1}^{M}
\Delta v_l^{(m)}.
\label{eq:hisa_layer_shapley}
\end{equation}
We convert the layer-level Shapley scores into coarse structural budgets through a temperature-controlled softmax:
\begin{equation}
B_l
=
B_{\mathrm{total}}
\frac{\exp(\Phi_l/\tau)}
{\sum_{j=1}^{L}\exp(\Phi_j/\tau)}.
\label{eq:hisa_layer_budget}
\end{equation}
Here, $B_{\mathrm{total}}$ is the total structural budget, and $\tau$ controls the allocation concentration. The resulting $B_l$ serves as the layer-level budget for subsequent within-layer module allocation.

\paragraph{Block-Level Fine-Grained Allocation via Dual-Sensitivity Shapley.}
To obtain finer-grained allocation within each layer, HiSA evaluates submodule importance by the output perturbation caused by compressing each module. However, the conventional MSE-based perturbation measure is limited, as it merges directional and magnitude drift into a single scalar despite their different failure modes. HiSA therefore measures two complementary distortions and estimates their Shapley marginal contributions separately. Specifically, $r_{\mathrm{dir}}$ captures angular distortion, while $r_{\mathrm{mag}}$ captures relative magnitude perturbation (with their formal definitions and distinct failure-mode analysis provided in Appendix~\ref{app:dual_distortion}). We then construct a module-level cooperative game inside layer $l$. Let $H_l^{\mathrm{ref}}$ be the original reference output of layer $l$, $\mathcal{N}_l$ be the candidate submodules, and $H_l(S)$ be the layer output when a subset $S\subseteq\mathcal{N}_l$ is degraded. For each distortion type $c\in\{\mathrm{dir},\mathrm{mag}\}$, we define the subset error as
\begin{equation}
E_c(S)=r_c\!\left(H_l^{\mathrm{ref}},H_l(S)\right).
\label{eq:hisa_module_error}
\end{equation}
Given a Monte Carlo permutation $\pi$, we progressively degrade submodules from $S_0=\varnothing$. To reduce negative marginals caused by accidental error cancellation, we use the monotonic accumulated error $\hat E_c(S_t)=\max(E_c(S_t),\hat E_c(S_{t-1}))$ and define the marginal contribution of module $i=\pi_t$ as
\begin{equation}
\Delta_c^{(i)}(\pi)
=
\hat E_c(S_t)-\hat E_c(S_{t-1}).
\label{eq:hisa_module_marginal}
\end{equation}
Following the same Monte Carlo averaging form as Eq.~\ref{eq:hisa_layer_shapley}, we obtain the module-level Shapley estimates $\Phi_{c}^{(i)}$ for $c\in\{\mathrm{dir},\mathrm{mag}\}$. Because the two scores have different scales, we normalize them with
a MAD-based robust Z-score~\citep{iglewicz1993volume}, equivalently written in the scaled-MAD form:
\begin{equation}
\resizebox{0.87\columnwidth}{!}{$
    \displaystyle
z_c^{(i)}
=
\frac{
\Phi_c^{(i)}-\mathrm{Median}(\Phi_c)
}{
1.4826\cdot\mathrm{MAD}(\Phi_c)+\epsilon
},
c\in\{\mathrm{dir},\mathrm{mag}\}.
$}
\label{eq:hisa_mad_norm}
\end{equation}
Then, we map the normalized scores by $P_c^{(i)}=1/(1+\exp(-z_c^{(i)}))$~\citep{zhang2026beyond}. To prevent vulnerability in either failure mode from being suppressed by averaging, we fuse the two modes with a Soft-OR rule
$v_{\mathrm{base}}^{(i)}
=
1-
(1-P_{\mathrm{dir}}^{(i)})
(1-P_{\mathrm{mag}}^{(i)})$.
Since submodules within the same layer can have substantially different parameter counts, we allocate the layer budget using a smoothed size-aware sensitivity (see Appendix~\ref{app:size_density}):
\begin{equation}
\resizebox{0.86\columnwidth}{!}{$
    \displaystyle
b_i
=
B_l
\cdot
\frac{
\left(\sum_{k\in\mathcal{N}_l}|\mathbf{W}_k|\right)
\left(v_{\mathrm{base}}^{(i)}/|\mathbf{W}_i|+\epsilon\right)^{\alpha}
}{
\sum_{j\in\mathcal{N}_l}
|\mathbf{W}_j|
\left(v_{\mathrm{base}}^{(j)}/|\mathbf{W}_j|+\epsilon\right)^{\alpha}
}.
$}
\label{eq:hisa_module_budget}
\end{equation}
This allocation preserves the layer-level budget and the resulting $b_i$ is then converted into the intermediate dimension used by NABF.

\begin{table}[t]
\centering
\resizebox{0.49\textwidth}{!}{%
\tablestyle{1.8pt}{1.2}
\begin{tabular}{clccccc}
\toprule
\textbf{Model} & \textbf{Method} & \textbf{\#Bits(W)} & \textbf{Model Size} & \textbf{MMLU-Pro} & \textbf{HumanEval} & \textbf{GSM8K} \\
\midrule
\multirow{6}{*}{\makecell{LLaMA-3.1\\-8B-Instruct}}
& BF16        & 16   & 16.07GB & 46.64 & 67.07 & 83.24 \\
\cline{2-7}
& BiLLM       & 2.88 & 3.75GB  & 8.77  & 15.33 & 21.36 \\
& STBLLM$_{\text{(4:8)}}$ & 3.50 & 4.32GB  & 13.74 & 10.15 & 10.83 \\
& ARB-LLM$_{\text{RC}} $ & 2.51 & 3.36GB  & 10.16 & 20.12 & 30.38 \\
& HBLLM$_{\text{row}}$    & 3.25 & 4.15GB  & \textbf{23.57} & \underline{28.97} & \underline{40.26} \\
\rowcolor{gray!5}
& AF1        & \textbf{1.00} & \textbf{1.56GB}  & \underline{20.12} & \textbf{36.71} & \textbf{44.78} \\
\midrule
\multirow{6}{*}{\makecell{Qwen3-14B\\-Instruct}}
& BF16        & 16   & 29.51GB & 69.11 & 89.63 & 94.77 \\
\cline{2-7}
& BiLLM       & 2.88 & 6.70GB  & 23.75 & 19.74 & 21.06 \\
& STBLLM$_{\text{(4:8)}}$ & 3.50 & 8.11GB  & 13.38 & 13.70 & 12.05 \\
& ARB-LLM$_{\text{RC}} $ & 2.51 & 5.99GB  & 26.64 & 26.37 & 37.74 \\
& HBLLM$_{\text{row}}$    & 3.25 & 7.43GB  & \textbf{43.84} & \textbf{52.32} & \textbf{72.48} \\
\rowcolor{gray!5}
& AF1        & \textbf{1.00} & \textbf{2.92GB}  & \underline{38.57} & \underline{49.76} & \underline{64.66} \\
\bottomrule
\end{tabular}}
\vspace{-0.3cm}
\caption{Results on instruction-tuned models across three challenging benchmarks. Bold and underline denote the \textbf{best} and \underline{second-best} quantized results.}
\vspace{-0.3cm}
\label{tab:instruct_reasoning}
\end{table}

\begin{figure*}[t!]
    \centering
    \includegraphics[width=0.98\linewidth]{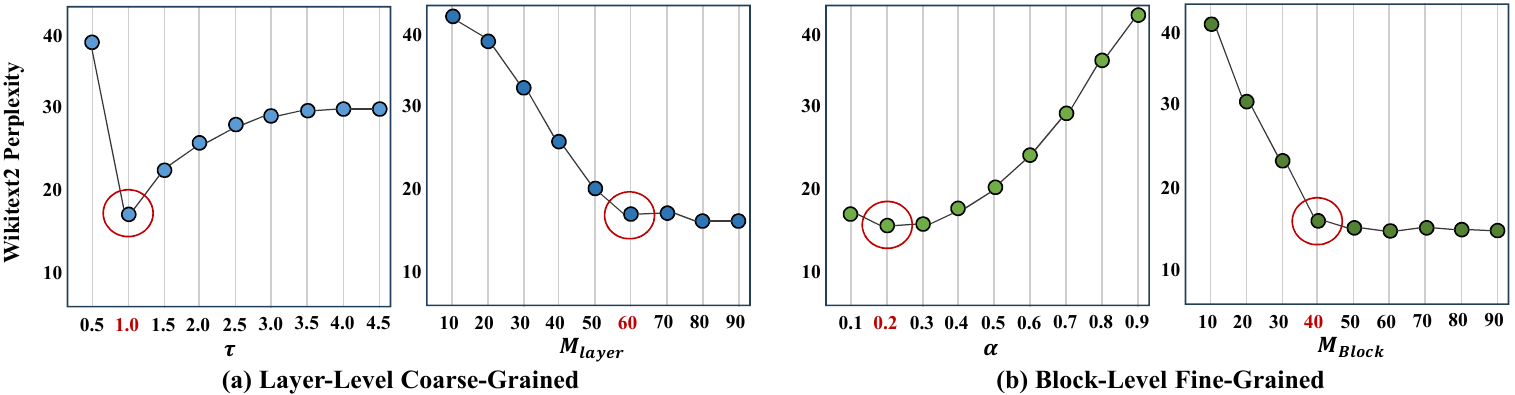}
    \caption{
Hyperparameter sensitivity of HiSA on WikiText2 perplexity.
(a) Layer-level coarse-grained allocation under different $\tau$ and $M_{\mathrm{layer}}$.
(b) Block-level fine-grained allocation under different $\alpha$ and $M_{\mathrm{block}}$.
}
\label{fig:hisa_hyperparam}
\end{figure*}
\section{Experiments}
\label{sec:Experiments}

\subsection{Experiment setup}
\paragraph{Models and Datasets.}
We evaluate \textbf{AF1} on representative LLM families, including LLaMA-2/3~\citep{llama3herdmodels}, Qwen3~\citep{qwen3technicalreport}, and Gemma-3~\citep{gemmateam2025gemma3technicalreport}, along with their instruction-tuned variants. We report WikiText2 perplexity~\citep{wikitext2} and seven zero-shot tasks: ARC-Challenge/Easy~\citep{clark2018thinksolvedquestionanswering}, HellaSwag~\citep{zellers2019hellaswagmachinereallyfinish}, LAMBADA-openai/standard~\citep{paperno2016lambadadatasetwordprediction}, PIQA~\citep{bisk2019piqareasoningphysicalcommonsense}, and WinoGrande~\citep{sakaguchi2019winograndeadversarialwinogradschema}. We also include stronger reasoning and generation benchmarks: MMLU-Pro~\citep{wang2024mmlu}, GSM8K~\citep{gsm8k}, and HumanEval~\citep{humaneval}.

\paragraph{Baseline Methods.} 
We compare \textbf{AF1} with representative binarization-based PTQ methods, including BiLLM~\citep{huang2024billmpushinglimitposttraining}, ARB-LLM~\citep{li2024arbllmalternatingrefinedbinarizations}, STBLLM~\citep{dong2024stbllm}, and HBLLM~\citep{chen2025hbllm} (with their effective BPW derived in Appendix~\ref{app:bpw_binary_ptq}). To provide additional reference under genuine 1-bit compression, we also include recent QAT methods, including OneBit~\citep{xu2024onebit} and LittleBit~\citep{lee2026littlebit}. Since QAT requires extra data and much higher optimization cost, these results serve as reference points rather than strict PTQ comparisons.

\paragraph{Implementation Details.}
All experiments are implemented with PyTorch and conducted on NVIDIA RTX PRO 6000 GPUs. Following the calibration setting of DBF, we randomly sample 128 sequences from the RedPajama~\citep{weber2024redpajamaopendatasettraining} dataset with a sequence length of 2048. For the ADMM-SVID optimization in NABF, we follow the default setting of DBF. For evaluation, we use lm-evaluation-harness~\citep{eval-harness} for the seven zero-shot tasks, and EvalScope~\citep{evalscope_2024} for MMLU-Pro, GSM8K, and HumanEval.

\subsection{Main Results}

\paragraph{Comparison Results.}
Table~\ref{tab:main_results} reports the main results of AF1 across LLaMA-3, Qwen3, and Gemma-3 model families. 
Under the 2.5-bit setting, which is close to the BPW of existing binarization-based PTQ methods, AF1 achieves the best results among quantized methods on all models. Compared with the strongest baseline HBLLM, AF1 improves the average zero-shot accuracy across all models by over 20\% and reduces the average WikiText2 PPL by over 90\%, showing substantially stronger low-bit reconstruction capability under comparable storage cost.
More importantly, AF1 remains highly competitive under a genuine 1-bit storage budget. On the largest model, LLaMA-3-70B, AF1 achieves 61.80 zero-shot accuracy and 8.91 WikiText2 PPL at 1.0 effective BPW for target linear weights, even surpassing the strongest baseline HBLLM. Similar trends on Qwen3 and Gemma-3 further indicate that AF1 delivers stable cross-family 1-bit reconstruction rather than overfitting to a single architecture (Practical generation examples with GPT-5.5~\citep{openai2026gpt55} annotations are provided in Appendix~\ref{sec:dialog}).

\begin{table}[t!]
\centering
\resizebox{0.95\linewidth}{!}{
\tablestyle{2.5pt}{1.2}
\begin{tabular}{c | c:c:c | c:c | cc}
\toprule
\multirow{2}{*}{\textbf{Model}} &
\multicolumn{3}{c|}{\textbf{NABF}} &
\multicolumn{2}{c|}{\textbf{HiSA}} &
\textbf{Wiki2} &
\textbf{0-shot$^{7}$} \\
& \textbf{Step1} & \textbf{Step2} & \textbf{Step3} 
& \textbf{Layer} & \textbf{Module} &
\textbf{PPL}($\downarrow$) & \textbf{Avg.}($\uparrow$) \\
\midrule
\multirow{6}{*}{\makecell{LLaMA-\\3.2-3B}}
& \xmark & \xmark & \xmark
& \multicolumn{2}{c|}{\multirow{4}{*}{\LARGE\xmark}}
& 8.9e2 & 25.48 \\
& \cmark & \xmark & \xmark
& \multicolumn{2}{c|}{}
& 1.3e2 & 29.31 \\
& \cmark & \cmark & \xmark
& \multicolumn{2}{c|}{}
& 41.34 & 34.85 \\
& \cmark & \cmark & \cmark
& \multicolumn{2}{c|}{}
& 30.47 & 40.79 \\
\cdashline{2-8}
& \multicolumn{3}{c|}{\multirow{2}{*}{\LARGE\cmark}}
& \cmark & \xmark
& 25.52 & 42.30 \\
& \multicolumn{3}{c|}{}
& \cmark & \cmark
& \textbf{24.21} & \textbf{43.09} \\
\bottomrule
\end{tabular}}
\vspace{-0.2cm}
\caption{Combined ablation on LLaMA-3.2-3B. Step1--3 denote surrogate reparameterization, null-space binary factorization, and scale-only reconstruction; Layer and Module denote the two stages of HiSA allocation.}
\vspace{-0.4cm}
\label{tab:nabf_ablation}
\end{table}

\paragraph{Experiments on Instruction-tuned Models.}
Instruction-tuned models are practically important but more vulnerable to quantization errors in reasoning, coding, and mathematics. As shown in Table~\ref{tab:instruct_reasoning}, AF1 maintains strong instruction-tuned performance with much smaller model sizes. For example, on Qwen3-14B-Instruct, AF1 consistently outperforms BiLLM, STBLLM, and ARB-LLM across all benchmarks. Compared with HBLLM, AF1 has only about a 9\% lower average score while using just 39.3\% of its model size ($7.43\,\mathrm{GB}\rightarrow2.92\,\mathrm{GB}$). These results show AF1 achieves a favorable accuracy--memory trade-off for instruction-tuned LLMs under genuine 1-bit PTQ.

\subsection{Ablation Studies}

\paragraph{Component Ablation.}
We analyze the contributions of NABF and HiSA using WikiText2 PPL and seven-task average accuracy. As shown in Table~\ref{tab:nabf_ablation}, NABF alone substantially improves the 1-bit reconstruction quality, reducing PPL from $8.9e2$ to 30.47 and achieving 40.79 average accuracy. Building on this stronger reconstruction, HiSA further improves the limited-budget allocation: the final model reduces PPL to 24.21 and raises the average accuracy to 43.09. 
Additional ablations on the residual-energy threshold and HiSA sensitivity metrics are provided in Appendices~\ref{app:ablation_nullspace_threshold} and~\ref{app:ablation_hisa_metric}.
These results show that NABF and HiSA are complementary: NABF provides a high-fidelity binary reconstruction, while HiSA allocates the scarce capacity to sensitive layers and submodules.

\paragraph{Hyperparameter Sensitivity of HiSA.}
We analyze HiSA's sensitivity on Qwen3-8B with respect to four key hyperparameters: the temperature coefficient $\tau$ in Eq.~\ref{eq:hisa_layer_budget}, the smoothing exponent $\alpha$ in Eq.~\ref{eq:hisa_module_budget}, and the layer- and block-level sampling numbers $M_{\mathrm{layer}}$ and $M_{\mathrm{block}}$. As shown in Fig.~\ref{fig:hisa_hyperparam}, $\tau$ controls the concentration of layer-wise budgets: overly small values over-concentrate capacity on a few layers, while overly large values approach uniform allocation. The best result is obtained at $\tau=1.0$. Increasing $M_{\mathrm{layer}}$ and $M_{\mathrm{block}}$ stabilizes Shapley estimation, saturating around 60 and 40 samples, respectively. $\alpha$ balances sensitivity-density redistribution and noise amplification, with the best result at $\alpha=0.2$. Similar trends on other models are reported in Appendix~\ref{app:hisa_more_ablation}, which supports our default configuration of $\tau=1.0$, $M_{\mathrm{layer}}/M_{\mathrm{block}}=60/40$, and $\alpha=0.2$. We also report the calibration ablation in Appendix~\ref{app:ablation_calibration_data}.

\subsection{Efficiency Analysis of AF1}

\paragraph{Quantization Time Comparison.}
\begin{figure}[t!]
    \centering
    \includegraphics[width=0.95\linewidth]{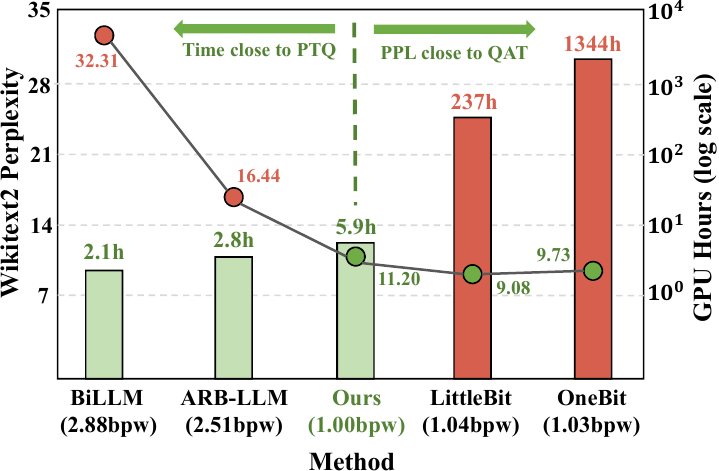}
   \caption{
Quantization time and WikiText2 perplexity comparison under near-1-bit compression.
}
\label{fig:efficiency_time}
\end{figure}

We compare AF1 with representative PTQ and QAT-based binarization methods on LLaMA-2-7B. As shown in Fig.~\ref{fig:efficiency_time}, prior PTQ methods such as BiLLM and ARB-LLM require only a few GPU hours, but their BPWs are close to 3 and PPL are much worse. In contrast, QAT-based methods such as LittleBit and OneBit achieve lower PPL near 1-bit storage, but require hundreds to thousands of GPU hours, limiting scalability. As a PTQ method, AF1 bridges this gap: under a 1.0 effective-BPW target-linear-weight setting, it reduces PPL from 16.44 (ARB-LLM) to 11.20 with only 5.9 GPU hours. Compared with QAT baselines, AF1 incurs only a small PPL gap relative to them while being about 40$\times$ faster than LittleBit and 228$\times$ faster than OneBit. These results show that AF1 provides PTQ-level efficiency and near-QAT performance.

\begin{figure}[t!]
    \centering
    \includegraphics[width=0.9\linewidth]{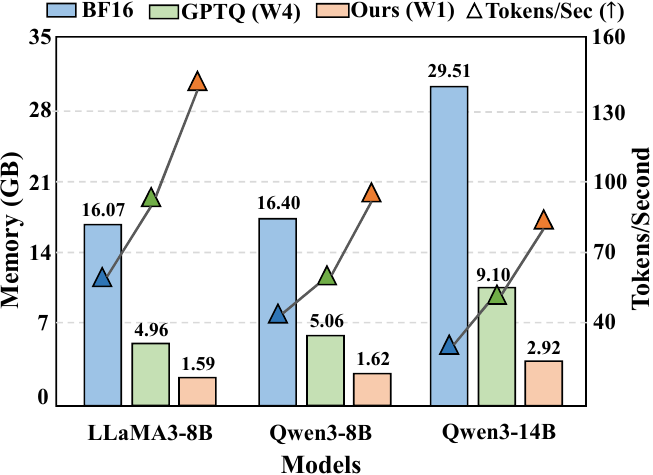}
   \caption{
Inference memory and decoding throughput comparison across BF16, GPTQ-W4, and AF1.
}
\label{fig:memory_speed}
\end{figure}

\paragraph{Speedup and Memory Savings.}
We benchmark the inference efficiency of AF1 on three models, comparing BF16, GPTQ-W4, and AF1 in memory usage and decoding throughput, with binary matrix multiplication implemented using GemLite~\citep{badri2023hqq}. Under a batch size of 1, a 1024-token prefill, and 4096-token decoding, Fig.~\ref{fig:memory_speed} shows that AF1 consistently reduces memory and improves throughput. Since long-context autoregressive decoding is memory-bandwidth-bound, reducing BPW to 1 substantially lowers memory traffic compared with BF16 and 4-bit PTQ, leading to faster decoding. Across three models, AF1 reduces memory by about 90\% over BF16 and 68\% over GPTQ-W4 on average, while improving decoding throughput by about 2.5$\times$ over BF16. These results highlight the practical memory and decoding efficiency gains enabled by AF1 in practice.

\section{Conclusion}
\label{sec:conclusion}

We present AF1, a genuine 1-bit PTQ framework for LLMs. By addressing the key challenges of low-capacity binary reconstruction, projection-error propagation, and non-uniform structural sensitivity, AF1 effectively enables binarization without costly QAT. Specifically, NABF enhances binary reconstruction, while HiSA allocates the limited structural budget across layers and submodules. Experiments show that AF1 substantially outperforms existing binary PTQ methods under comparable or lower BPW, approaches the accuracy of QAT-based binarization methods with lower quantization cost, and brings practical memory reduction and decoding acceleration. These results demonstrate that AF1 offers a scalable and deployment-friendly pathway toward genuine 1-bit LLM compression.

\clearpage
\section*{Limitations}
\label{sec:limitations}
AF1 focuses on weight-only post-training binarization under a strict 1.0 effective-BPW budget for target linear weights. This setting substantially reduces model memory and decoding traffic, but it does not directly compress activations or the KV cache, which can become important in long-context or large-batch serving. In addition, our efficiency evaluation is based on a representative GPU implementation with optimized binary matrix multiplication; further system-level integration into broader inference stacks remains an important future direction. Finally, our experiments cover representative LLaMA, Qwen, and Gemma models and standard language benchmarks, while evaluations on more architectures, broader instruction-tuned variants, and real deployment workloads would further validate generality.

\section*{Ethical Considerations}
\label{sec:ethics}

AF1 is a model compression method and does not introduce new training data or modify the semantic objectives of the original models. Its main societal benefit is reducing the memory and bandwidth cost of LLM deployment, which can improve accessibility and lower energy consumption. However, more efficient LLM deployment may also lower the barrier to misuse of existing models, such as automated spam, misinformation generation, or other harmful applications. We therefore encourage deploying AF1-compressed models with the same safety, monitoring, and access-control mechanisms used for the corresponding full-precision models.

\bibliography{custom}

@misc{llama3herdmodels,
      title={The Llama 3 Herd of Models}, 
      author={Aaron Grattafiori and Abhimanyu Dubey and Abhinav Jauhri and Abhinav Pandey and Abhishek Kadian and Ahmad Al-Dahle and Aiesha Letman and Akhil Mathur and Alan Schelten and Alex Vaughan and Amy Yang and Angela Fan and Anirudh Goyal and Anthony Hartshorn and Aobo Yang and Archi Mitra and Archie Sravankumar and Artem Korenev and Arthur Hinsvark and Arun Rao and Aston Zhang and Aurelien Rodriguez and Austen Gregerson and Ava Spataru and Baptiste Roziere and Bethany Biron and Binh Tang and Bobbie Chern and Charlotte Caucheteux and Chaya Nayak and Chloe Bi and Chris Marra and Chris McConnell and Christian Keller and Christophe Touret and Chunyang Wu and Corinne Wong and Cristian Canton Ferrer and Cyrus Nikolaidis and Damien Allonsius and Daniel Song and Danielle Pintz and Danny Livshits and Danny Wyatt and David Esiobu and Dhruv Choudhary and Dhruv Mahajan and Diego Garcia-Olano and Diego Perino and Dieuwke Hupkes and Egor Lakomkin and Ehab AlBadawy and Elina Lobanova and Emily Dinan and Eric Michael Smith and Filip Radenovic and Francisco Guzmán and Frank Zhang and Gabriel Synnaeve and Gabrielle Lee and Georgia Lewis Anderson and Govind Thattai and Graeme Nail and Gregoire Mialon and Guan Pang and Guillem Cucurell and Hailey Nguyen and Hannah Korevaar and Hu Xu and Hugo Touvron and Iliyan Zarov and Imanol Arrieta Ibarra and Isabel Kloumann and Ishan Misra and Ivan Evtimov and Jack Zhang and Jade Copet and Jaewon Lee and Jan Geffert and Jana Vranes and Jason Park and Jay Mahadeokar and Jeet Shah and Jelmer van der Linde and Jennifer Billock and Jenny Hong and Jenya Lee and Jeremy Fu and Jianfeng Chi and Jianyu Huang and Jiawen Liu and Jie Wang and Jiecao Yu and Joanna Bitton and Joe Spisak and Jongsoo Park and Joseph Rocca and Joshua Johnstun and Joshua Saxe and Junteng Jia and Kalyan Vasuden Alwala and Karthik Prasad and Kartikeya Upasani and Kate Plawiak and Ke Li and Kenneth Heafield and Kevin Stone and Khalid El-Arini and Krithika Iyer and Kshitiz Malik and Kuenley Chiu and Kunal Bhalla and Kushal Lakhotia and Lauren Rantala-Yeary and Laurens van der Maaten and Lawrence Chen and Liang Tan and Liz Jenkins and Louis Martin and Lovish Madaan and Lubo Malo and Lukas Blecher and Lukas Landzaat and Luke de Oliveira and Madeline Muzzi and Mahesh Pasupuleti and Mannat Singh and Manohar Paluri and Marcin Kardas and Maria Tsimpoukelli and Mathew Oldham and Mathieu Rita and Maya Pavlova and Melanie Kambadur and Mike Lewis and Min Si and Mitesh Kumar Singh and Mona Hassan and Naman Goyal and Narjes Torabi and Nikolay Bashlykov and Nikolay Bogoychev and Niladri Chatterji and Ning Zhang and Olivier Duchenne and Onur Çelebi and Patrick Alrassy and Pengchuan Zhang and Pengwei Li and Petar Vasic and Peter Weng and Prajjwal Bhargava and Pratik Dubal and Praveen Krishnan and Punit Singh Koura and Puxin Xu and Qing He and Qingxiao Dong and Ragavan Srinivasan and Raj Ganapathy and Ramon Calderer and Ricardo Silveira Cabral and Robert Stojnic and Roberta Raileanu and Rohan Maheswari and Rohit Girdhar and Rohit Patel and Romain Sauvestre and Ronnie Polidoro and Roshan Sumbaly and Ross Taylor and Ruan Silva and Rui Hou and Rui Wang and Saghar Hosseini and Sahana Chennabasappa and Sanjay Singh and Sean Bell and Seohyun Sonia Kim and Sergey Edunov and Shaoliang Nie and Sharan Narang and Sharath Raparthy and Sheng Shen and Shengye Wan and Shruti Bhosale and Shun Zhang and Simon Vandenhende and Soumya Batra and Spencer Whitman and Sten Sootla and Stephane Collot and Suchin Gururangan and Sydney Borodinsky and Tamar Herman and Tara Fowler and Tarek Sheasha and Thomas Georgiou and Thomas Scialom and Tobias Speckbacher and Todor Mihaylov and Tong Xiao and Ujjwal Karn and Vedanuj Goswami and Vibhor Gupta and Vignesh Ramanathan and Viktor Kerkez and Vincent Gonguet and Virginie Do and Vish Vogeti and Vítor Albiero and Vladan Petrovic and Weiwei Chu and Wenhan Xiong and Wenyin Fu and Whitney Meers and Xavier Martinet and Xiaodong Wang and Xiaofang Wang and Xiaoqing Ellen Tan and Xide Xia and Xinfeng Xie and Xuchao Jia and Xuewei Wang and Yaelle Goldschlag and Yashesh Gaur and Yasmine Babaei and Yi Wen and Yiwen Song and Yuchen Zhang and Yue Li and Yuning Mao and Zacharie Delpierre Coudert and Zheng Yan and Zhengxing Chen and Zoe Papakipos and Aaditya Singh and Aayushi Srivastava and Abha Jain and Adam Kelsey and Adam Shajnfeld and Adithya Gangidi and Adolfo Victoria and Ahuva Goldstand and Ajay Menon and Ajay Sharma and Alex Boesenberg and Alexei Baevski and Allie Feinstein and Amanda Kallet and Amit Sangani and Amos Teo and Anam Yunus and Andrei Lupu and Andres Alvarado and Andrew Caples and Andrew Gu and Andrew Ho and Andrew Poulton and Andrew Ryan and Ankit Ramchandani and Annie Dong and Annie Franco and Anuj Goyal and Aparajita Saraf and Arkabandhu Chowdhury and Ashley Gabriel and Ashwin Bharambe and Assaf Eisenman and Azadeh Yazdan and Beau James and Ben Maurer and Benjamin Leonhardi and Bernie Huang and Beth Loyd and Beto De Paola and Bhargavi Paranjape and Bing Liu and Bo Wu and Boyu Ni and Braden Hancock and Bram Wasti and Brandon Spence and Brani Stojkovic and Brian Gamido and Britt Montalvo and Carl Parker and Carly Burton and Catalina Mejia and Ce Liu and Changhan Wang and Changkyu Kim and Chao Zhou and Chester Hu and Ching-Hsiang Chu and Chris Cai and Chris Tindal and Christoph Feichtenhofer and Cynthia Gao and Damon Civin and Dana Beaty and Daniel Kreymer and Daniel Li and David Adkins and David Xu and Davide Testuggine and Delia David and Devi Parikh and Diana Liskovich and Didem Foss and Dingkang Wang and Duc Le and Dustin Holland and Edward Dowling and Eissa Jamil and Elaine Montgomery and Eleonora Presani and Emily Hahn and Emily Wood and Eric-Tuan Le and Erik Brinkman and Esteban Arcaute and Evan Dunbar and Evan Smothers and Fei Sun and Felix Kreuk and Feng Tian and Filippos Kokkinos and Firat Ozgenel and Francesco Caggioni and Frank Kanayet and Frank Seide and Gabriela Medina Florez and Gabriella Schwarz and Gada Badeer and Georgia Swee and Gil Halpern and Grant Herman and Grigory Sizov and Guangyi and Zhang and Guna Lakshminarayanan and Hakan Inan and Hamid Shojanazeri and Han Zou and Hannah Wang and Hanwen Zha and Haroun Habeeb and Harrison Rudolph and Helen Suk and Henry Aspegren and Hunter Goldman and Hongyuan Zhan and Ibrahim Damlaj and Igor Molybog and Igor Tufanov and Ilias Leontiadis and Irina-Elena Veliche and Itai Gat and Jake Weissman and James Geboski and James Kohli and Janice Lam and Japhet Asher and Jean-Baptiste Gaya and Jeff Marcus and Jeff Tang and Jennifer Chan and Jenny Zhen and Jeremy Reizenstein and Jeremy Teboul and Jessica Zhong and Jian Jin and Jingyi Yang and Joe Cummings and Jon Carvill and Jon Shepard and Jonathan McPhie and Jonathan Torres and Josh Ginsburg and Junjie Wang and Kai Wu and Kam Hou U and Karan Saxena and Kartikay Khandelwal and Katayoun Zand and Kathy Matosich and Kaushik Veeraraghavan and Kelly Michelena and Keqian Li and Kiran Jagadeesh and Kun Huang and Kunal Chawla and Kyle Huang and Lailin Chen and Lakshya Garg and Lavender A and Leandro Silva and Lee Bell and Lei Zhang and Liangpeng Guo and Licheng Yu and Liron Moshkovich and Luca Wehrstedt and Madian Khabsa and Manav Avalani and Manish Bhatt and Martynas Mankus and Matan Hasson and Matthew Lennie and Matthias Reso and Maxim Groshev and Maxim Naumov and Maya Lathi and Meghan Keneally and Miao Liu and Michael L. Seltzer and Michal Valko and Michelle Restrepo and Mihir Patel and Mik Vyatskov and Mikayel Samvelyan and Mike Clark and Mike Macey and Mike Wang and Miquel Jubert Hermoso and Mo Metanat and Mohammad Rastegari and Munish Bansal and Nandhini Santhanam and Natascha Parks and Natasha White and Navyata Bawa and Nayan Singhal and Nick Egebo and Nicolas Usunier and Nikhil Mehta and Nikolay Pavlovich Laptev and Ning Dong and Norman Cheng and Oleg Chernoguz and Olivia Hart and Omkar Salpekar and Ozlem Kalinli and Parkin Kent and Parth Parekh and Paul Saab and Pavan Balaji and Pedro Rittner and Philip Bontrager and Pierre Roux and Piotr Dollar and Polina Zvyagina and Prashant Ratanchandani and Pritish Yuvraj and Qian Liang and Rachad Alao and Rachel Rodriguez and Rafi Ayub and Raghotham Murthy and Raghu Nayani and Rahul Mitra and Rangaprabhu Parthasarathy and Raymond Li and Rebekkah Hogan and Robin Battey and Rocky Wang and Russ Howes and Ruty Rinott and Sachin Mehta and Sachin Siby and Sai Jayesh Bondu and Samyak Datta and Sara Chugh and Sara Hunt and Sargun Dhillon and Sasha Sidorov and Satadru Pan and Saurabh Mahajan and Saurabh Verma and Seiji Yamamoto and Sharadh Ramaswamy and Shaun Lindsay and Shaun Lindsay and Sheng Feng and Shenghao Lin and Shengxin Cindy Zha and Shishir Patil and Shiva Shankar and Shuqiang Zhang and Shuqiang Zhang and Sinong Wang and Sneha Agarwal and Soji Sajuyigbe and Soumith Chintala and Stephanie Max and Stephen Chen and Steve Kehoe and Steve Satterfield and Sudarshan Govindaprasad and Sumit Gupta and Summer Deng and Sungmin Cho and Sunny Virk and Suraj Subramanian and Sy Choudhury and Sydney Goldman and Tal Remez and Tamar Glaser and Tamara Best and Thilo Koehler and Thomas Robinson and Tianhe Li and Tianjun Zhang and Tim Matthews and Timothy Chou and Tzook Shaked and Varun Vontimitta and Victoria Ajayi and Victoria Montanez and Vijai Mohan and Vinay Satish Kumar and Vishal Mangla and Vlad Ionescu and Vlad Poenaru and Vlad Tiberiu Mihailescu and Vladimir Ivanov and Wei Li and Wenchen Wang and Wenwen Jiang and Wes Bouaziz and Will Constable and Xiaocheng Tang and Xiaojian Wu and Xiaolan Wang and Xilun Wu and Xinbo Gao and Yaniv Kleinman and Yanjun Chen and Ye Hu and Ye Jia and Ye Qi and Yenda Li and Yilin Zhang and Ying Zhang and Yossi Adi and Youngjin Nam and Yu and Wang and Yu Zhao and Yuchen Hao and Yundi Qian and Yunlu Li and Yuzi He and Zach Rait and Zachary DeVito and Zef Rosnbrick and Zhaoduo Wen and Zhenyu Yang and Zhiwei Zhao and Zhiyu Ma},
      year={2024},
      eprint={2407.21783},
      archivePrefix={arXiv},
      primaryClass={cs.AI},
      url={https://arxiv.org/abs/2407.21783}, 
}

@misc{qwen3technicalreport,
      title={Qwen3 Technical Report}, 
      author={An Yang and Anfeng Li and Baosong Yang and Beichen Zhang and Binyuan Hui and Bo Zheng and Bowen Yu and Chang Gao and Chengen Huang and Chenxu Lv and Chujie Zheng and Dayiheng Liu and Fan Zhou and Fei Huang and Feng Hu and Hao Ge and Haoran Wei and Huan Lin and Jialong Tang and Jian Yang and Jianhong Tu and Jianwei Zhang and Jianxin Yang and Jiaxi Yang and Jing Zhou and Jingren Zhou and Junyang Lin and Kai Dang and Keqin Bao and Kexin Yang and Le Yu and Lianghao Deng and Mei Li and Mingfeng Xue and Mingze Li and Pei Zhang and Peng Wang and Qin Zhu and Rui Men and Ruize Gao and Shixuan Liu and Shuang Luo and Tianhao Li and Tianyi Tang and Wenbiao Yin and Xingzhang Ren and Xinyu Wang and Xinyu Zhang and Xuancheng Ren and Yang Fan and Yang Su and Yichang Zhang and Yinger Zhang and Yu Wan and Yuqiong Liu and Zekun Wang and Zeyu Cui and Zhenru Zhang and Zhipeng Zhou and Zihan Qiu},
      year={2025},
      eprint={2505.09388},
      archivePrefix={arXiv},
      primaryClass={cs.CL},
      url={https://arxiv.org/abs/2505.09388}, 
}

@misc{qwen3.5,
    title  = {{Qwen3.5}: Towards Native Multimodal Agents},
    author = {{Qwen Team}},
    month  = {February},
    year   = {2026},
    url    = {https://qwen.ai/blog?id=qwen3.5}
}

@misc{huang2024billmpushinglimitposttraining,
      title={BiLLM: Pushing the Limit of Post-Training Quantization for LLMs}, 
      author={Wei Huang and Yangdong Liu and Haotong Qin and Ying Li and Shiming Zhang and Xianglong Liu and Michele Magno and Xiaojuan Qi},
      year={2024},
      eprint={2402.04291},
      archivePrefix={arXiv},
      primaryClass={cs.LG},
      url={https://arxiv.org/abs/2402.04291}, 
}

@misc{li2024arbllmalternatingrefinedbinarizations,
      title={ARB-LLM: Alternating Refined Binarizations for Large Language Models}, 
      author={Zhiteng Li and Xianglong Yan and Tianao Zhang and Haotong Qin and Dong Xie and Jiang Tian and zhongchao shi and Linghe Kong and Yulun Zhang and Xiaokang Yang},
      year={2024},
      eprint={2410.03129},
      archivePrefix={arXiv},
      primaryClass={cs.CV},
      url={https://arxiv.org/abs/2410.03129}, 
}

@misc{shang2023pbllmpartiallybinarizedlarge,
      title={PB-LLM: Partially Binarized Large Language Models}, 
      author={Yuzhang Shang and Zhihang Yuan and Qiang Wu and Zhen Dong},
      year={2023},
      eprint={2310.00034},
      archivePrefix={arXiv},
      primaryClass={cs.LG},
      url={https://arxiv.org/abs/2310.00034}, 
}

@misc{clark2018thinksolvedquestionanswering,
      title={Think you have Solved Question Answering? Try ARC, the AI2 Reasoning Challenge}, 
      author={Peter Clark and Isaac Cowhey and Oren Etzioni and Tushar Khot and Ashish Sabharwal and Carissa Schoenick and Oyvind Tafjord},
      year={2018},
      eprint={1803.05457},
      archivePrefix={arXiv},
      primaryClass={cs.AI},
      url={https://arxiv.org/abs/1803.05457}, 
}

@misc{zellers2019hellaswagmachinereallyfinish,
      title={HellaSwag: Can a Machine Really Finish Your Sentence?}, 
      author={Rowan Zellers and Ari Holtzman and Yonatan Bisk and Ali Farhadi and Yejin Choi},
      year={2019},
      eprint={1905.07830},
      archivePrefix={arXiv},
      primaryClass={cs.CL},
      url={https://arxiv.org/abs/1905.07830}, 
}

@misc{paperno2016lambadadatasetwordprediction,
      title={The LAMBADA dataset: Word prediction requiring a broad discourse context}, 
      author={Denis Paperno and Germán Kruszewski and Angeliki Lazaridou and Quan Ngoc Pham and Raffaella Bernardi and Sandro Pezzelle and Marco Baroni and Gemma Boleda and Raquel Fernández},
      year={2016},
      eprint={1606.06031},
      archivePrefix={arXiv},
      primaryClass={cs.CL},
      url={https://arxiv.org/abs/1606.06031}, 
}

@misc{bisk2019piqareasoningphysicalcommonsense,
      title={PIQA: Reasoning about Physical Commonsense in Natural Language}, 
      author={Yonatan Bisk and Rowan Zellers and Ronan Le Bras and Jianfeng Gao and Yejin Choi},
      year={2019},
      eprint={1911.11641},
      archivePrefix={arXiv},
      primaryClass={cs.CL},
      url={https://arxiv.org/abs/1911.11641}, 
}

@misc{sakaguchi2019winograndeadversarialwinogradschema,
      title={WinoGrande: An Adversarial Winograd Schema Challenge at Scale}, 
      author={Keisuke Sakaguchi and Ronan Le Bras and Chandra Bhagavatula and Yejin Choi},
      year={2019},
      eprint={1907.10641},
      archivePrefix={arXiv},
      primaryClass={cs.CL},
      url={https://arxiv.org/abs/1907.10641}, 
}

@article{wikitext2,
  title={Pointer sentinel mixture models},
  author={Merity, Stephen and Xiong, Caiming and Bradbury, James and Socher, Richard},
  journal={arXiv preprint arXiv:1609.07843},
  year={2016}
}

@misc{humaneval,
      title={Evaluating Large Language Models Trained on Code}, 
      author={Mark Chen and Jerry Tworek and Heewoo Jun and Qiming Yuan and Henrique Ponde de Oliveira Pinto and Jared Kaplan and Harri Edwards and Yuri Burda and Nicholas Joseph and Greg Brockman and Alex Ray and Raul Puri and Gretchen Krueger and Michael Petrov and Heidy Khlaaf and Girish Sastry and Pamela Mishkin and Brooke Chan and Scott Gray and Nick Ryder and Mikhail Pavlov and Alethea Power and Lukasz Kaiser and Mohammad Bavarian and Clemens Winter and Philippe Tillet and Felipe Petroski Such and Dave Cummings and Matthias Plappert and Fotios Chantzis and Elizabeth Barnes and Ariel Herbert-Voss and William Hebgen Guss and Alex Nichol and Alex Paino and Nikolas Tezak and Jie Tang and Igor Babuschkin and Suchir Balaji and Shantanu Jain and William Saunders and Christopher Hesse and Andrew N. Carr and Jan Leike and Josh Achiam and Vedant Misra and Evan Morikawa and Alec Radford and Matthew Knight and Miles Brundage and Mira Murati and Katie Mayer and Peter Welinder and Bob McGrew and Dario Amodei and Sam McCandlish and Ilya Sutskever and Wojciech Zaremba},
      year={2021},
      eprint={2107.03374},
      archivePrefix={arXiv},
      primaryClass={cs.LG},
      url={https://arxiv.org/abs/2107.03374}, 
}

@misc{gsm8k,
      title={Training Verifiers to Solve Math Word Problems}, 
      author={Karl Cobbe and Vineet Kosaraju and Mohammad Bavarian and Mark Chen and Heewoo Jun and Lukasz Kaiser and Matthias Plappert and Jerry Tworek and Jacob Hilton and Reiichiro Nakano and Christopher Hesse and John Schulman},
      year={2021},
      eprint={2110.14168},
      archivePrefix={arXiv},
      primaryClass={cs.LG},
      url={https://arxiv.org/abs/2110.14168}, 
}

@inproceedings{zhao2025quark,
  title={QUARK: Quantization-enabled circuit sharing for transformer acceleration by exploiting common patterns in nonlinear operations},
  author={Zhao, Zhixiong and Li, Haomin and Liu, Fangxin and Lu, Yuncheng and Wang, Zongwu and Yang, Tao and Jiang, Li and Guan, Haibing},
  booktitle={2025 IEEE/ACM International Conference On Computer Aided Design (ICCAD)},
  pages={1--9},
  year={2025},
  organization={IEEE}
}

@inproceedings{zhao2026specquant,
  title={Specquant: Spectral decomposition and adaptive truncation for ultra-low-bit llms quantization},
  author={Zhao, Zhixiong and Liu, Fangxin and Wang, Junjie and Guan, Chenyang and Wang, Zongwu and Jiang, Li and Guan, Haibing},
  booktitle={Proceedings of the AAAI Conference on Artificial Intelligence},
  volume={40},
  number={34},
  pages={28786--28794},
  year={2026}
}

@article{xu2026kbvq,
  title={Kbvq-moe: Klt-guided svd with bias-corrected vector quantization for moe large language models},
  author={Xu, Zukang and Zhao, Zhixiong and Hu, Xing and Chen, Zhixuan and Yang, Dawei},
  journal={arXiv preprint arXiv:2602.11184},
  year={2026}
}

@article{zhao2026twla,
  title={TWLA: Achieving Ternary Weights and Low-Bit Activations for LLMs via Post-Training Quantization},
  author={Zhao, Zhixiong and Xu, Zukang and Chen, Zhixuan and Hu, Xing and Jiang, Zhe and Yang, Dawei},
  journal={arXiv preprint arXiv:2606.13054},
  year={2026}
}

@inproceedings{zhao2026bwla,
  title={Bwla: Breaking the barrier of w1ax post-training quantization for llms},
  author={Zhao, Zhixiong and Xu, Zukang and Yang, Dawei},
  booktitle={Proceedings of the 64th Annual Meeting of the Association for Computational Linguistics (Volume 1: Long Papers)},
  pages={19264--19290},
  year={2026}
}

@article{chen2025hbllm,
  title={HBLLM: A Haar-Based Approach for Accurate Structured 1-Bit Quantized LLMs},
  author={Chen, Ningning and Ye, Weicai and Jiang, Ying},
  journal={arXiv e-prints},
  pages={arXiv--2512},
  year={2025}
}

@article{xu2024onebit,
  title={Onebit: Towards extremely low-bit large language models},
  author={Xu, Yuzhuang and Han, Xu and Yang, Zonghan and Wang, Shuo and Zhu, Qingfu and Liu, Zhiyuan and Liu, Weidong and Che, Wanxiang},
  journal={Advances in Neural Information Processing Systems},
  volume={37},
  pages={66357--66382},
  year={2024}
}

@article{dong2024stbllm,
  title={Stbllm: Breaking the 1-bit barrier with structured binary llms},
  author={Dong, Peijie and Li, Lujun and Zhong, Yuedong and Du, Dayou and Fan, Ruibo and Chen, Yuhan and Tang, Zhenheng and Wang, Qiang and Xue, Wei and Guo, Yike and others},
  journal={arXiv preprint arXiv:2408.01803},
  year={2024}
}

@inproceedings{guan2024aptq,
  title={APTQ: Attention-aware post-training mixed-precision quantization for large language models},
  author={Guan, Ziyi and Huang, Hantao and Su, Yupeng and Huang, Hong and Wong, Ngai and Yu, Hao},
  booktitle={Proceedings of the 61st ACM/IEEE Design Automation Conference},
  pages={1--6},
  year={2024}
}

@article{dettmers2023spqr,
  title={Spqr: A sparse-quantized representation for near-lossless llm weight compression},
  author={Dettmers, Tim and Svirschevski, Ruslan and Egiazarian, Vage and Kuznedelev, Denis and Frantar, Elias and Ashkboos, Saleh and Borzunov, Alexander and Hoefler, Torsten and Alistarh, Dan},
  journal={arXiv preprint arXiv:2306.03078},
  year={2023}
}

@article{zhao2024atom,
  title={Atom: Low-bit quantization for efficient and accurate llm serving},
  author={Zhao, Yilong and Lin, Chien-Yu and Zhu, Kan and Ye, Zihao and Chen, Lequn and Zheng, Size and Ceze, Luis and Krishnamurthy, Arvind and Chen, Tianqi and Kasikci, Baris},
  journal={Proceedings of Machine Learning and Systems},
  volume={6},
  pages={196--209},
  year={2024}
}

@article{huang2024slim,
  title={SliM-LLM: Salience-driven mixed-precision quantization for large language models},
  author={Huang, Wei and Qin, Haotong and Liu, Yangdong and Li, Yawei and Liu, Qinshuo and Liu, Xianglong and Benini, Luca and Magno, Michele and Zhang, Shiming and Qi, Xiaojuan},
  journal={arXiv preprint arXiv:2405.14917},
  year={2024}
}

@article{chen2024channel,
  title={Channel-wise mixed-precision quantization for large language models},
  author={Chen, Zihan and Xie, Bike and Li, Jundong and Shen, Cong},
  journal={arXiv preprint arXiv:2410.13056},
  year={2024}
}

@inproceedings{lee2025amq,
  title={Amq: Enabling automl for mixed-precision weight-only quantization of large language models},
  author={Lee, Sangjun and Woo, Seung-taek and Jin, Jun-gyu and Lee, Changhun and Park, Eunhyeok},
  booktitle={Proceedings of the 2025 Conference on Empirical Methods in Natural Language Processing},
  pages={35520--35538},
  year={2025}
}

@article{fang2024alphaedit,
  title={Alphaedit: Null-space constrained knowledge editing for language models},
  author={Fang, Junfeng and Jiang, Houcheng and Wang, Kun and Ma, Yunshan and Jie, Shi and Wang, Xiang and He, Xiangnan and Chua, Tat-Seng},
  journal={arXiv preprint arXiv:2410.02355},
  year={2024}
}

@article{winter2002shapley,
  title={The shapley value},
  author={Winter, Eyal},
  journal={Handbook of game theory with economic applications},
  volume={3},
  pages={2025--2054},
  year={2002},
  publisher={Elsevier}
}

@article{zhao2025impq,
  title={IMPQ: Interaction-Aware Layerwise Mixed Precision Quantization for LLMs},
  author={Zhao, Junchen and Derakhshan, Ali and Bharadwaj, Dushyant and Kana Hyman, Jayden and Dong, Junhao and Abdu Jyothi, Sangeetha and Harris, Ian},
  journal={arXiv e-prints},
  pages={arXiv--2509},
  year={2025}
}

@inproceedings{martens2015optimizing,
  title={Optimizing neural networks with kronecker-factored approximate curvature},
  author={Martens, James and Grosse, Roger},
  booktitle={International conference on machine learning},
  pages={2408--2417},
  year={2015},
  organization={PMLR}
}

@article{bovza2025addition,
  title={Addition is almost all you need: Compressing neural networks with double binary factorization},
  author={Bo{\v{z}}a, Vladim{\'\i}r and Macko, Vladim{\'\i}r},
  journal={arXiv preprint arXiv:2505.11076},
  year={2025}
}

@inproceedings{wang2021training,
  title={Training networks in null space of feature covariance for continual learning},
  author={Wang, Shipeng and Li, Xiaorong and Sun, Jian and Xu, Zongben},
  booktitle={Proceedings of the IEEE/CVF conference on Computer Vision and Pattern Recognition},
  pages={184--193},
  year={2021}
}

@book{iglewicz1993volume,
  title={Volume 16: how to detect and handle outliers},
  author={Iglewicz, Boris and Hoaglin, David C},
  year={1993},
  publisher={Quality Press}
}

@article{zhang2026beyond,
  title={Beyond Outliers: A Data-Free Layer-wise Mixed-Precision Quantization Approach Driven by Numerical and Structural Dual-Sensitivity},
  author={Zhang, Hengyuan and Chen, Xinrong and Su, Zunhai and Liang, Xiao and Xiong, Jing and Xu, Wendong and Xiao, He and Tao, Chaofan and Zhang, Wei and Xie, Ruobing and others},
  journal={arXiv preprint arXiv:2603.17354},
  year={2026}
}

@misc{gemmateam2025gemma3technicalreport,
      title={Gemma 3 Technical Report}, 
      author={Gemma Team and Aishwarya Kamath and Johan Ferret and Shreya Pathak and Nino Vieillard and Ramona Merhej and Sarah Perrin and Tatiana Matejovicova and Alexandre Ramé and Morgane Rivière and Louis Rouillard and Thomas Mesnard and Geoffrey Cideron and Jean-bastien Grill and Sabela Ramos and Edouard Yvinec and Michelle Casbon and Etienne Pot and Ivo Penchev and Gaël Liu and Francesco Visin and Kathleen Kenealy and Lucas Beyer and Xiaohai Zhai and Anton Tsitsulin and Robert Busa-Fekete and Alex Feng and Noveen Sachdeva and Benjamin Coleman and Yi Gao and Basil Mustafa and Iain Barr and Emilio Parisotto and David Tian and Matan Eyal and Colin Cherry and Jan-Thorsten Peter and Danila Sinopalnikov and Surya Bhupatiraju and Rishabh Agarwal and Mehran Kazemi and Dan Malkin and Ravin Kumar and David Vilar and Idan Brusilovsky and Jiaming Luo and Andreas Steiner and Abe Friesen and Abhanshu Sharma and Abheesht Sharma and Adi Mayrav Gilady and Adrian Goedeckemeyer and Alaa Saade and Alex Feng and Alexander Kolesnikov and Alexei Bendebury and Alvin Abdagic and Amit Vadi and András György and André Susano Pinto and Anil Das and Ankur Bapna and Antoine Miech and Antoine Yang and Antonia Paterson and Ashish Shenoy and Ayan Chakrabarti and Bilal Piot and Bo Wu and Bobak Shahriari and Bryce Petrini and Charlie Chen and Charline Le Lan and Christopher A. Choquette-Choo and CJ Carey and Cormac Brick and Daniel Deutsch and Danielle Eisenbud and Dee Cattle and Derek Cheng and Dimitris Paparas and Divyashree Shivakumar Sreepathihalli and Doug Reid and Dustin Tran and Dustin Zelle and Eric Noland and Erwin Huizenga and Eugene Kharitonov and Frederick Liu and Gagik Amirkhanyan and Glenn Cameron and Hadi Hashemi and Hanna Klimczak-Plucińska and Harman Singh and Harsh Mehta and Harshal Tushar Lehri and Hussein Hazimeh and Ian Ballantyne and Idan Szpektor and Ivan Nardini and Jean Pouget-Abadie and Jetha Chan and Joe Stanton and John Wieting and Jonathan Lai and Jordi Orbay and Joseph Fernandez and Josh Newlan and Ju-yeong Ji and Jyotinder Singh and Kat Black and Kathy Yu and Kevin Hui and Kiran Vodrahalli and Klaus Greff and Linhai Qiu and Marcella Valentine and Marina Coelho and Marvin Ritter and Matt Hoffman and Matthew Watson and Mayank Chaturvedi and Michael Moynihan and Min Ma and Nabila Babar and Natasha Noy and Nathan Byrd and Nick Roy and Nikola Momchev and Nilay Chauhan and Noveen Sachdeva and Oskar Bunyan and Pankil Botarda and Paul Caron and Paul Kishan Rubenstein and Phil Culliton and Philipp Schmid and Pier Giuseppe Sessa and Pingmei Xu and Piotr Stanczyk and Pouya Tafti and Rakesh Shivanna and Renjie Wu and Renke Pan and Reza Rokni and Rob Willoughby and Rohith Vallu and Ryan Mullins and Sammy Jerome and Sara Smoot and Sertan Girgin and Shariq Iqbal and Shashir Reddy and Shruti Sheth and Siim Põder and Sijal Bhatnagar and Sindhu Raghuram Panyam and Sivan Eiger and Susan Zhang and Tianqi Liu and Trevor Yacovone and Tyler Liechty and Uday Kalra and Utku Evci and Vedant Misra and Vincent Roseberry and Vlad Feinberg and Vlad Kolesnikov and Woohyun Han and Woosuk Kwon and Xi Chen and Yinlam Chow and Yuvein Zhu and Zichuan Wei and Zoltan Egyed and Victor Cotruta and Minh Giang and Phoebe Kirk and Anand Rao and Kat Black and Nabila Babar and Jessica Lo and Erica Moreira and Luiz Gustavo Martins and Omar Sanseviero and Lucas Gonzalez and Zach Gleicher and Tris Warkentin and Vahab Mirrokni and Evan Senter and Eli Collins and Joelle Barral and Zoubin Ghahramani and Raia Hadsell and Yossi Matias and D. Sculley and Slav Petrov and Noah Fiedel and Noam Shazeer and Oriol Vinyals and Jeff Dean and Demis Hassabis and Koray Kavukcuoglu and Clement Farabet and Elena Buchatskaya and Jean-Baptiste Alayrac and Rohan Anil and Dmitry and Lepikhin and Sebastian Borgeaud and Olivier Bachem and Armand Joulin and Alek Andreev and Cassidy Hardin and Robert Dadashi and Léonard Hussenot},
      year={2025},
      eprint={2503.19786},
      archivePrefix={arXiv},
      primaryClass={cs.CL},
      url={https://arxiv.org/abs/2503.19786}, 
}

@article{wang2024mmlu,
  title={Mmlu-pro: A more robust and challenging multi-task language understanding benchmark},
  author={Wang, Yubo and Ma, Xueguang and Zhang, Ge and Ni, Yuansheng and Chandra, Abhranil and Guo, Shiguang and Ren, Weiming and Arulraj, Aaran and He, Xuan and Jiang, Ziyan and others},
  journal={Advances in Neural Information Processing Systems},
  volume={37},
  pages={95266--95290},
  year={2024}
}

@article{lee2026littlebit,
  title={LittleBit: Ultra low-bit quantization via latent factorization},
  author={Lee, Banseok and Kim, Dongkyu and You, Youngcheon and Kim, Youngmin},
  journal={Advances in Neural Information Processing Systems},
  volume={38},
  pages={116379--116411},
  year={2026}
}

@misc{evalscope_2024,
    title={{EvalScope}: Evaluation Framework for Large Models},
    author={ModelScope Team},
    year={2024},
    url={https://github.com/modelscope/evalscope}
}

@misc{eval-harness,
  author       = {Gao, Leo and Tow, Jonathan and Abbasi, Baber and Biderman, Stella and Black, Sid and DiPofi, Anthony and Foster, Charles and Golding, Laurence and Hsu, Jeffrey and Le Noac'h, Alain and Li, Haonan and McDonell, Kyle and Muennighoff, Niklas and Ociepa, Chris and Phang, Jason and Reynolds, Laria and Schoelkopf, Hailey and Skowron, Aviya and Sutawika, Lintang and Tang, Eric and Thite, Anish and Wang, Ben and Wang, Kevin and Zou, Andy},
  title        = {The Language Model Evaluation Harness},
  month        = 07,
  year         = 2024,
  publisher    = {Zenodo},
  version      = {v0.4.3},
  doi          = {10.5281/zenodo.12608602},
  url          = {https://zenodo.org/records/12608602}
}

@inproceedings{kwon2022alphatuning,
  title={Alphatuning: Quantization-aware parameter-efficient adaptation of large-scale pre-trained language models},
  author={Kwon, Se Jung and Kim, Jeonghoon and Bae, Jeongin and Yoo, Kang Min and Kim, Jin-Hwa and Park, Baeseong and Kim, Byeongwook and Ha, Jung-Woo and Sung, Nako and Lee, Dongsoo},
  booktitle={Findings of the Association for Computational Linguistics: EMNLP 2022},
  pages={3288--3305},
  year={2022}
}

@misc{openai2026gpt55,
  title        = {{GPT-5.5 System Card}},
  author       = {{OpenAI}},
  year         = {2026},
  month        = apr,
  howpublished = {\url{https://openai.com/index/gpt-5-5-system-card/}},
  note         = {Accessed: 2026-05-14}
}

@article{neal2011distributed,
  title={Distributed optimization and statistical learning via the alternating direction method of multipliers},
  author={Neal, Parikh and Eric, Chu and Borja, Peleato and Jonathan, Eckstein},
  journal={Foundations and Trends{\textregistered} in Machine learning},
  volume={3},
  number={1},
  pages={1--122},
  year={2011},
  publisher={Emerald Publishing Limited}
}

@misc{badri2023hqq,
title  = {Half-Quadratic Quantization of Large Machine Learning Models},
url    = {https://dropbox.github.io/hqq_blog/},
author = {Hicham Badri and Appu Shaji},
month  = {November},
year   = {2023}
}

@article{Jolliffe2016PrincipalCA,
  title={Principal component analysis: a review and recent developments},
  author={Ian T. Jolliffe and Jorge Cadima},
  journal={Philosophical Transactions of the Royal Society A: Mathematical, Physical and Engineering Sciences},
  year={2016},
  volume={374},
  url={https://api.semanticscholar.org/CorpusID:20101754}
}

@misc{weber2024redpajamaopendatasettraining,
      title={RedPajama: an Open Dataset for Training Large Language Models}, 
      author={Maurice Weber and Daniel Fu and Quentin Anthony and Yonatan Oren and Shane Adams and Anton Alexandrov and Xiaozhong Lyu and Huu Nguyen and Xiaozhe Yao and Virginia Adams and Ben Athiwaratkun and Rahul Chalamala and Kezhen Chen and Max Ryabinin and Tri Dao and Percy Liang and Christopher Ré and Irina Rish and Ce Zhang},
      year={2024},
      eprint={2411.12372},
      archivePrefix={arXiv},
      primaryClass={cs.CL},
      url={https://arxiv.org/abs/2411.12372}, 
}

@inproceedings{ye-etal-2026-tableqa,
    title = "When {T}able{QA} Meets Noise: A Dual Denoising Framework for Complex Questions and Large-scale Tables",
    author = "Ye, Shenghao  and
      Guo, Yu  and
      Jin, Dong  and
      Wang, Yuxiang  and
      Shen, Yikai  and
      Hou, Yunpeng  and
      Chen, Shuangwu  and
      Jianyang  and
      Jiang, Xiaofeng",
    editor = "Liakata, Maria  and
      Moreira, Viviane P.  and
      Zhang, Jiajun  and
      Jurgens, David",
    booktitle = "Proceedings of the 64th Annual Meeting of the {A}ssociation for {C}omputational {L}inguistics (Volume 1: Long Papers)",
    month = jul,
    year = "2026",
    address = "San Diego, California, United States",
    publisher = "Association for Computational Linguistics",
    url = "https://aclanthology.org/2026.acl-long.1102/",
    doi = "10.18653/v1/2026.acl-long.1102",
    pages = "24022--24045",
    ISBN = "979-8-89176-390-6"
}

@inproceedings{guo-etal-2026-rethinking-table,
    title = "Rethinking Table Pruning in {T}able{QA}: From Sequential Revisions to Gold Trajectory-Supervised Parallel Search",
    author = "Guo, Yu  and
      Ye, Shenghao  and
      Chen, Shuangwu  and
      Wen, Zijian  and
      Zhang, Tao  and
      Qirui, Bai  and
      Jin, Dong  and
      Hou, Yunpeng  and
      He, Huasen  and
      Jianyang  and
      Tan, Xiaobin",
    editor = "Liakata, Maria  and
      Moreira, Viviane P.  and
      Zhang, Jiajun  and
      Jurgens, David",
    booktitle = "Proceedings of the 64th Annual Meeting of the {A}ssociation for {C}omputational {L}inguistics (Volume 1: Long Papers)",
    month = jul,
    year = "2026",
    address = "San Diego, California, United States",
    publisher = "Association for Computational Linguistics",
    url = "https://aclanthology.org/2026.acl-long.591/",
    doi = "10.18653/v1/2026.acl-long.591",
    pages = "12960--12976",
    ISBN = "979-8-89176-390-6"
}

@misc{ye2026rubricguidedprocessrewardstepwise,
      title={Rubric-Guided Process Reward for Stepwise Model Routing}, 
      author={Shenghao Ye and Yu Guo and Zhengheng Li and Shuangwu Chen and Jian Yang},
      year={2026},
      eprint={2605.29310},
      archivePrefix={arXiv},
      primaryClass={cs.AI},
      url={https://arxiv.org/abs/2605.29310}, 
}

@misc{ye2026rethinkingstepwisemodelrouting,
      title={Rethinking Stepwise Model Routing: A Cost-Efficient Table Reasoning Perspective}, 
      author={Shenghao Ye and Yuxiang Wang and Yu Guo and Dong Jin and Shuangwu Chen and Jian Yang},
      year={2026},
      eprint={2605.29319},
      archivePrefix={arXiv},
      primaryClass={cs.CL},
      url={https://arxiv.org/abs/2605.29319}, 
}

@inproceedings{guo-etal-2025-sqlforge,
    title = "{SQLF}orge: Synthesizing Reliable and Diverse Data to Enhance Text-to-{SQL} Reasoning in {LLM}s",
    author = "Guo, Yu  and
      Jin, Dong  and
      Ye, Shenghao  and
      Chen, Shuangwu  and
      Yang, Jian  and
      Tan, Xiaobin",
    editor = "Che, Wanxiang  and
      Nabende, Joyce  and
      Shutova, Ekaterina  and
      Pilehvar, Mohammad Taher",
    booktitle = "Findings of the Association for Computational Linguistics: ACL 2025",
    month = jul,
    year = "2025",
    address = "Vienna, Austria",
    publisher = "Association for Computational Linguistics",
    url = "https://aclanthology.org/2025.findings-acl.443/",
    doi = "10.18653/v1/2025.findings-acl.443",
    pages = "8441--8452",
    ISBN = "979-8-89176-256-5"
}

@inproceedings{hao2026rethinking,
  title={Rethinking entropy interventions in rlvr: An entropy change perspective},
  author={Hao, Zhezheng and Wang, Hong and Liu, Haoyang and Luo, Jian and Yu, Jiarui and Dong, Hande and Lin, Qiang and Wang, Can and Chen, Jiawei},
  booktitle={Proceedings of the 64th Annual Meeting of the Association for Computational Linguistics (Volume 1: Long Papers)},
  pages={31105--31133},
  year={2026}
}

@inproceedings{hao2026recreate,
  title={Recreate: Reasoning and creating domain agents driven by experience},
  author={Hao, Zhezheng and Wang, Hong and Luo, Jian and Zhang, Jianqing and Zhou, Yuyan and Lin, Qiang and Wang, Can and Dong, Hande and Chen, Jiawei},
  booktitle={Proceedings of the 64th Annual Meeting of the Association for Computational Linguistics (Volume 1: Long Papers)},
  pages={31018--31046},
  year={2026}
}

@article{hao2026evolve,
  title={Evolve as a Team: Collaborative Self-Evolution for LLM-based Multi-Agent Systems},
  author={Hao, Zhezheng and Wang, Tianfu and Dong, Huanshuo and Liu, Ziyan and Wang, Hong and Lin, Xiankun and Lin, Qiang and Wang, Can and Dong, Hande and Chen, Jiawei},
  journal={arXiv preprint arXiv:2605.29790},
  year={2026}
}

@inproceedings{wang2026scheduling,
  title={Scheduling your llm reinforcement learning with reasoning trees},
  author={Wang, Hong and Hao, Zhezheng and Luo, Jian and Wei, Chenxing and Shu, Yao and Liu, Lei and Lin, Qiang and Dong, Hande and Chen, Jiawei},
  booktitle={International Conference on Learning Representations},
  volume={2026},
  pages={154734--154753},
  year={2026}
}

@article{liu2026meta,
  title={Meta-Cognitive Memory Policy Optimization for Long-Horizon LLM Agents},
  author={Liu, Ziyan and Hao, Zhezheng and Chen, Yeqiu and Wang, Hong and Hou, Jingren and Ding, Ruiyi and Yang, Yongkang and Ji, Wence and Xia, Wei and Liu, Feng},
  journal={arXiv preprint arXiv:2605.30159},
  year={2026}
}

@inproceedings{zhu2025pathology,
  title={Pathology-Aware Prototype Evolution via LLM-Driven Semantic Disambiguation for Multicenter Diabetic Retinopathy Diagnosis},
  author={Zhu, Chunzheng and Lin, Yangfang and Shao, Jialin and Lin, Jianxin and Wang, Yijun},
  booktitle={Proceedings of the 33rd ACM International Conference on Multimedia},
  pages={9196--9205},
  year={2025}
}

@inproceedings{zhu2026medeyes,
  title={MedEyes: Learning Dynamic Visual Focus for Medical Progressive Diagnosis},
  author={Zhu, Chunzheng and Lin, Yangfang and Chen, Shen and Wang, Yijun and Lin, Jianxin},
  booktitle={Proceedings of the AAAI Conference on Artificial Intelligence},
  volume={40},
  number={16},
  pages={13916--13924},
  year={2026}
}

@article{lin2026medcausalx,
  title={MedCausalX: Adaptive Causal Reasoning with Self-Reflection for Trustworthy Medical Vision-Language Models},
  author={Lin, Jianxin and Zhu, Chunzheng and Kneuertz, Peter J and Bai, Yunfei and Xue, Yuan},
  journal={arXiv preprint arXiv:2603.23085},
  year={2026}
}

@misc{zhu2026medsynapsevbridgingvisualperception,
      title={MedSynapse-V: Bridging Visual Perception and Clinical Intuition via Latent Memory Evolution}, 
      author={Chunzheng Zhu and Jiaqi Zeng and Junyu Jiang and Jianxin Lin and Yijun Wang},
      year={2026},
      eprint={2604.26283},
      archivePrefix={arXiv},
      primaryClass={cs.CV},
      url={https://arxiv.org/abs/2604.26283}, 
}

@inproceedings{zhu2024advancing,
  title={Advancing Ultrasound Medical Continuous Learning with Task-Specific Generalization and Adaptability},
  author={Zhu, Chunzheng and Lin, Jianxin and Tan, Guanghua and Zhu, Ningbo and Li, Kenli and Wang, Chunlian and Li, Shengli},
  booktitle={2024 IEEE International Conference on Bioinformatics and Biomedicine (BIBM)},
  pages={3019--3025},
  year={2024},
  organization={IEEE}
}

@article{zhu2025fmri2ges,
  title={fMRI2GES: Co-speech Gesture Reconstruction from fMRI Signal with Dual Brain Decoding Alignment},
  author={Zhu, Chunzheng and Shao, Jialin and Lin, Jianxin and Wang, Yijun and Wang, Jing and Tang, Jinhui and Li, Kenli},
  journal={IEEE Transactions on Circuits and Systems for Video Technology},
  year={2025},
  publisher={IEEE}
}

@article{zhu2026anatomy,
  title={Anatomy-Anchored Self-Supervision: Distilling Vision Foundation Models for Invariant Ultrasound Representation},
  author={Zhu, Chunzheng and Wang, Yijun and Lin, Jianxin and Wang, Feng and Wang, Hongwei and Zhao, Lei and Li, Shengli and Li, Kenli},
  journal={MICCAI 2026 (Early Accept)},
  year={2026}
}

@misc{li2026mindmarginboundarydistilled,
    title={Mind Your Margin and Boundary: Are Your Distilled Datasets Truly Robust?},
    author={Muquan Li and Yingyi Ma and Yihong Huang and Hang Gou and Ke Qin and Ming Li and Yuan-Fang Li and Tao He},
    year={2026},
    eprint={2605.20606},
    archivePrefix={arXiv},
    primaryClass={cs.CV},
    url={https://arxiv.org/abs/2605.20606},
}

@misc{li2026fixedanchorsenoughdynamic,
title={Fixed Anchors Are Not Enough: Dynamic Retrieval and Persistent Homology for Dataset Distillation},
author={Muquan Li and Hang Gou and Yingyi Ma and Rongzheng Wang and Ke Qin and Tao He},
year={2026},
eprint={2602.24144},
archivePrefix={arXiv},
primaryClass={cs.CV},
url={https://arxiv.org/abs/2602.24144},
}

@misc{li2026randomautomaticinnerloopoptimization,
title={Beyond Random: Automatic Inner-loop Optimization in Dataset Distillation},
author={Muquan Li and Hang Gou and Dongyang Zhang and Shuang Liang and Xiurui Xie and Deqiang Ouyang and Ke Qin},
year={2026},
eprint={2510.04838},
archivePrefix={arXiv},
primaryClass={cs.CV},
url={https://arxiv.org/abs/2510.04838},
}

@inproceedings{Li_2024, series={MM ’24},
title={Towards Effective Data-Free Knowledge Distillation via Diverse Diffusion Augmentation},
url={http://dx.doi.org/10.1145/3664647.3680711},
DOI={10.1145/3664647.3680711},
booktitle={Proceedings of the 32nd ACM International Conference on Multimedia},
publisher={ACM},
author={Li, Muquan and Zhang, Dongyang and He, Tao and Xie, Xiurui and Li, Yuan-Fang and Qin, Ke},
year={2024},
month=Oct, pages={4416–4425},
collection={MM ’24} }

@inproceedings{li2025adaptive, 
title={Adaptive dataset quantization}, 
author={Li, Muquan and Zhang, Dongyang and Dong, Qiang and Xie, Xiurui and Qin, Ke}, 
booktitle={Proceedings of the AAAI Conference on Artificial Intelligence}, 
volume={39}, 
number={11}, 
pages={12093--12101}, 
year={2025} 
}

@article{yang2025robuq, 
title={Robuq: Pushing dits to w1. 58a2 via robust activation quantization}, 
author={Yang, Kaicheng and Zhang, Xun and Qin, Haotong and Lin, Yucheng and Yang, Kaisen and Yan, Xianglong and Zhang, Yulun}, journal={arXiv preprint arXiv:2509.23582}, 
year={2025} }

@inproceedings{yan2026pt, 
title={PT $^2$ -LLM: Post-Training Ternarization for Large Language Models}, 
author={Yan, Xianglong and Bao, Chengzhu and Li, Zhiteng and Zhang, Tianao and Yang, Kaicheng and Qin, Haotong and Xie, Ruobing and Sun, Samm and Zhang, Yulun}, 
booktitle={International Conference on Learning Representations}, volume={2026}, 
pages={11677--11689}, 
year={2026} }

@article{yang2025treeq, 
title={TreeQ: Pushing the Quantization Boundary of Diffusion Transformer via Tree-Structured Mixed-Precision Search}, 
author={Yang, Kaicheng and Yang, Kaisen and Wu, Baiting and Zhang, Xun and Yang, Qianrui and Qin, Haotong and Zhang, He and Zhang, Yulun}, 
journal={arXiv preprint arXiv:2512.06353}, 
year={2025} }

@article{zhu2025qartsr, 
title={QArtSR: Quantization via Reverse-Module and Timestep-Retraining in One-Step Diffusion based Image Super-Resolution}, author={Zhu, Libo and Qin, Haotong and Yang, Kaicheng and Li, Wenbo and Guo, Yong and Zhang, Yulun and Rahardja, Susanto and Yang, Xiaokang}, 
journal={arXiv preprint arXiv:2503.05584}, 
year={2025} }

@article{zhang2026q, 
title={Q-DiT4SR: Exploration of Detail-Preserving Diffusion Transformer Quantization for Real-World Image Super-Resolution}, author={Zhang, Xun and Yang, Kaicheng and Lu, Hongliang and Qin, Haotong and Guo, Yong and Zhang, Yulun}, 
journal={arXiv preprint arXiv:2602.01273}, 
year={2026} }

\clearpage
\appendix

\section{Appendix}
\label{sec:appendix}

\section*{Appendix Overview}
\begin{itemize}
    \item Section~\ref{app:algorithm}: Detailed Algorithms of AF1.
    \begin{itemize}
        \item Section~\ref{app:overall_af1}: Overall Workflow of AF1.
        \item Section~\ref{app:nabf_algorithm}: Null-space-Aware Binary Factorization.
        \item Section~\ref{app:hisa_algorithm}: Hierarchical Shapley Allocation.
        \item Section~\ref{app:dbf_admm}: ADMM-SVID Framework in DBF.
    \end{itemize}
    \item Section~\ref{sec:proof}: Detailed Proofs.
    \begin{itemize}
        \item Section~\ref{app:double_factor_capacity}: Representation Capacity of Double-Factor Binary Parameterization.
        \item Section~\ref{app:nullspace_proof}: Equivalence Between the Null Eigenspace of $\mathbf{B}\mathbf{B}^{\top}$ and the Left Null Space of $\mathbf{B}$.
        \item Section~\ref{app:adaptive_nullspace_cutoff}: Adaptive Residual-Energy Cutoff.
        \item Section~\ref{app:projector_proof}: Proof of the Approximate Null-Space Projection Property.
        \item Section~\ref{app:gamma_proof}: Closed-form Solution for Scale Compensation.
        \item Section~\ref{app:shapley_interaction}: Interaction-aware Property of Shapley Allocation.
        \item Section~\ref{app:dual_distortion}: Dual Distortion Decomposition for Block-Level Allocation.
        \item Section~\ref{app:size_density}: Motivation for Size-Aware Sensitivity Density.
    \end{itemize}
    \item Section~\ref{app:bpw_binary_ptq}: Effective Bit-Width of Binary PTQ Methods.
    \item Section~\ref{sec:more ablation}: More Experimental Results.
    \begin{itemize}
        \item Section~\ref{app:detailed main results}: More Detailed Results.
        \item Section~\ref{app:ablation_nullspace_threshold}: Ablation on the Residual-Energy Threshold.
        \item Section~\ref{app:ablation_hisa_metric}: Ablation on Allocation-Sensitivity Metrics.
        \item Section~\ref{app:hisa_more_ablation}: Additional Ablation on HiSA Hyperparameters.
        \item Section~\ref{app:ablation_calibration_data}: Ablation on Calibration Data.
    \end{itemize}
    \item Section~\ref{sec:dialog}: Dialog Examples.
    \item Section~\ref{sec:LLMs}: Use of Large Language Models.
\end{itemize}

\subsection{Detailed Algorithms of AF1}
\label{app:algorithm}

This appendix provides the algorithmic workflow of AF1. We present three pseudocode blocks corresponding to the overall PTQ pipeline, Null-space-Aware Binary Factorization (NABF), and Hierarchical Shapley Allocation (HiSA), respectively.

\subsubsection{Overall Workflow of AF1}
\label{app:overall_af1}

\paragraph{Pipeline description.}
AF1 combines hierarchical structural allocation with operator-level binary reconstruction. Given a pretrained model and a calibration set, AF1 first collects the statistics required by HiSA and NABF. HiSA determines the layer-wise and module-wise structural budgets, while NABF reconstructs each target linear operator under the assigned intermediate dimension. After all target operators are replaced, AF1 freezes the binary sign matrices and performs scale-only global reconstruction. The overall procedure is summarized in Algorithm~\ref{alg:af1_overall}.

\begin{algorithm*}[t!]
\caption{AF1: Overall PTQ Pipeline}
\label{alg:af1_overall}
\KwIn{Pretrained LLM $M$; calibration set $\mathcal{D}_{\mathrm{cal}}$; total budget $B_{\mathrm{total}}$; original dimension $r_{\mathrm{orig}}$; proxy dimension $r_{\mathrm{proxy}}$}
\KwOut{Compressed model $\hat M$}

Collect calibration statistics from $M$ on $\mathcal{D}_{\mathrm{cal}}$\;

$\{B_l\},\{b_i\} \leftarrow \mathrm{HiSA}(M,\mathcal{D}_{\mathrm{cal}},B_{\mathrm{total}},r_{\mathrm{orig}},r_{\mathrm{proxy}})$\;

\For{each target linear operator $\mathbf{W}_i$}{
    $r_i \leftarrow \mathrm{RankFromBudget}(b_i,\mathbf{W}_i)$\;
    $\hat{\mathbf{W}}_i \leftarrow \mathrm{NABF}(\mathbf{W}_i,r_i,\mathcal{D}_{\mathrm{cal}})$\;
    Replace $\mathbf{W}_i$ in $M$ with $\hat{\mathbf{W}}_i$\;
}

Freeze all binary sign matrices\;
Collect learnable scale vectors $\mathcal{S}$\;
Optimize only $\mathcal{S}$ using the KL objective in Eq.~\ref{eq:scale_only_kl}\;
Assemble the final compressed model $\hat M$\;

\Return{$\hat M$}
\end{algorithm*}

\subsubsection{Null-space-Aware Binary Factorization}
\label{app:nabf_algorithm}

\paragraph{Hessian-aware surrogate space.}
NABF operates on each target linear operator under the intermediate dimension allocated by HiSA. It first constructs a Hessian-aware surrogate weight by scaling the original weight with input-activation and output-gradient statistics. This converts the Hessian-weighted reconstruction problem into a Frobenius reconstruction problem in the surrogate space, while retaining task-sensitive channel information.

\paragraph{Binary factorization and null-space compensation.}
After surrogate reparameterization, NABF uses a double-factor binary structure with two binary sign matrices and several continuous scale vectors. The continuous factors are updated by alternating ADMM steps and then projected back to the binary-scaled domain through SVID. Since SVID projection introduces discrete residuals, NABF constructs an approximate null-space projector from the fixed factor and folds the compensated residual into the existing intermediate scaling vectors. The complete procedure is given in Algorithm~\ref{alg:nabf}.

\paragraph{Compact operators.}
To keep the single-column pseudocode readable, we use several compact operators. $\mathrm{ADMMLeft}$ and $\mathrm{ADMMRight}$ denote the closed-form ADMM proxy updates in the main text. $\mathrm{NullProj}(\mathbf{C},\delta)$ returns the projector spanned by eigenvectors of $\mathbf{C}$ whose eigenvalues are below $\delta$. $\mathrm{ColScaleFit}(\mathbf{X},\mathbf{Y})$ solves the column-wise least-squares scaling from $\mathbf{X}$ to $\mathbf{Y}$, and $\mathrm{RowScaleFit}(\mathbf{X},\mathbf{Y})$ is its row-wise counterpart.

\begin{algorithm*}[t!]
\caption{NABF: Null-space-Aware Binary Factorization}
\label{alg:nabf}
\KwIn{Weight $\mathbf{W}$; calibration activations $\mathbf{x}$; output gradients $\mathbf{g}$; intermediate dimension $r$; ADMM penalty $\rho$; max iterations $T$; threshold $\delta$}
\KwOut{Binary factors $\mathbf{A}_{\pm1},\mathbf{B}_{\pm1}$ and scales $\mathbf{a},\mathbf{m}_1,\mathbf{m}_2,\mathbf{b}$}

Compute $i_j=\sqrt{\mathbb{E}[x_j^2]}$ and $o_u=\sqrt{\mathbb{E}[g_u^2]}$\;
Construct $\mathbf{W}'=\mathbf{o}\odot\mathbf{W}\odot\mathbf{i}^{\top}$\;

Initialize $\mathbf{A}_{\pm1},\mathbf{B}_{\pm1},\mathbf{a},\mathbf{m}_1,\mathbf{m}_2,\mathbf{b}$ by SVID or low-rank approximation\;
Form scaled factors $\mathbf{A}$ and $\mathbf{B}$\;
Initialize ADMM dual variables $\mathbf{U}_A$ and $\mathbf{U}_B$\;

\For{$t=1$ \KwTo $T$}{
    \tcp{Left-factor update}
    $\hat{\mathbf{A}} \leftarrow \mathrm{ADMMLeft}(\mathbf{W}',\mathbf{A},\mathbf{B},\mathbf{U}_A,\rho)$\;
    $\mathbf{A}_q \leftarrow \mathrm{SVID}(\hat{\mathbf{A}}+\mathbf{U}_A)$\;
    $\mathbf{E}_A \leftarrow \hat{\mathbf{A}}-\mathbf{A}_q$\;
    $\mathbf{P}_B \leftarrow \mathrm{NullProj}(\mathbf{B}\mathbf{B}^{\top},\delta)$\;
    $\mathbf{A}_{\mathrm{target}} \leftarrow \hat{\mathbf{A}}-\mathbf{E}_A\mathbf{P}_B$\;
    $\boldsymbol{\gamma}^{*}\leftarrow \mathrm{ColScaleFit}(\mathbf{A}_q,\mathbf{A}_{\mathrm{target}})$\;
    $\mathbf{m}_1 \leftarrow \mathbf{m}_1\odot\boldsymbol{\gamma}^{*}$\;
    Update scaled factor $\mathbf{A}$\;

    \tcp{Right-factor update}
    $\hat{\mathbf{B}} \leftarrow \mathrm{ADMMRight}(\mathbf{W}',\mathbf{A},\mathbf{B},\mathbf{U}_B,\rho)$\;
    $\mathbf{B}_q \leftarrow \mathrm{SVID}(\hat{\mathbf{B}}+\mathbf{U}_B)$\;
    $\mathbf{E}_B \leftarrow \hat{\mathbf{B}}-\mathbf{B}_q$\;
    $\mathbf{P}_A \leftarrow \mathrm{NullProj}(\mathbf{A}^{\top}\mathbf{A},\delta)$\;
    $\mathbf{B}_{\mathrm{target}} \leftarrow \hat{\mathbf{B}}-\mathbf{P}_A\mathbf{E}_B$\;
    $\boldsymbol{\eta}^{*}\leftarrow \mathrm{RowScaleFit}(\mathbf{B}_q,\mathbf{B}_{\mathrm{target}})$\;
    $\mathbf{m}_2 \leftarrow \boldsymbol{\eta}^{*}\odot\mathbf{m}_2$\;
    Update scaled factor $\mathbf{B}$\;

    Update ADMM dual variables $\mathbf{U}_A$ and $\mathbf{U}_B$\;
}

\Return{$\mathbf{A}_{\pm1},\mathbf{B}_{\pm1},\mathbf{a},\mathbf{m}_1,\mathbf{m}_2,\mathbf{b}$}
\end{algorithm*}

\subsubsection{Hierarchical Shapley Allocation}
\label{app:hisa_algorithm}

\paragraph{Layer-level allocation.}
HiSA first estimates the global structural sensitivity of each Transformer layer. Instead of evaluating layers independently, it progressively degrades layers from the original intermediate dimension to a proxy dimension under sampled permutations, and measures the resulting change in end-to-end NLL. This gives a layer-level Shapley estimate that reflects cross-layer interactions and error propagation.

\paragraph{Block-level allocation.}
Given the layer budget, HiSA further assigns capacity to submodules within each layer. For each sampled submodule permutation, HiSA progressively degrades submodules and measures the distortion of the layer output. Directional and magnitude distortions are estimated separately, normalized by MAD, and fused by a Soft-OR rule. The fused score is converted into a size-aware density and then into a module-wise budget. The complete allocation procedure is shown in Algorithm~\ref{alg:hisa}.

\paragraph{Budget-to-rank conversion.}
After HiSA returns module-wise budgets, AF1 converts each budget into the intermediate dimension used by NABF. For a target matrix $\mathbf{W}_i\in\mathbb{R}^{d_{\mathrm{out}}\times d_{\mathrm{in}}}$, the dominant binary storage of the double-factor representation is $r_i(d_{\mathrm{out}}+d_{\mathrm{in}})$ bits. Thus, the practical rank can be computed as
\begin{equation}
r_i =
\operatorname{Round}
\left(
\frac{
b_i d_{\mathrm{out}}d_{\mathrm{in}}
}{
d_{\mathrm{out}}+d_{\mathrm{in}}
}
\right),
\end{equation}
with clipping to a feasible range in implementation.

\begin{algorithm*}[t!]
\caption{HiSA: Hierarchical Shapley Allocation}
\label{alg:hisa}
\KwIn{Model $M$; calibration set $\mathcal{D}_{\mathrm{cal}}$; layer set $T$; submodules $\mathcal{N}_l$; total budget $B_{\mathrm{total}}$; dimensions $r_{\mathrm{orig}},r_{\mathrm{proxy}}$; temperature $\tau$; smoothing exponent $\alpha$; permutation numbers $M_{\mathrm{layer}},M_{\mathrm{block}}$}
\KwOut{Layer budgets $\{B_l\}$ and module budgets $\{b_i\}$}

Initialize $\Phi_l=0$ for all layers\;

\For{$m=1$ \KwTo $M_{\mathrm{layer}}$}{
    Sample a random layer permutation $\pi_m$\;
    Initialize $S=T$ and evaluate $v_{\mathrm{NLL}}(S)$\;

    \For{each layer $l$ following $\pi_m$}{
        Degrade $l$ from $r_{\mathrm{orig}}$ to $r_{\mathrm{proxy}}$\;
        Evaluate $v_{\mathrm{NLL}}(S\setminus\{l\})$\;
        $\Phi_l \leftarrow \Phi_l+v_{\mathrm{NLL}}(S\setminus\{l\})-v_{\mathrm{NLL}}(S)$\;
        $S\leftarrow S\setminus\{l\}$\;
    }
}
$\Phi_l\leftarrow \Phi_l/M_{\mathrm{layer}}$ for all layers\;
$\{B_l\}\leftarrow \mathrm{SoftmaxBudget}(\{\Phi_l\},B_{\mathrm{total}},\tau)$\;

\For{each layer $l=1,\dots,L$}{
    Compute reference output $H_l^{\mathrm{ref}}$\;
    Initialize $\Phi_{\mathrm{dir}}^{(i)}=0$ and $\Phi_{\mathrm{mag}}^{(i)}=0$ for all $i\in\mathcal{N}_l$\;

    \For{$m=1$ \KwTo $M_{\mathrm{block}}$}{
        Sample a random submodule permutation $\pi_m$\;
        Initialize $S_0=\varnothing$, $\hat E_{\mathrm{dir}}(S_0)=0$, and $\hat E_{\mathrm{mag}}(S_0)=0$\;

        \For{each submodule $i$ following $\pi_m$}{
            Degrade $i$ to the low-capacity state and compute $H_l(S_t)$\;
            $E_{\mathrm{dir}}(S_t)\leftarrow r_{\mathrm{dir}}(H_l^{\mathrm{ref}},H_l(S_t))$\;
            $E_{\mathrm{mag}}(S_t)\leftarrow r_{\mathrm{mag}}(H_l^{\mathrm{ref}},H_l(S_t))$\;
            $\hat E_c(S_t)\leftarrow \max(E_c(S_t),\hat E_c(S_{t-1}))$, $c\in\{\mathrm{dir},\mathrm{mag}\}$\;
            $\Phi_c^{(i)}\leftarrow \Phi_c^{(i)}+\hat E_c(S_t)-\hat E_c(S_{t-1})$, $c\in\{\mathrm{dir},\mathrm{mag}\}$\;
        }
    }

    $\Phi_c^{(i)}\leftarrow \Phi_c^{(i)}/M_{\mathrm{block}}$, $c\in\{\mathrm{dir},\mathrm{mag}\}$\;
    $z_c^{(i)}\leftarrow \mathrm{MADNorm}(\Phi_c^{(i)})$, $c\in\{\mathrm{dir},\mathrm{mag}\}$\;
    $P_c^{(i)}\leftarrow 1/(1+\exp(-z_c^{(i)}))$\;
    $v_{\mathrm{base}}^{(i)}\leftarrow 1-(1-P_{\mathrm{dir}}^{(i)})(1-P_{\mathrm{mag}}^{(i)})$\;
    $\omega_i\leftarrow \left(v_{\mathrm{base}}^{(i)}/(|\mathbf{W}_i|+\epsilon)+\epsilon\right)^{\alpha}$\;
    $b_i\leftarrow \mathrm{ModuleBudget}(B_l,\omega_i,|\mathbf{W}_i|)$ for all $i\in\mathcal{N}_l$\;
}

\Return{$\{B_l\},\{b_i\}$}
\end{algorithm*}

\subsubsection{ADMM-SVID Framework in DBF}
\label{app:dbf_admm}

\paragraph{Double-binary factorization objective.}
NABF builds upon the double-binary factorization framework introduced in DBF~\citep{bovza2025addition}. Given a surrogate weight matrix $\mathbf{W}'$, the goal is to approximate it by two binary-scaled factors:
\begin{equation}
\resizebox{\columnwidth}{!}{$
    \displaystyle
\begin{aligned}
\min \quad
&
\left\|
\mathbf{W}'
-
\left(
    \mathbf{a}\odot\mathbf{A}_{\pm1}\odot\mathbf{m}^{\top}
\right)
\left(
    \mathbf{B}_{\pm1}\odot\mathbf{b}^{\top}
\right)
\right\|_F^2 .
\end{aligned}
$}
\end{equation}
Following DBF, the middle scaling vector is split into two parts, which can be merged back after optimization:
\begin{equation}
\resizebox{\columnwidth}{!}{$
    \displaystyle
\begin{aligned}
\min \quad
&
\left\|
\mathbf{W}'
-
\left(
    \mathbf{a}\odot\mathbf{A}_{\pm1}\odot\mathbf{m}_1^{\top}
\right)
\left(
    \mathbf{m}_2\odot\mathbf{B}_{\pm1}\odot\mathbf{b}^{\top}
\right)
\right\|_F^2 .
\end{aligned}
$}
\end{equation}
We denote the two scaled factors as
\begin{equation}
\begin{aligned}
\mathbf{A}
&=
\mathbf{a}\odot\mathbf{A}_{\pm1}\odot\mathbf{m}_1^{\top},
\\
\mathbf{B}
&=
\mathbf{m}_2\odot\mathbf{B}_{\pm1}\odot\mathbf{b}^{\top}.
\end{aligned}
\end{equation}
The factorization is then optimized by alternating minimization: fixing $\mathbf{B}$ to update $\mathbf{A}$, and fixing $\mathbf{A}$ to update $\mathbf{B}$.

\paragraph{ADMM subproblem.}
When updating the left factor with $\mathbf{B}$ fixed, the constrained subproblem is
\begin{equation}
\begin{aligned}
\min_{\mathbf{A}}
\quad
&
\left\|
\mathbf{A}\mathbf{B}
-
\mathbf{W}'
\right\|_F^2,
\\
\mathrm{s.t.}
\quad
&
\mathbf{A}
=
\mathbf{a}\odot\mathbf{A}_{\pm1}\odot\mathbf{m}_1^{\top}.
\end{aligned}
\end{equation}
This constraint is non-convex because of the binary sign matrix. DBF solves this subproblem using ADMM~\citep{neal2011distributed}. For a generic constrained problem
\begin{equation}
    \min_{\mathbf{X}} f(\mathbf{X}),
    \qquad
    \mathrm{s.t.}\ \mathbf{X}\in\mathcal{C},
\end{equation}
one ADMM iteration is
\begin{equation}
\begin{aligned}
\mathbf{X}^{(k+1)}
&=
\arg\min_{\mathbf{X}}
f(\mathbf{X})
+
\frac{\rho}{2}
\left\|
\mathbf{X}
-
\mathbf{Z}^{(k)}
+
\mathbf{U}^{(k)}
\right\|_F^2,
\\
\mathbf{Z}^{(k+1)}
&=
\Pi_{\mathcal{C}}
\left(
\mathbf{X}^{(k+1)}
+
\mathbf{U}^{(k)}
\right),
\\
\mathbf{U}^{(k+1)}
&=
\mathbf{U}^{(k)}
+
\mathbf{X}^{(k+1)}
-
\mathbf{Z}^{(k+1)},
\end{aligned}
\end{equation}
where $\rho$ is the penalty coefficient, $\mathbf{U}$ is the scaled dual variable, and $\Pi_{\mathcal{C}}$ denotes Euclidean projection onto the feasible set.

\paragraph{SVID projection.}
In DBF, the feasible set $\mathcal{C}$ consists of matrices that can be written as $\mathbf{a}\odot\mathbf{A}_{\pm1}\odot\mathbf{m}_1^{\top}$. The projection $\Pi_{\mathcal{C}}$ is implemented by the SVID projection~\citep{xu2024onebit}. Given a matrix $\mathbf{Z}$, SVID first sets
\begin{equation}
    \mathbf{Z}_{\pm1}=\operatorname{Sign}(\mathbf{Z}),
\end{equation}
and then approximates the magnitude matrix by a rank-1 factorization:
\begin{equation}
    |\mathbf{Z}|
    \approx
    \mathbf{a}\mathbf{m}_1^{\top}.
\end{equation}
The projected matrix is therefore
\begin{equation}
    \mathrm{SVID}(\mathbf{Z})
    =
    \mathbf{a}
    \odot
    \mathbf{Z}_{\pm1}
    \odot
    \mathbf{m}_1^{\top}.
\end{equation}
In practice, the rank-1 magnitude approximation can be efficiently computed by power iteration.

\paragraph{Closed-form proxy updates.}
For the left-factor update, applying ADMM to the above subproblem gives the continuous proxy
\begin{equation}
\begin{aligned}
\hat{\mathbf{A}}^{(k+1)}
&=
\left(
\mathbf{W}'\mathbf{B}^{\top}
+
\rho
\left(
\mathbf{A}^{(k)}
-
\mathbf{U}^{(k)}
\right)
\right)
\\
&\quad
\left(
\mathbf{B}\mathbf{B}^{\top}
+
\rho\mathbf{I}
\right)^{-1},
\end{aligned}
\end{equation}
followed by SVID projection and dual update:
\begin{equation}
\begin{aligned}
\mathbf{A}^{(k+1)}
&=
\mathrm{SVID}
\left(
\hat{\mathbf{A}}^{(k+1)}
+
\mathbf{U}^{(k)}
\right),
\\
\mathbf{U}^{(k+1)}
&=
\mathbf{U}^{(k)}
+
\hat{\mathbf{A}}^{(k+1)}
-
\mathbf{A}^{(k+1)}.
\end{aligned}
\end{equation}
The right-factor update is derived symmetrically:
\begin{equation}
\begin{aligned}
\hat{\mathbf{B}}^{(k+1)}
&=
\left(
\mathbf{A}^{\top}\mathbf{A}
+
\rho\mathbf{I}
\right)^{-1}
\\
&\quad
\left(
\mathbf{A}^{\top}\mathbf{W}'
+
\rho
\left(
\mathbf{B}^{(k)}
-
\mathbf{U}^{(k)}
\right)
\right),
\end{aligned}
\end{equation}
followed by SVID projection onto the right-factor binary-scaled structure. NABF uses this ADMM-SVID procedure as the baseline discrete optimization framework, and then introduces null-space-aware compensation to suppress the projection residual left by SVID.

\subsection{Detailed Proofs}
\label{sec:proof}

\subsubsection{Representation Capacity of Double-Factor Binary Parameterization}
\label{app:double_factor_capacity}

We provide a justification for why the double-factor binary parameterization used in NABF has higher representation capacity than a single binary-scaled matrix. The argument follows the general motivation of double-factor compression in DBF/DSF~\citep{bovza2025addition}: instead of approximating a weight matrix by one constrained matrix, representing it as the product of two constrained factors introduces intermediate paths that increase expressiveness under a controlled structural budget.

\paragraph{Single-factor binary-scaled representation.}
A standard SVID-style binary-scaled matrix can be written as
\begin{equation}
\begin{aligned}
    \widehat{\mathbf{W}}_{\mathrm{single}}
    =
    \mathbf{a}
    \odot
    \mathbf{S}
    \odot
    \mathbf{b}^{\top}
    =
    \operatorname{Diag}(\mathbf{a})
    \mathbf{S}
    \operatorname{Diag}(\mathbf{b}),
    \\
    \mathbf{S}\in\{-1,+1\}^{n\times m}.
\end{aligned}
\end{equation}
Although the sign matrix $\mathbf{S}$ is fully binary and flexible, the magnitude part is separable:
\begin{equation}
    \left|
    \widehat{\mathbf{W}}_{\mathrm{single},ij}
    \right|
    =
    |a_i||b_j|.
\end{equation}
Therefore, the magnitude matrix of a single-factor binary-scaled representation is rank one. Equivalently, for any two rows $i_1,i_2$ and two columns $j_1,j_2$, it must satisfy the multiplicative consistency constraint
\begin{equation}
\begin{aligned}
&
\left|
\widehat{\mathbf{W}}_{\mathrm{single},i_1j_1}
\right|
\left|
\widehat{\mathbf{W}}_{\mathrm{single},i_2j_2}
\right|
\\
&\qquad =
\left|
\widehat{\mathbf{W}}_{\mathrm{single},i_1j_2}
\right|
\left|
\widehat{\mathbf{W}}_{\mathrm{single},i_2j_1}
\right|.
\end{aligned}
\label{eq:single_rank1_constraint}
\end{equation}
This constraint limits the ability of a single binary-scaled matrix to fit heterogeneous magnitude patterns, which are common in LLM weights after Hessian-aware reparameterization.

\paragraph{Double-factor binary representation.}
NABF instead uses
\begin{equation}
    \widehat{\mathbf{W}}_{\mathrm{double}}
    =
    \left(
        \mathbf{a}
        \odot
        \mathbf{A}_{\pm1}
        \odot
        \mathbf{m}_1^{\top}
    \right)
    \left(
        \mathbf{m}_2
        \odot
        \mathbf{B}_{\pm1}
        \odot
        \mathbf{b}^{\top}
    \right),
\end{equation}
where $\mathbf{A}_{\pm1}\in\{-1,+1\}^{n\times r}$ and $\mathbf{B}_{\pm1}\in\{-1,+1\}^{r\times m}$. Let
\begin{equation}
    \boldsymbol{\lambda}
    =
    \mathbf{m}_1
    \odot
    \mathbf{m}_2.
\end{equation}
Then the double-factor form can be equivalently written as
\begin{equation}
    \widehat{\mathbf{W}}_{\mathrm{double}}
    =
    \operatorname{Diag}(\mathbf{a})
    \mathbf{A}_{\pm1}
    \operatorname{Diag}(\boldsymbol{\lambda})
    \mathbf{B}_{\pm1}
    \operatorname{Diag}(\mathbf{b}).
\label{eq:double_factor_equiv}
\end{equation}
Thus, each entry is given by
\begin{equation}
    \widehat{\mathbf{W}}_{\mathrm{double},ij}
    =
    a_i b_j
    \sum_{k=1}^{r}
    \lambda_k
    A_{\pm1,ik}
    B_{\pm1,kj}.
\label{eq:binary_paths}
\end{equation}
Equation~\ref{eq:binary_paths} shows that each input-output connection is no longer determined by a single binary sign and a separable magnitude term. Instead, it is the sum of $r$ binary paths through the intermediate dimension. The path coefficients $\lambda_k$ provide continuous degrees of freedom, while $\mathbf{A}_{\pm1}$ and $\mathbf{B}_{\pm1}$ keep the dominant parameters binary.

\paragraph{Strictly richer magnitude patterns.}
The double-factor form can represent magnitude patterns that violate the rank-one constraint in Eq.~\ref{eq:single_rank1_constraint}. Consider a $2\times2$ example with $\mathbf{a}=\mathbf{b}=\mathbf{1}$, $r=2$,
\begin{equation}
    \mathbf{A}_{\pm1}
    =
    \begin{bmatrix}
    1 & 1 \\
    1 & -1
    \end{bmatrix},
    \quad
    \mathbf{B}_{\pm1}
    =
    \begin{bmatrix}
    1 & 1 \\
    1 & -1
    \end{bmatrix},
    \quad
    \boldsymbol{\lambda}
    =
    \begin{bmatrix}
    1 \\
    1/2
    \end{bmatrix}.
\end{equation}
Then
\begin{equation}
\begin{aligned}
    \mathbf{A}_{\pm1}
    \operatorname{Diag}(\boldsymbol{\lambda})
    \mathbf{B}_{\pm1}
    &=
    \begin{bmatrix}
    1 & 1 \\
    1 & -1
    \end{bmatrix}
    \begin{bmatrix}
    1 & 0 \\
    0 & 1/2
    \end{bmatrix}
    \begin{bmatrix}
    1 & 1 \\
    1 & -1
    \end{bmatrix}
    \\
    &=
    \begin{bmatrix}
    3/2 & 1/2 \\
    1/2 & 3/2
    \end{bmatrix}.
\end{aligned}
\label{eq:double_example}
\end{equation}
The magnitude matrix in Eq.~\ref{eq:double_example} has determinant
\begin{equation}
    \frac{3}{2}\cdot\frac{3}{2}
    -
    \frac{1}{2}\cdot\frac{1}{2}
    =
    2
    \neq
    0.
\end{equation}
Hence, it is not rank one and cannot be represented by any single-factor binary-scaled form $\operatorname{Diag}(\mathbf{a})\mathbf{S}\operatorname{Diag}(\mathbf{b})$, whose magnitude matrix must satisfy Eq.~\ref{eq:single_rank1_constraint}. This example shows that even with $r=2$, the double-factor form can express non-separable magnitude structures that are impossible for a single binary-scaled matrix.

\paragraph{Path-sum interpretation.}
The capacity gain can also be understood from Eq.~\ref{eq:binary_paths}. For a fixed pair $(i,j)$, the normalized entry
\begin{equation}
    \frac{
    \widehat{\mathbf{W}}_{\mathrm{double},ij}
    }{
    a_i b_j
    }
    =
    \sum_{k=1}^{r}
    \lambda_k
    A_{\pm1,ik}
    B_{\pm1,kj}
\end{equation}
is a signed combination of $r$ path weights. With distinct $\lambda_k$, different input-output pairs can realize different sums depending on the binary path signs. Therefore, the model can produce multiple effective magnitude levels and correlation patterns, instead of being restricted to the separable magnitude $|a_i||b_j|$ of a single-factor representation.

\paragraph{Implication for NABF.}
The above analysis explains why NABF adopts a double-factor binary parameterization before applying null-space compensation. The double-factor structure increases the representational capacity available under a genuine 1-bit storage budget by introducing an intermediate binary path space. At the same time, the dominant matrices $\mathbf{A}_{\pm1}$ and $\mathbf{B}_{\pm1}$ remain binary, and the additional continuous parameters are limited to lightweight scale vectors. This provides a better starting point for ADMM-SVID projection and allows the subsequent null-space-aware scale recalibration to absorb discrete projection errors without introducing dense correction matrices.

\subsubsection{Equivalence Between the Null Eigenspace of $\mathbf{B}\mathbf{B}^{\top}$ and the Left Null Space of $\mathbf{B}$}
\label{app:nullspace_proof}

We prove that the null eigenspace of the uncentered covariance matrix $\mathbf{B}\mathbf{B}^{\top}$ is equivalent to the left null space of $\mathbf{B}$. This justifies constructing the approximate null-space projector from $\mathbf{C}_B=\mathbf{B}\mathbf{B}^{\top}$ in NABF.

Let $\mathbf{B}\in\mathbb{R}^{r\times d}$ and define
\begin{equation}
    \mathbf{C}_B=\mathbf{B}\mathbf{B}^{\top}\in\mathbb{R}^{r\times r}.
\end{equation}
For any vector $\mathbf{z}\in\mathbb{R}^{r}$, we have
\begin{equation}
\begin{aligned}
    \mathbf{z}^{\top}\mathbf{C}_B\mathbf{z}
    &=
    \mathbf{z}^{\top}
    \mathbf{B}\mathbf{B}^{\top}
    \mathbf{z}  \\
    &=
    \left(
        \mathbf{B}^{\top}\mathbf{z}
    \right)^{\top}
    \left(
        \mathbf{B}^{\top}\mathbf{z}
    \right)  \\
    &=
    \left\|
        \mathbf{B}^{\top}\mathbf{z}
    \right\|_2^2 .
\end{aligned}
\end{equation}
Since $\mathbf{C}_B$ is positive semidefinite, $\mathbf{z}$ belongs to the null eigenspace of $\mathbf{C}_B$ if and only if the above quadratic form is zero:
\begin{equation}
\begin{aligned}
    \mathbf{z}\in\mathrm{Null}(\mathbf{C}_B)
    &\Longleftrightarrow
    \mathbf{z}^{\top}\mathbf{C}_B\mathbf{z}=0  \\
    &\Longleftrightarrow
    \left\|
        \mathbf{B}^{\top}\mathbf{z}
    \right\|_2^2=0  \\
    &\Longleftrightarrow
    \mathbf{B}^{\top}\mathbf{z}=\mathbf{0}.
\end{aligned}
\end{equation}
Therefore,
\begin{equation}
    \mathrm{Null}
    \left(
        \mathbf{B}\mathbf{B}^{\top}
    \right)
    =
    \mathrm{Null}
    \left(
        \mathbf{B}^{\top}
    \right).
\end{equation}
That is, the zero-eigenvalue eigenspace of $\mathbf{B}\mathbf{B}^{\top}$ is exactly the left null space of $\mathbf{B}$.

Let $\widehat{\mathbf{U}}$ be the eigenvectors of $\mathbf{C}_B$ corresponding to zero eigenvalues, and define
\begin{equation}
    \mathbf{P}_B
    =
    \widehat{\mathbf{U}}
    \widehat{\mathbf{U}}^{\top}.
\end{equation}
Then $\mathbf{P}_B$ is the orthogonal projector onto $\mathrm{Null}(\mathbf{B}^{\top})$. For any vector $\mathbf{v}$, $\mathbf{P}_B\mathbf{v}\in\mathrm{Null}(\mathbf{B}^{\top})$, and thus
\begin{equation}
    \mathbf{B}^{\top}
    \mathbf{P}_B\mathbf{v}
    =
    \mathbf{0}.
\end{equation}
Equivalently, this implies
\begin{equation}
    \mathbf{P}_B\mathbf{B}
    =
    \mathbf{0}.
\end{equation}
Hence, projecting a residual component onto this subspace makes it vanish after multiplication by the fixed factor $\mathbf{B}$:
\begin{equation}
    \left(
        \mathbf{E}_A\mathbf{P}_B
    \right)
    \mathbf{B}
    =
    \mathbf{E}_A
    \left(
        \mathbf{P}_B\mathbf{B}
    \right)
    =
    \mathbf{0}.
\end{equation}
This property is the basis of the null-space compensation used in NABF.

In practice, due to numerical noise and the finite intermediate dimension, the eigenvalues of $\mathbf{C}_B$ are rarely exactly zero. Therefore, we use eigenvectors associated with zero or near-zero eigenvalues to construct an approximate projector $\mathbf{P}_B$, which satisfies
\begin{equation}
    \mathbf{P}_B\mathbf{B}
    \approx
    \mathbf{0}.
\end{equation}
This approximate property is sufficient for suppressing the propagated projection residual $\mathbf{E}_A\mathbf{B}$.

\paragraph{Symmetric case for the right-factor update.}
The same argument applies when updating the right factor. Given a fixed left factor $\mathbf{A}\in\mathbb{R}^{n\times r}$, define
\begin{equation}
    \mathbf{C}_A=\mathbf{A}^{\top}\mathbf{A}\in\mathbb{R}^{r\times r}.
\end{equation}
For any $\mathbf{z}\in\mathbb{R}^{r}$,
\begin{equation}
    \mathbf{z}^{\top}\mathbf{C}_A\mathbf{z}
    =
    \left\|
        \mathbf{A}\mathbf{z}
    \right\|_2^2.
\end{equation}
Thus,
\begin{equation}
    \mathrm{Null}
    \left(
        \mathbf{A}^{\top}\mathbf{A}
    \right)
    =
    \mathrm{Null}(\mathbf{A}).
\end{equation}
If $\mathbf{P}_A$ is constructed from the zero or near-zero eigenspace of $\mathbf{A}^{\top}\mathbf{A}$, then
\begin{equation}
    \mathbf{A}\mathbf{P}_A
    \approx
    \mathbf{0}.
\end{equation}
Consequently, for the right-factor projection residual $\mathbf{E}_B$, the compensated component satisfies
\begin{equation}
    \mathbf{A}
    \left(
        \mathbf{P}_A\mathbf{E}_B
    \right)
    =
    \left(
        \mathbf{A}\mathbf{P}_A
    \right)
    \mathbf{E}_B
    \approx
    \mathbf{0}.
\end{equation}
This gives the symmetric null-space compensation used for the right-factor update.

\subsubsection{Adaptive Residual-Energy Cutoff}
\label{app:adaptive_nullspace_cutoff}

We describe how the cutoff index used in the adaptive residual-energy criterion is determined.
Given the decomposition
\begin{equation}
\begin{aligned}
\mathbf{C}_B
=
\mathbf{U}\boldsymbol{\Lambda}\mathbf{U}^{\top},
\quad
\boldsymbol{\Lambda}
=
\mathrm{diag}(\lambda_1,\ldots,\lambda_r),
\\
\qquad
\lambda_1\geq\lambda_2\geq\cdots\geq\lambda_r\geq 0,
\end{aligned}
\end{equation}
the leading eigenspace captures most of the spectral energy, while the trailing eigenspace contains the remaining low-energy directions. 
A fixed absolute eigenvalue threshold is sensitive to the scale of $\mathbf{C}_B$, which may vary across layers and submodules. 
We therefore determine the cutoff by the fraction of residual spectral energy.

For a candidate cutoff index $k$, we define the total and residual spectral energies as
\begin{equation}
E_{\mathrm{tot}}
=
\sum_{i=1}^{r}\lambda_i,
\qquad
E_{\mathrm{res}}(k)
=
\sum_{i=k+1}^{r}\lambda_i.
\end{equation}
The residual-energy fraction is then
\begin{equation}
\rho_k
=
\frac{E_{\mathrm{res}}(k)}
{E_{\mathrm{tot}}+\epsilon}
=
\frac{\sum_{i=k+1}^{r}\lambda_i}
{\sum_{i=1}^{r}\lambda_i+\epsilon},
\label{eq:app_residual_energy_fraction}
\end{equation}
where $\epsilon$ is a small constant for numerical stability. 
We choose the first index $k$ satisfying
\begin{equation}
\rho_k \leq \eta,
\label{eq:app_residual_energy_condition}
\end{equation}
where $\eta$ controls the maximum fraction of spectral energy assigned to the trailing subspace. 
Equivalently, the retained eigenspace explains at least $1-\eta$ of the total spectral energy. 
The selected approximate null-space basis is then
\begin{equation}
\widehat{\mathbf{U}}
=
\mathbf{U}_{:,k+1:r}.
\end{equation}

This criterion is scale-invariant: multiplying $\mathbf{C}_B$ by any positive constant scales both $E_{\mathrm{tot}}$ and $E_{\mathrm{res}}(k)$ by the same factor, leaving $\rho_k$ unchanged up to the numerical stabilizer $\epsilon$. 
Therefore, the cutoff adapts to the spectrum of each factor covariance matrix and avoids manually choosing a global absolute eigenvalue threshold across layers and submodules.

\subsubsection{Proof of the Approximate Null-Space Projection Property}
\label{app:projector_proof}

We prove that the projector constructed from near-zero singular vectors satisfies $\mathbf{P}_B\mathbf{B}\approx\mathbf{0}$. Let $\widehat{\mathbf{U}}=[\mathbf{u}_1,\ldots,\mathbf{u}_r]$ collect the singular vectors of $\mathbf{C}_B=\mathbf{B}\mathbf{B}^{\top}$ whose singular values are zero or near-zero. The projector is defined as
\begin{equation}
    \mathbf{P}_B
    =
    \widehat{\mathbf{U}}\widehat{\mathbf{U}}^{\top}.
\end{equation}
For each selected vector $\mathbf{u}_k$, the near-zero singular value condition gives
\begin{equation}
    \mathbf{C}_B\mathbf{u}_k
    \approx
    \mathbf{0}.
\end{equation}
Since $\mathbf{C}_B=\mathbf{B}\mathbf{B}^{\top}$ is positive semi-definite, we have
\begin{equation}
    \mathbf{u}_k^{\top}\mathbf{C}_B\mathbf{u}_k
    =
    \mathbf{u}_k^{\top}\mathbf{B}\mathbf{B}^{\top}\mathbf{u}_k
    =
    \|\mathbf{B}^{\top}\mathbf{u}_k\|_2^2
    \approx
    0.
\end{equation}
Thus, $\mathbf{B}^{\top}\mathbf{u}_k\approx\mathbf{0}$, or equivalently $\mathbf{u}_k^{\top}\mathbf{B}\approx\mathbf{0}$. Stacking all selected vectors yields
\begin{equation}
    \widehat{\mathbf{U}}^{\top}\mathbf{B}
    \approx
    \mathbf{0}.
\end{equation}
Therefore,
\begin{equation}
    \mathbf{P}_B\mathbf{B}
    =
    \widehat{\mathbf{U}}\widehat{\mathbf{U}}^{\top}\mathbf{B}
    \approx
    \mathbf{0}.
\end{equation}
This shows that $\mathbf{P}_B$ approximately projects onto the left null space of $\mathbf{B}$, which justifies its use for steering projection errors into directions that minimally affect the product with $\mathbf{B}$.

\subsubsection{Closed-form Solution for Scale Compensation}
\label{app:gamma_proof}

We provide the derivation of the closed-form solution used for scale compensation in NABF. After null-space compensation, the continuous target $\mathbf{A}_{\mathrm{target}}$ preserves the propagated behavior of the ADMM proxy, but it cannot be directly deployed as a binary factor. Therefore, we absorb its effect into the existing intermediate scaling vector by solving
\begin{equation}
    \boldsymbol{\gamma}^{*}
    =
    \arg\min_{\boldsymbol{\gamma}}
    \left\|
        \mathbf{A}_{\mathrm{target}}
        -
        \mathbf{A}_q
        \operatorname{Diag}(\boldsymbol{\gamma})
    \right\|_F^2 ,
    \label{eq:app_gamma_lstsq}
\end{equation}
where $\mathbf{A}_q$ is the SVID-projected discrete factor and $\boldsymbol{\gamma}$ is a column-wise correction vector along the intermediate dimension.

Let $\mathbf{a}_{t,k}$ and $\mathbf{a}_{q,k}$ denote the $k$-th columns of $\mathbf{A}_{\mathrm{target}}$ and $\mathbf{A}_q$, respectively. Since right multiplication by $\operatorname{Diag}(\boldsymbol{\gamma})$ independently rescales each column of $\mathbf{A}_q$, the objective in Eq.~\ref{eq:app_gamma_lstsq} can be decomposed into independent scalar problems:
\begin{equation}
\begin{aligned}
    &
    \left\|
        \mathbf{A}_{\mathrm{target}}
        -
        \mathbf{A}_q
        \operatorname{Diag}(\boldsymbol{\gamma})
    \right\|_F^2
    \\
    &=
    \sum_{k=1}^{r}
    \left\|
        \mathbf{a}_{t,k}
        -
        \gamma_k
        \mathbf{a}_{q,k}
    \right\|_2^2 .
\end{aligned}
\end{equation}
Thus, each $\gamma_k$ can be optimized separately:
\begin{equation}
    \gamma_k^{*}
    =
    \arg\min_{\gamma_k}
    \left\|
        \mathbf{a}_{t,k}
        -
        \gamma_k
        \mathbf{a}_{q,k}
    \right\|_2^2 .
    \label{eq:app_gamma_scalar}
\end{equation}

Define the scalar objective
\begin{equation}
    f_k(\gamma_k)
    =
    \left\|
        \mathbf{a}_{t,k}
        -
        \gamma_k
        \mathbf{a}_{q,k}
    \right\|_2^2 .
\end{equation}
Expanding it gives
\begin{equation}
\begin{aligned}
    f_k(\gamma_k)
    &=
    \mathbf{a}_{t,k}^{\top}
    \mathbf{a}_{t,k}
    -
    2\gamma_k
    \mathbf{a}_{q,k}^{\top}
    \mathbf{a}_{t,k}
    +
    \gamma_k^2
    \mathbf{a}_{q,k}^{\top}
    \mathbf{a}_{q,k}.
\end{aligned}
\end{equation}
This is a quadratic function in $\gamma_k$. Its second derivative is
\begin{equation}
    \frac{\partial^2 f_k}{\partial \gamma_k^2}
    =
    2
    \mathbf{a}_{q,k}^{\top}
    \mathbf{a}_{q,k}
    =
    2
    \left\|
        \mathbf{a}_{q,k}
    \right\|_2^2
    \ge 0.
\end{equation}
Therefore, $f_k(\gamma_k)$ is convex. Moreover, for the SVID-projected binary-scaled factor, $\mathbf{a}_{q,k}$ is nonzero in normal cases, so $\left\|\mathbf{a}_{q,k}\right\|_2^2>0$ and the objective is strictly convex. Hence, the minimizer exists and is unique.

Taking the first derivative and setting it to zero yields
\begin{equation}
\begin{aligned}
    \frac{\partial f_k}{\partial \gamma_k}
    &=
    -2
    \mathbf{a}_{q,k}^{\top}
    \mathbf{a}_{t,k}
    +
    2\gamma_k
    \mathbf{a}_{q,k}^{\top}
    \mathbf{a}_{q,k}
    =
    0.
\end{aligned}
\end{equation}
Solving for $\gamma_k$ gives
\begin{equation}
    \gamma_k^{*}
    =
    \frac{
        \mathbf{a}_{q,k}^{\top}
        \mathbf{a}_{t,k}
    }{
        \mathbf{a}_{q,k}^{\top}
        \mathbf{a}_{q,k}
    }.
\end{equation}
Stacking all coordinates yields the vectorized closed-form solution
\begin{equation}
    \boldsymbol{\gamma}^{*}
    =
    \frac{
        \operatorname{diag}
        \left(
            \mathbf{A}_q^{\top}
            \mathbf{A}_{\mathrm{target}}
        \right)
    }{
        \operatorname{diag}
        \left(
            \mathbf{A}_q^{\top}
            \mathbf{A}_q
        \right)
    } .
    \label{eq:app_gamma_solution}
\end{equation}
In practice, a small constant $\epsilon$ can be added to the denominator for numerical stability:
\begin{equation}
    \boldsymbol{\gamma}^{*}
    =
    \frac{
        \operatorname{diag}
        \left(
            \mathbf{A}_q^{\top}
            \mathbf{A}_{\mathrm{target}}
        \right)
    }{
        \operatorname{diag}
        \left(
            \mathbf{A}_q^{\top}
            \mathbf{A}_q
        \right)
        +
        \epsilon
    } .
\end{equation}

Finally, the obtained correction vector is folded into the intermediate scaling vector:
\begin{equation}
    \mathbf{m}_1
    \leftarrow
    \mathbf{m}_1
    \odot
    \boldsymbol{\gamma}^{*}.
\end{equation}
This operation keeps the binary sign matrix $\mathbf{A}_{\pm1}$ unchanged and introduces no additional dense parameters. The compensated discrete factor remains within the double-factor binary parameterization, since
\begin{equation}
\begin{aligned}
    \mathbf{A}_q
    \operatorname{Diag}(\boldsymbol{\gamma}^{*})
    &=
    \left(
        \mathbf{a}
        \odot
        \mathbf{A}_{\pm1}
        \odot
        \mathbf{m}_1^{\top}
    \right)
    \operatorname{Diag}(\boldsymbol{\gamma}^{*})
    \\
    &=
    \mathbf{a}
    \odot
    \mathbf{A}_{\pm1}
    \odot
    \left(
        \mathbf{m}_1
        \odot
        \boldsymbol{\gamma}^{*}
    \right)^{\top}.
\end{aligned}
\end{equation}
Therefore, the continuous null-space-compensated target can be approximated by recalibrating only the existing scale vector, preserving the deployable binary structure.

The derivation for the right-factor update is symmetric. Given the compensated target $\mathbf{B}_{\mathrm{target}}$ and the SVID-projected factor $\mathbf{B}_q$, we solve a row-wise scaling problem:
\begin{equation}
    \boldsymbol{\eta}^{*}
    =
    \arg\min_{\boldsymbol{\eta}}
    \left\|
        \mathbf{B}_{\mathrm{target}}
        -
        \operatorname{Diag}(\boldsymbol{\eta})
        \mathbf{B}_q
    \right\|_F^2 .
\end{equation}
By the same least-squares argument, the closed-form solution is
\begin{equation}
    \boldsymbol{\eta}^{*}
    =
    \frac{
        \operatorname{diag}
        \left(
            \mathbf{B}_{\mathrm{target}}
            \mathbf{B}_q^{\top}
        \right)
    }{
        \operatorname{diag}
        \left(
            \mathbf{B}_q
            \mathbf{B}_q^{\top}
        \right)
    } ,
\end{equation}
or, with numerical stabilization,
\begin{equation}
    \boldsymbol{\eta}^{*}
    =
    \frac{
        \operatorname{diag}
        \left(
            \mathbf{B}_{\mathrm{target}}
            \mathbf{B}_q^{\top}
        \right)
    }{
        \operatorname{diag}
        \left(
            \mathbf{B}_q
            \mathbf{B}_q^{\top}
        \right)
        +
        \epsilon
    } .
\end{equation}
This correction is folded into the corresponding intermediate scaling vector of the right factor.

\subsubsection{Interaction-aware Property of Shapley Allocation}
\label{app:shapley_interaction}

We explain why the Shapley-based structural-capacity game in HiSA can capture cross-layer interactions. The key is that Shapley values do not evaluate a layer in isolation. Instead, they measure the expected marginal effect of a layer under different coalition contexts.

Let $T=\{1,\dots,L\}$ denote the set of Transformer layers, and let $v(S)$ be the value of a structural state where layers in $S$ retain the original intermediate dimension while the remaining layers are degraded to the proxy dimension. For layer $l$, its Shapley value is
\begin{equation}
\phi_l
=
\sum_{S\subseteq T\setminus\{l\}}
\frac{|S|!(L-|S|-1)!}{L!}
\left[
v(S\cup\{l\})-v(S)
\right].
\label{eq:app_shapley}
\end{equation}
The marginal term $v(S\cup\{l\})-v(S)$ evaluates the contribution of layer $l$ under a specific context $S$. Different subsets $S$ correspond to different compression states of the remaining layers. Therefore, the same layer is evaluated under many upstream and downstream compression conditions, rather than under a fixed isolated setting.

If compression errors were independent and additive across layers, the marginal effect of layer $l$ would be context-invariant:
\begin{equation}
v(S\cup\{l\})-v(S)
=
\Delta_l,
\quad
\forall S\subseteq T\setminus\{l\}.
\end{equation}
In this special case, isolated local sensitivity would be sufficient, and the Shapley value would reduce to the same contribution $\Delta_l$.

However, deep Transformers generally violate this additive assumption. Compressing one layer changes the hidden-state distribution received by subsequent layers, and downstream layers may amplify, suppress, or compensate the earlier error. As a result, the marginal effect of layer $l$ depends on the coalition context:
\begin{equation}
v(S\cup\{l\})-v(S)
\neq
v(S'\cup\{l\})-v(S'),
\quad
S\neq S'.
\end{equation}
This context dependence is exactly the cross-layer interaction that isolated local reconstruction error cannot capture.

The permutation view of Shapley values makes this interaction-aware property clearer. For a random permutation $\pi$ of layers, let $\mathrm{Pre}_{\pi}(l)$ denote the set of layers appearing before $l$ in $\pi$. Then,
\begin{equation}
\phi_l
=
\mathbb{E}_{\pi}
\left[
v(\mathrm{Pre}_{\pi}(l)\cup\{l\})
-
v(\mathrm{Pre}_{\pi}(l))
\right].
\label{eq:app_perm_shapley}
\end{equation}
Thus, each sampled permutation evaluates layer $l$ under a different structural context.

In HiSA, we approximate this expectation by progressively degrading layers along sampled permutations and measuring the induced increase in end-to-end NLL:
\begin{equation}
\Delta v_l^{(m)}
=
v_{\mathrm{NLL}}(S_l^{(m)}\setminus\{l\})
-
v_{\mathrm{NLL}}(S_l^{(m)}),
\end{equation}
where $S_l^{(m)}$ denotes the set of layers that still retain the original dimension before layer $l$ is degraded in the $m$-th permutation. Averaging over $M$ sampled permutations gives
\begin{equation}
\Phi_l
=
\frac{1}{M}
\sum_{m=1}^{M}
\Delta v_l^{(m)}.
\end{equation}

This estimator differs from isolated local sensitivity in two aspects. First, the value function is the full-model NLL, so the measured effect includes downstream propagation and prediction-level changes. Second, each marginal contribution is evaluated under a different coalition of compressed and uncompressed layers, so the estimate reflects how layer $l$ interacts with other layers under different structural states. Therefore, the layer-level Shapley score in HiSA is interaction-aware and is more suitable for allocating global structural budgets than uniform dimensions or isolated reconstruction errors.

\subsubsection{Dual Distortion Decomposition for Block-Level Allocation}
\label{app:dual_distortion}

We provide the motivation for using both directional and magnitude distortions in block-level HiSA. The key observation is that a single MSE score entangles two different types of representation errors: angular drift and activation-energy shift. Let $\mathbf{h}$ and $\hat{\mathbf{h}}$ denote the reference and compressed hidden states of a token, respectively. Writing
\begin{equation}
    \mathbf{h}
    =
    \|\mathbf{h}\|_2 \mathbf{u},
    \qquad
    \hat{\mathbf{h}}
    =
    \|\hat{\mathbf{h}}\|_2 \hat{\mathbf{u}},
\end{equation}
where $\|\mathbf{u}\|_2=\|\hat{\mathbf{u}}\|_2=1$, the squared error can be decomposed as
\begin{equation}
\begin{aligned}
    \|\mathbf{h}-\hat{\mathbf{h}}\|_2^2
    &=
    \|\mathbf{h}\|_2^2
    +
    \|\hat{\mathbf{h}}\|_2^2
    -
    2
    \langle
        \mathbf{h},
        \hat{\mathbf{h}}
    \rangle
    \\
    &=
    \left(
        \|\mathbf{h}\|_2
        -
        \|\hat{\mathbf{h}}\|_2
    \right)^2
    \\
    & \qquad
    +2
    \|\mathbf{h}\|_2
    \|\hat{\mathbf{h}}\|_2
    \left(
        1-
        \langle
            \mathbf{u},
            \hat{\mathbf{u}}
        \rangle
    \right).
\end{aligned}
\end{equation}
The first term measures the mismatch in representation magnitude, while the second term measures the angular deviation between the two normalized directions. Therefore, MSE is a coupled measure of magnitude and directional errors, and a single scalar may hide which failure mode dominates.

This distinction is important for Transformer blocks. In self-attention, the attention logits are computed from query--key similarities:
\begin{equation}
    \mathbf{S}
    =
    \frac{
        \mathbf{Q}\mathbf{K}^{\top}
    }{
        \sqrt{d}
    }.
\end{equation}
For two token vectors $\mathbf{q}$ and $\mathbf{k}$,
\begin{equation}
    \mathbf{q}^{\top}\mathbf{k}
    =
    \|\mathbf{q}\|_2
    \|\mathbf{k}\|_2
    \cos(\theta_{\mathbf{q},\mathbf{k}}).
\end{equation}
Thus, angular changes in query or key representations can directly alter token-matching patterns and attention distributions, even when vector norms are not severely changed. This makes Q/K projections particularly sensitive to directional distortion.

By contrast, value projections, output projections, and MLP branches mainly inject transformed activations into the residual stream. Their compression errors often manifest as changes in activation energy, channel scale, or residual update strength. For a residual update
\begin{equation}
    \mathbf{h}_{l+1}
    =
    \mathbf{h}_{l}
    +
    F_l(\mathbf{h}_{l}),
\end{equation}
a magnitude mismatch in the compressed branch $\hat F_l$ changes the scale of the injected update:
\begin{equation}
    \|\hat F_l(\mathbf{h}_{l})\|_2
    \neq
    \|F_l(\mathbf{h}_{l})\|_2.
\end{equation}
Such scale shifts can accumulate across layers and disturb downstream normalization, residual balance, and activation dynamic range. This makes magnitude distortion an important signal for V/O projections and MLP modules.

However, these module tendencies should not be hard-coded as fixed priors, since their sensitivity can vary across layers, model families, and compression budgets. HiSA therefore measures both distortions directly from layer outputs. The directional error
\begin{equation}
r_{\mathrm{dir}}
=
\mathbb{E}_{b,t}
\left[
1-
\frac{
\langle
h_{b,t}^{\mathrm{ref}},
\hat h_{b,t}
\rangle
}{
\|h_{b,t}^{\mathrm{ref}}\|_2
\|\hat h_{b,t}\|_2
+
\epsilon
}
\right]
\end{equation}
captures angular drift of token representations, while the magnitude error
\begin{equation}
r_{\mathrm{mag}}
=
\mathbb{E}_{b,t}
\left[
\frac{
\left|
\|\hat h_{b,t}\|_2
-
\|h_{b,t}^{\mathrm{ref}}\|_2
\right|
}{
\|h_{b,t}^{\mathrm{ref}}\|_2
+
\epsilon
}
\right]
\end{equation}
captures relative activation-energy shift. By estimating Shapley marginal contributions separately under these two metrics, HiSA identifies whether a submodule is vulnerable through directional mismatch, magnitude mismatch, or both. The subsequent Soft-OR fusion ensures that a module with strong sensitivity in either failure mode can still receive sufficient structural capacity.

\subsubsection{Motivation for Size-Aware Sensitivity Density}
\label{app:size_density}

We explain why HiSA converts the fused vulnerability score into a size-aware sensitivity density before allocating the module-level budget. For brevity, let $v_i$ denote $v_{\mathrm{base}}^{(i)}$, and let $|\mathbf{W}_i|$ denote the parameter count of submodule $i$. If the allocation directly uses $v_i$ as the module weight, then the total budget received by a module is affected by both its vulnerability and its size. Specifically, under a layer budget $B_l$, a generic allocation weight $\omega_i$ gives
\begin{equation}
    b_i
    =
    B_l
    \cdot
    \frac{
        P_l\omega_i
    }{
        \sum_{j\in\mathcal{N}_l}
        |\mathbf{W}_j|\omega_j
    },
    \qquad
    P_l
    =
    \sum_{j\in\mathcal{N}_l}
    |\mathbf{W}_j|.
\end{equation}
The actual number of allocated bits for module $i$ is therefore
\begin{equation}
    |\mathbf{W}_i|b_i
    =
    B_l P_l
    \cdot
    \frac{
        |\mathbf{W}_i|\omega_i
    }{
        \sum_{j\in\mathcal{N}_l}
        |\mathbf{W}_j|\omega_j
    }.
\end{equation}
If we set $\omega_i=v_i$, then the total allocated bits become proportional to $|\mathbf{W}_i|v_i$. As a result, a large module can dominate the layer budget even when its vulnerability per parameter is not high.

To reduce this size-induced bias, HiSA first converts vulnerability into a sensitivity density:
\begin{equation}
    d_i
    =
    v_i/|\mathbf{W}_i|.
\end{equation}
This quantity measures vulnerability per parameter, rather than the aggregate vulnerability of the whole module. However, using density alone may over-emphasize very small modules, whose Shapley estimates can be noisier due to fewer parameters and smaller output contribution. Therefore, HiSA applies a smoothing exponent:
\begin{equation}
    \omega_i
    =
    (d_i+\epsilon)^{\alpha}.
\end{equation}
Substituting this into the total allocated bits gives
\begin{equation}
    |\mathbf{W}_i|b_i
    \propto
    |\mathbf{W}_i|
    \left(
        v_i/|\mathbf{W}_i|+\epsilon
    \right)^{\alpha}.
\end{equation}
Ignoring the small $\epsilon$ for clarity, this becomes
\begin{equation}
    |\mathbf{W}_i|b_i
    \propto
    |\mathbf{W}_i|^{1-\alpha} v_i^{\alpha}.
\end{equation}
Thus, $\alpha$ controls the trade-off between size-proportional allocation and vulnerability-density allocation. When $\alpha=0$, all modules receive the same average bit-width and total allocated bits are proportional to module size. When $\alpha=1$, the total allocated bits are mainly determined by the fused vulnerability $v_i$, reducing the dominance of large modules. Intermediate values of $\alpha$ interpolate between these two extremes.

As an illustrative example, consider a Qwen3-8B-style decoder layer. With hidden size $4096$ and intermediate size $22016$, an attention projection such as $\mathbf{W}_q$ has approximately
\begin{equation}
    4096\times4096 \approx 16.8\mathrm{M}
\end{equation}
parameters, while an MLP projection such as $\mathbf{W}_{\mathrm{up}}$ has approximately
\begin{equation}
    4096\times22016 \approx 90.2\mathrm{M}
\end{equation}
parameters. The MLP projection is therefore about $5.4\times$ larger than the attention projection. Suppose the fused vulnerabilities are $v_{\mathrm{mlp}}=0.6$ and $v_{\mathrm{attn}}=0.3$. If raw vulnerability is used as the allocation weight, the ratio of allocated bits is
\begin{equation}
    \frac{
        |\mathbf{W}_{\mathrm{mlp}}|v_{\mathrm{mlp}}
    }{
        |\mathbf{W}_{\mathrm{attn}}|v_{\mathrm{attn}}
    }
    \approx
    5.4\times2
    =
    10.8.
\end{equation}
Thus, the larger MLP module receives more than ten times the total budget, even though its vulnerability score is only twice as large. By contrast, with density smoothing and $\alpha=1$, the size factor is approximately canceled in the total-bit allocation:
\begin{equation}
    \frac{
        |\mathbf{W}_{\mathrm{mlp}}|d_{\mathrm{mlp}}
    }{
        |\mathbf{W}_{\mathrm{attn}}|d_{\mathrm{attn}}
    }
    =
    \frac{
        v_{\mathrm{mlp}}
    }{
        v_{\mathrm{attn}}
    }
    =
    2.
\end{equation}
With $0<\alpha<1$, HiSA obtains a smoother interpolation, reducing large-module dominance while avoiding excessive sensitivity to noisy estimates from small modules. This motivates the size-aware density and smoothing design used in Eq.~\ref{eq:hisa_module_budget}, where $P_l$ and $\omega_i$ are expanded inline in the main text for compactness.

\subsection{Effective Bit-Width of Binary PTQ Methods}
\label{app:bpw_binary_ptq}

Although binary PTQ methods represent most quantized weights with 1-bit signs, their deployable storage is not determined by binary weights alone. To reconstruct the quantized matrix, one also needs to store residual binary components, BF16 scaling factors and means, salient-column indicators, group bitmaps, and, for structured sparsity, sparsity indices. Therefore, we report the effective bits per weight (BPW) by counting all bits required for weight reconstruction.

For a weight matrix $W\in\mathbb{R}^{n\times m}$, let $n$ and $m$ denote the number of rows and columns, respectively. Let $c$ be the number of salient columns selected in $W$, $k$ be the block size, and $q=\lceil m/k\rceil$ be the number of column blocks per row. Following common implementations, all reconstruction coefficients, including scaling factors and means, are stored in BF16. If the quantized representation requires $M$ bits in total, then
\begin{equation}
\mathrm{BPW}=\frac{M}{nm}.
\label{eq:app_bpw_def}
\end{equation}

\paragraph{BiLLM~\citep{huang2024billmpushinglimitposttraining}.}
BiLLM separates the weight matrix into salient and non-salient columns. For the salient columns, BiLLM adopts second-order residual binarization, which stores two binary tensors for the original and residual components. Thus, the binary storage of the salient part is $2nc$. In addition, each column block stores three BF16 row-wise reconstruction coefficients, corresponding to $\alpha_1$, $\alpha_2$, and the merged mean $\mu$, which costs $3\times16nq=48nq$ bits. For the non-salient columns, BiLLM uses first-order binarization with two distribution groups, which stores $n(m-c)$ binary signs and two sets of BF16 scale/mean coefficients, costing $2\times2\times16nq=64nq$ bits. Finally, BiLLM stores a group bitmap over the whole matrix and a salient-column bitmap, which cost $nm$ and $m$ bits, respectively. Therefore,
\begin{equation}
\begin{aligned}
M_{\mathrm{BiLLM}}
&~=(2nc+48nq)+\big(n(m-c)+64nq\big)\\
&~+nm+m=2nm+nc+m+112nq .
\end{aligned}
\label{eq:app_billm_memory}
\end{equation}
Dividing Eq.~\ref{eq:app_billm_memory} by $nm$ gives
\begin{equation}
\mathrm{BPW}_{\mathrm{BiLLM}}
=
2+\frac{c}{m}+\frac{112q}{m}+\frac{1}{n}.
\label{eq:app_billm_bpw}
\end{equation}

\paragraph{STBLLM$_{4:8}$~\citep{dong2024stbllm}.}
STBLLM further introduces structured $N\!:\!M$ sparsity on top of BiLLM-style binary PTQ. We only consider the 4:8 setting used in our comparison. The salient part still follows residual binarization, so it costs $2nc+48nq$ bits, as in BiLLM. For the non-salient part, 4:8 sparsity retains only half of the binary weights. Therefore, the retained binary signs contribute $\frac{1}{2}n(m-c)$ bits. STBLLM also divides non-salient weights into sparse, intermediate, and dense regions using trisection search, and each retained value needs a 2-bit group identifier; this contributes $\frac{1}{2}\cdot2nm$ bits. Moreover, each 4:8 block needs to store which four positions are retained among eight positions. Since $\lceil\log_2 {8\choose4}\rceil=7$, the sparsity-index overhead is $\frac{7}{8}n(m-c)$ bits. The three non-salient groups require three sets of BF16 scale/mean coefficients, costing $3\times2\times16nq=96nq$ bits. Together with the salient-column bitmap, the total storage is
\begin{equation}
\resizebox{\columnwidth}{!}{$
    \displaystyle
    \begin{aligned}
    M_{\mathrm{STBLLM}_{4:8}}
    =&~(2nc+48nq)+\frac{1}{2}\big[n(m-c)+2nm\big]  \\
    &+\frac{7}{8}n(m-c)+96nq+m  \\
    =&~\frac{19}{8}nm+\frac{5}{8}nc+144nq+m .
    \end{aligned}
$}
\label{eq:app_stbllm48_memory}
\end{equation}
Thus, the effective BPW is
\begin{equation}
\mathrm{BPW}_{\mathrm{STBLLM}_{4:8}}
=
\frac{19}{8}+\frac{5c}{8m}+\frac{144q}{m}+\frac{1}{n}.
\label{eq:app_stbllm48_bpw}
\end{equation}

\paragraph{ARB-LLM$_{\mathrm{RC}}$~\citep{li2024arbllmalternatingrefinedbinarizations}.}
ARB-LLM$_{\mathrm{RC}}$ improves BiLLM by alternatingly refining binarization parameters and introducing row-column-wise scaling to better preserve column-wise deviations. For salient columns, it stores two binary residual components, giving $2nc$ bits. Different from purely row-wise BiLLM, ARB-LLM$_{\mathrm{RC}}$ stores both row-wise and column-wise BF16 coefficients for the residual representation. Specifically, the salient part requires $(2nq+2c)\times16$ bits, where $2nq$ corresponds to row-wise coefficients over two residual components and $2c$ corresponds to column-wise coefficients for the salient columns. For the non-salient part, ARB-LLM$_{\mathrm{RC}}$ stores $n(m-c)$ binary signs and two groups of row-column BF16 coefficients, yielding $(nq+m-c)\times16\times2$ bits. It also requires a group bitmap and a salient-column bitmap, costing $nm$ and $m$ bits. Therefore,
\begin{equation}
\begin{aligned}
M_{\mathrm{ARB}_{\mathrm{RC}}}
=&~\big(2nc+16(2nq+2c)\big)\\
+&\big(n(m-c)+32(nq+m-c)\big) \\
+&nm+m=2nm+nc+33m+64nq .
\end{aligned}
\label{eq:app_arb_rc_memory}
\end{equation}
The corresponding BPW is
\begin{equation}
\mathrm{BPW}_{\mathrm{ARB}_{\mathrm{RC}}}
=
2+\frac{c}{m}+\frac{64q}{m}+\frac{33}{n}.
\label{eq:app_arb_rc_bpw}
\end{equation}

\paragraph{HBLLM$_{\mathrm{row}}$~\citep{chen2025hbllm}.}
HBLLM$_{\mathrm{row}}$ extends BiLLM with Haar-domain row-wise quantization and frequency-aware grouping. In the non-salient branch, HBLLM$_{\mathrm{row}}$ uses FillAvg to fill salient positions before applying row-wise Haar quantization. As a result, the non-salient Haar branch stores a dense binary representation over $nm$ positions rather than only $n(m-c)$ positions. Its BF16 coefficient overhead is $3\times16nq\times2=96nq$ bits because the row-wise Haar branch uses two frequency-aware groups with three BF16 coefficients per group. For the salient branch, HBLLM$_{\mathrm{row}}$ stores $nc$ binary signs and two groups of BF16 coefficients, costing $2\times2\times16nq=64nq$ bits. In addition, the FillAvg-based row branch requires a group bitmap covering both the original matrix and the filled salient positions, giving $n(m+c)$ bits, while the salient-column bitmap costs $m$ bits. Hence,
\begin{equation}
\begin{aligned}
M_{\mathrm{HBLLM}_{\mathrm{row}}}
=&~(nm+96nq)+(nc+64nq)\\
&~+n(m+c)+m  \\
=&~2nm+2nc+m+160nq .
\end{aligned}
\label{eq:app_hbllm_row_memory}
\end{equation}
Therefore,
\begin{equation}
\mathrm{BPW}_{\mathrm{HBLLM}_{\mathrm{row}}}
=
2+\frac{2c}{m}+\frac{160q}{m}+\frac{1}{n}.
\label{eq:app_hbllm_row_bpw}
\end{equation}

\paragraph{AF1 (ours).}
For a target linear matrix $W_i\in\mathbb{R}^{d_{\mathrm{out},i}\times d_{\mathrm{in},i}}$, AF1 stores two binary sign factors $A_i\in\{-1,+1\}^{d_{\mathrm{out},i}\times r_i}$ and $B_i\in\{-1,+1\}^{r_i\times d_{\mathrm{in},i}}$, together with three BF16 scale vectors $a_i\in\mathbb{R}^{d_{\mathrm{out},i}}$, $m_i\in\mathbb{R}^{r_i}$, and $b_i\in\mathbb{R}^{d_{\mathrm{in},i}}$. Therefore, the deployable storage of this matrix is
\begin{equation}
M_i^{\mathrm{AF1}}
=
r_i(d_{\mathrm{out},i}+d_{\mathrm{in},i})
+
16(d_{\mathrm{out},i}+r_i+d_{\mathrm{in},i}).
\label{eq:app_af1_memory}
\end{equation}
The corresponding model-level effective BPW is
\begin{equation}
\mathrm{BPW}_{\mathrm{AF1}}
=
\frac{
\sum_i M_i^{\mathrm{AF1}}
}{
\sum_i d_{\mathrm{out},i}d_{\mathrm{in},i}
}.
\label{eq:app_af1_bpw}
\end{equation}
In practice, HiSA assigns the module-wise budgets and AF1 chooses the intermediate dimensions $\{r_i\}$ under the model-level constraint in Eq.~\ref{eq:app_af1_bpw}. The calibration statistics, ADMM variables, null-space projectors, and teacher logits are used only during quantization and are not stored for deployment. Thus, unlike prior binary PTQ baselines, AF1 does not require salient-column masks, group bitmaps, residual binary tensors, or structured-sparsity indices at inference time.

These derivations show that existing binary PTQ baselines are not purely 1-bit in effective storage. Their BPW is increased by second-order residual binary tensors, BF16 reconstruction coefficients, saliency masks, group bitmaps, and structured-sparsity metadata. In contrast, AF1 explicitly controls the deployable storage through its binary-factor dimensions and lightweight scale vectors, explaining why the nominal binary width and the actual deployable BPW can differ substantially for prior methods but remain budget-controlled in AF1.

\subsection{More Experimental Results}
\label{sec:more ablation}

\subsubsection{More Detailed Results}
\label{app:detailed main results}
In this appendix, we provide the full expanded results corresponding to Table~\ref{tab:main_results} in the main paper. 
Table~\ref{tab:app:zero_shot_full} reports the task-level zero-shot accuracy on ARC-Easy, ARC-Challenge, HellaSwag, LAMBADA-openai, LAMBADA-standard, PIQA, and WinoGrande for each evaluated model. 
The expanded breakdown shows that the advantage of AF1 is not caused by isolated improvements on a single benchmark, but is consistently reflected across different task categories, including commonsense reasoning, language modeling, and cloze-style completion. 
Under the 2.50-bit setting, AF1 achieves the best averaged zero-shot accuracy for all evaluated model families and obtains the strongest or near-strongest performance on most individual tasks, while also yielding the lowest WikiText2 perplexity in Table~\ref{tab:main_results}. 
Notably, even at the strict 1.00-bit setting, AF1 remains competitive with or superior to several higher-bit baselines, demonstrating that the proposed binary factorization and hierarchical budget allocation preserve task-level generalization under an extremely constrained weight budget. 
These detailed results further support the main conclusion that AF1 improves both average performance and per-task robustness over existing binary PTQ methods.

\begin{table*}[h!]
\centering
\resizebox{0.84\textwidth}{!}{
\tablestyle{3pt}{1.1}
\begin{tabular}{llccccccccc}
\toprule
\textbf{Model} & \textbf{Method} & \textbf{\#Bits(W)}
& \textbf{AE}$\uparrow$ & \textbf{AC}$\uparrow$ & \textbf{HS}$\uparrow$
& \textbf{LO}$\uparrow$ & \textbf{LS}$\uparrow$
& \textbf{PQ}$\uparrow$ & \textbf{WG}$\uparrow$
& \textbf{Avg.}$^7$ $\uparrow$ \\
\midrule

\multirow{8}{*}{\makecell[l]{LLaMA-3-8B}} & BF16 & 16 & 77.90 & 52.82 & 79.07 & 75.63 & 68.58 & 80.63 & 72.93 & 72.51 \\
\cdashline{2-11}
 & RTN & 1.00 & 24.71 & 25.94 & 26.63 & 0.00 & 0.00 & 51.85 & 51.62 & 25.82 \\
 & BiLLM & 2.88 & 36.78 & 22.18 & 34.81 & 28.76 & 13.00 & 58.49 & 53.67 & 35.38 \\
 & STBLLM$_{\text{(4:8)}}$ & 3.50 & 31.10 & 21.76 & 31.29 & 11.84 & 5.88 & 54.52 & 49.64 & 29.43 \\
 & ARB-LLM$_{\text{RC}}$ & 2.51 & 48.78 & 28.33 & 43.55 & 50.65 & 29.83 & 65.61 & 56.75 & 46.21 \\
 & HBLLM$_{\text{row}}$ & 3.25 & \underline{54.12} & \underline{32.68} & \underline{60.22} & \underline{58.51} & \underline{46.40} & \underline{70.35} & \underline{65.19} & \underline{55.35} \\
\rowcolor{gray!5}  & \textbf{AF1} & 2.50 & \textbf{72.43} & \textbf{44.62} & \textbf{73.81} & \textbf{70.56} & \textbf{62.51} & \textbf{78.29} & \textbf{69.22} & \textbf{67.35} \\
\rowcolor{gray!5}  & \textbf{AF1} & 1.00 & 44.51 & 26.49 & 42.01 & 47.35 & 36.48 & 65.00 & 56.54 & 45.48 \\
\midrule

\multirow{8}{*}{\makecell[l]{LLaMA-3-70B}} & BF16 & 16 & 86.07 & 64.33 & 84.92 & 79.31 & 73.96 & 84.60 & 80.35 & 79.08 \\
\cdashline{2-11}
 & RTN & 1.00 & 25.31 & 24.33 & 25.59 & 0.00 & 0.00 & 50.61 & 48.29 & 24.88 \\
 & BiLLM & 2.88 & 25.34 & 20.73 & 33.18 & 9.41 & 5.44 & 52.39 & 51.78 & 28.32 \\
 & STBLLM$_{\text{(4:8)}}$ & 3.50 & 25.67 & 25.94 & 27.23 & 1.84 & 4.29 & 51.03 & 49.88 & 26.55 \\
 & ARB-LLM$_{\text{RC}}$ & 2.51 & \underline{61.71} & \underline{42.48} & 67.44 & 65.68 & 53.54 & \underline{69.97} & 64.53 & 60.76 \\
 & HBLLM$_{\text{row}}$ & 3.25 & 24.83 & 27.05 & \underline{74.70} & \underline{70.66} & 50.96 & 50.60 & \underline{73.56} & 53.19 \\
\rowcolor{gray!5}  & \textbf{AF1} & 2.50 & \textbf{71.89} & \textbf{48.04} & \textbf{83.43} & \textbf{76.93} & \textbf{70.74} & \textbf{77.20} & \textbf{77.66} & \textbf{72.27} \\
\rowcolor{gray!5}  & \textbf{AF1} & 1.00 & 55.56 & 36.69 & 73.92 & 68.60 & \underline{61.36} & 68.39 & 68.11 & \underline{61.80} \\
\midrule

\multirow{8}{*}{\makecell[l]{Qwen3-8B}} & BF16 & 16 & 80.93 & 56.74 & 74.98 & 64.18 & 61.11 & 77.37 & 68.35 & 69.09 \\
\cdashline{2-11}
 & RTN & 1.00 & 25.84 & 25.17 & 26.56 & 0.00 & 0.00 & 49.95 & 49.41 & 25.28 \\
 & BiLLM & 2.88 & 34.97 & 23.81 & 37.69 & 18.75 & 12.71 & 56.80 & 51.14 & 33.70 \\
 & STBLLM$_{\text{(4:8)}}$ & 3.50 & 31.78 & 22.61 & 29.89 & 5.72 & 2.99 & 53.81 & 49.33 & 28.02 \\
 & ARB-LLM$_{\text{RC}}$ & 2.51 & 59.09 & 36.35 & 49.84 & 41.86 & 34.29 & 68.72 & 58.80 & 49.85 \\
 & HBLLM$_{\text{row}}$ & 3.25 & 58.38 & 36.18 & \underline{60.51} & 50.32 & 44.91 & \underline{71.82} & \underline{61.17} & 54.76 \\
\rowcolor{gray!5}  & \textbf{AF1} & 2.50 & \textbf{78.16} & \textbf{55.03} & \textbf{72.91} & \textbf{61.42} & \textbf{58.98} & \textbf{77.31} & \textbf{71.82} & \textbf{67.95} \\
\rowcolor{gray!5}  & \textbf{AF1} & 1.00 & \underline{62.07} & \underline{37.91} & 55.76 & \underline{50.95} & \underline{45.24} & 70.87 & 60.97 & \underline{54.82} \\
\midrule

\multirow{8}{*}{\makecell[l]{Qwen3-14B}} & BF16 & 16 & 83.08 & 60.49 & 78.82 & 67.84 & 64.47 & 79.76 & 72.85 & 72.47 \\
\cdashline{2-11}
 & RTN & 1.00 & 24.62 & 26.88 & 26.14 & 0.00 & 0.00 & 51.41 & 49.80 & 25.55 \\
 & BiLLM & 2.88 & 56.57 & 34.56 & 55.70 & 48.67 & 43.90 & 70.62 & 64.72 & 53.53 \\
 & STBLLM$_{\text{(4:8)}}$ & 3.50 & 47.52 & 27.47 & 44.33 & 24.72 & 22.86 & 63.06 & 56.20 & 40.88 \\
 & ARB-LLM$_{\text{RC}}$ & 2.51 & \underline{72.43} & 44.28 & 60.55 & 61.50 & 51.85 & 73.88 & 66.69 & 61.60 \\
 & HBLLM$_{\text{row}}$ & 3.25 & 69.19 & \underline{46.93} & \underline{69.60} & \underline{63.61} & \underline{56.53} & \underline{74.92} & \underline{68.03} & \underline{64.12} \\
\rowcolor{gray!5}  & \textbf{AF1} & 2.50 & \textbf{78.83} & \textbf{55.72} & \textbf{77.37} & \textbf{66.80} & \textbf{62.58} & \textbf{79.27} & \textbf{71.74} & \textbf{70.33} \\
\rowcolor{gray!5}  & \textbf{AF1} & 1.00 & 67.85 & 41.98 & 64.22 & 57.11 & 49.41 & 72.25 & 64.48 & 59.61 \\
\midrule

\multirow{8}{*}{\makecell[l]{Qwen3-32B}} & BF16 & 16 & 83.21 & 61.09 & 82.60 & 67.24 & 58.04 & 81.99 & 72.77 & 72.42 \\
\cdashline{2-11}
 & RTN & 1.00 & 25.21 & 26.71 & 26.37 & 0.00 & 0.00 & 50.00 & 49.57 & 25.41 \\
 & BiLLM & 2.88 & 63.80 & 41.30 & 67.62 & 62.24 & 53.13 & 73.83 & 66.69 & 61.23 \\
 & STBLLM$_{\text{(4:8)}}$ & 3.50 & 60.65 & 37.63 & 62.42 & 51.21 & 43.02 & 72.58 & 63.54 & 55.86 \\
 & ARB-LLM$_{\text{RC}}$ & 2.51 & \underline{75.97} & \underline{54.35} & 71.91 & \textbf{71.28} & \underline{62.62} & \underline{78.78} & \underline{71.98} & \underline{69.56} \\
 & HBLLM$_{\text{row}}$ & 3.25 & 75.17 & 50.34 & \underline{75.63} & 66.50 & 56.16 & 76.99 & 70.17 & 67.28 \\
\rowcolor{gray!5}  & \textbf{AF1} & 2.50 & \textbf{81.23} & \textbf{59.56} & \textbf{80.28} & \underline{70.54} & \textbf{62.97} & \textbf{80.52} & \textbf{72.30} & \textbf{72.49} \\
\rowcolor{gray!5}  & \textbf{AF1} & 1.00 & 71.72 & 46.25 & 70.37 & 61.61 & 55.13 & 75.35 & 69.38 & 64.26 \\
\midrule

\multirow{8}{*}{\makecell[l]{Gemma-3-4B-it}} & BF16 & 16 & 78.20 & 56.74 & 74.09 & 59.17 & 54.49 & 77.15 & 70.01 & 67.12 \\
\cdashline{2-11}
 & RTN & 1.00 & 25.02 & 26.15 & 25.73 & 0.00 & 0.00 & 50.45 & 49.60 & 25.28 \\
 & BiLLM & 2.88 & 36.57 & 24.91 & 35.29 & 9.33 & 6.64 & 60.01 & 53.59 & 32.33 \\
 & STBLLM$_{\text{(4:8)}}$ & 3.50 & 33.88 & 22.18 & 31.34 & 5.67 & 4.68 & 55.28 & 52.17 & 29.31 \\
 & ARB-LLM$_{\text{RC}}$ & 2.51 & 49.20 & 29.10 & 40.62 & 30.35 & 23.33 & 65.61 & 54.62 & 41.83 \\
 & HBLLM$_{\text{row}}$ & 3.25 & \underline{60.40} & \underline{36.52} & \underline{54.13} & 41.68 & 33.01 & \underline{69.42} & 57.30 & \underline{50.35} \\
\rowcolor{gray!5}  & \textbf{AF1} & 2.50 & \textbf{71.46} & \textbf{45.05} & \textbf{68.47} & \textbf{57.89} & \textbf{51.78} & \textbf{76.06} & \textbf{65.35} & \textbf{62.29} \\
\rowcolor{gray!5}  & \textbf{AF1} & 1.00 & 57.74 & 35.60 & 49.83 & \underline{42.49} & \underline{37.58} & 67.35 & \underline{58.43} & 49.86 \\
\midrule

\multirow{8}{*}{\makecell[l]{Gemma-3-12B-it}} & BF16 & 16 & 76.73 & 60.32 & 81.89 & 67.71 & 65.61 & 78.51 & 74.66 & 72.20 \\
\cdashline{2-11}
 & RTN & 1.00 & 25.33 & 26.01 & 25.88 & 0.00 & 0.00 & 50.23 & 49.66 & 25.30 \\
 & BiLLM & 2.88 & 43.60 & 28.67 & 41.82 & 22.06 & 13.97 & 62.79 & 55.96 & 38.41 \\
 & STBLLM$_{\text{(4:8)}}$ & 3.50 & 34.01 & 24.06 & 33.57 & 13.45 & 8.52 & 54.68 & 51.93 & 31.46 \\
 & ARB-LLM$_{\text{RC}}$ & 2.51 & 46.93 & 29.27 & 45.98 & 50.73 & 34.60 & 64.64 & 57.93 & 47.15 \\
 & HBLLM$_{\text{row}}$ & 3.25 & \underline{67.80} & \underline{43.69} & \underline{67.00} & 43.92 & 37.61 & 74.70 & 67.09 & 57.40 \\
\rowcolor{gray!5}  & \textbf{AF1} & 2.50 & \textbf{80.35} & \textbf{59.30} & \textbf{76.99} & \textbf{64.39} & \textbf{61.28} & \textbf{79.27} & \textbf{71.51} & \textbf{70.44} \\
\rowcolor{gray!5}  & \textbf{AF1} & 1.00 & 66.95 & 41.86 & 57.42 & \underline{54.45} & \underline{47.81} & \underline{77.36} & \underline{67.30} & \underline{59.02} \\
\bottomrule
\end{tabular}}
\vspace{-0.2cm}
\caption{Detailed zero-shot accuracy on ARC-Easy (AE), ARC-Challenge (AC), HellaSwag (HS), LAMBADA-openai (LO), LAMBADA-standard (LS), PIQA (PQ), and WinoGrande (WG). The average score is computed over the seven zero-shot tasks. Bold and underline denote the best and second-best quantized results within each model.}
\label{tab:app:zero_shot_full}
\end{table*}

\subsubsection{Ablation on the Residual-Energy Threshold}
\label{app:ablation_nullspace_threshold}
We further study the sensitivity of NABF to the residual-energy threshold $\eta$ used in the adaptive null-space cutoff. 
This threshold controls the size of the selected trailing subspace: a smaller $\eta$ yields a more conservative null-space estimate, while a larger $\eta$ introduces more low-energy directions for compensation. 
Table~\ref{tab:nullspace_threshold_ablation} reports WikiText2 perplexity on LLaMA-3-8B, Qwen3-14B, and Gemma-3-4B-it when sweeping $\eta$ from $0.01$ to $0.09$. 
Overall, $\eta=0.01$ achieves the lowest average perplexity and is therefore used as the default setting.
\begin{table*}[t]
    \centering
    \tablestyle{3pt}{1.3}
    \begin{tabular}{c|ccc|c}
    \hline
    $\eta$ 
    & LLaMA-3-8B 
    & Qwen3-14B 
    & Gemma-3-4B-it 
    & Avg. \\
    \hline
    0.01 & \textbf{22.18} & \textbf{13.32} & \underline{32.82} & \textbf{22.77} \\
    0.02 & \underline{22.35} & \underline{13.45} & \textbf{32.74} & \underline{22.85} \\
    0.03 & 22.51 & 13.58 & 33.02 & 23.04 \\
    0.04 & 22.43 & 13.72 & 33.11 & 23.09 \\
    0.05 & 22.69 & 13.66 & 33.23 & 23.19 \\
    0.06 & 22.58 & 13.84 & 33.34 & 23.25 \\
    0.07 & 22.83 & 13.79 & 33.50 & 23.37 \\
    0.08 & 22.74 & 14.01 & 33.61 & 23.45 \\
    0.09 & 23.05 & 13.95 & 33.70 & 23.57 \\
    \hline
    \end{tabular}
    \caption{
    Ablation on the residual-energy threshold $\eta$ used in the adaptive null-space cutoff. 
    Results are reported as WikiText2 perplexity on LLaMA-3-8B, Qwen3-14B, and Gemma-3-4B-it. 
    Bold and underline denote the best and second-best results, respectively.
    }
    \label{tab:nullspace_threshold_ablation}
\end{table*}

\subsubsection{Ablation on Allocation-Sensitivity Metrics}
\label{app:ablation_hisa_metric}

\begin{table*}[t!]
\centering
\tablestyle{3pt}{1.2}
\begin{tabular}{c c c c c c c c c c}
\toprule
\multirow{2}{*}{\textbf{Model}} 
& \multirow{2}{*}{\makecell{\textbf{Avg.}\\\textbf{Bits(W)}}}
& \multicolumn{5}{c}{\textbf{General Allocation Baselines (PPL$\downarrow$)}}
& \multicolumn{3}{c}{\textbf{HiSA Variants (PPL$\downarrow$)}} \\
\cmidrule(lr){3-7}
\cmidrule(lr){8-10}
& & \textbf{Uniform} & \textbf{LIM} & \textbf{ZD} & \textbf{Act.} & \textbf{NLL} 
& \makecell{\textbf{Only}\\\textbf{Dir.}} 
& \makecell{\textbf{Only}\\\textbf{Mag.}} 
& \textbf{HiSA} \\
\midrule
LLaMA-3-8B & 1.00 & 35.84 & 29.67 & 32.51 & 27.94 & 25.86 & 24.59 & 23.33 & \textbf{22.18} \\
Qwen3-8B  & 1.00 & 28.92 & 23.38 & 25.71 & 21.45 & 19.37 & 18.15 & 17.59 & \textbf{16.26} \\
\bottomrule
\end{tabular}
\caption{Ablation on different allocation-sensitivity metrics and HiSA variants under the same 1.00-bit weight budget. We report WikiText2 perplexity (PPL; lower is better). Only Dir. and Only Mag. use only directional and magnitude distortion, respectively, while HiSA uses the full Soft-OR fusion of both distortion modes. Bold denotes the best result.}
\label{tab:app_hisa_metric}
\end{table*}

We further investigate how different sensitivity metrics affect structural budget allocation under the same average weight-bit budget. 
Specifically, we compare five representative allocation metrics and three HiSA variants on \textsc{LLaMA-3-8B} and \textsc{Qwen3-8B}, while keeping the binary factorization procedure, calibration data, and reconstruction settings unchanged. 
The quantization quality is evaluated by WikiText2 perplexity (PPL; lower is better).

\begin{itemize}
    \item \textbf{Uniform} assigns the same structural budget to all layers and submodules, ignoring the non-uniform sensitivity of different components.
    \item \textbf{LIM} (Layer Input Modification) measures layer importance by the cosine drift between the layer input and output representations, where larger input--output changes indicate higher sensitivity.
    \item \textbf{ZD} (Z-score Distribution) estimates sensitivity by the fraction of statistically abnormal weights in each layer, assuming that layers with more outlier weights are harder to binarize.
    \item \textbf{Activation-based scoring} uses the Frobenius norm of layer activations as the allocation criterion, based on the assumption that layers with larger activation energy carry more important information.
    \item \textbf{NLL Increase} measures the validation negative log-likelihood increase caused by quantizing one layer at a time, while keeping the remaining layers at a higher-precision reference.
    \item \textbf{Only Directional} keeps the hierarchical Shapley allocation framework but uses only the directional distortion for module-level budget allocation.
    \item \textbf{Only Magnitude} keeps the hierarchical Shapley allocation framework but uses only the magnitude distortion for module-level budget allocation.
    \item \textbf{HiSA} denotes the full dual-distortion allocation strategy, which fuses directional and magnitude sensitivities with the Soft-OR rule.
\end{itemize}

As shown in Table~\ref{tab:app_hisa_metric}, HiSA consistently achieves the lowest WikiText2 perplexity on both models under the same 1.00-bit weight budget. 
Compared with uniform allocation, HiSA reduces perplexity from 35.84 to 22.18 on \textsc{LLaMA-3-8B} and from 28.92 to 16.26 on \textsc{Qwen3-8B}, indicating that uniformly assigning the same structural capacity to all layers is highly suboptimal for genuine 1-bit PTQ. 
Local sensitivity metrics such as LIM, ZD, and activation-based scoring improve over the uniform baseline in most cases, but their performance remains clearly worse than HiSA because they rely on isolated representation shifts, weight outlier statistics, or activation magnitudes, and therefore cannot fully reflect the end-to-end effect of quantization errors. 
NLL Increase provides a stronger baseline by directly measuring the loss perturbation caused by single-layer quantization, and it achieves the best result among these non-Shapley allocation baselines. 
Nevertheless, HiSA further reduces PPL from 25.86 to 22.18 on \textsc{LLaMA-3-8B} and from 19.37 to 16.26 on \textsc{Qwen3-8B}. 
This improvement shows that evaluating layer sensitivity in isolation is still insufficient, since quantization errors can accumulate, propagate, and be partially compensated across layers.

We further isolate the contribution of the dual-distortion design by comparing full HiSA with two single-distortion variants. 
Only Directional uses the directional Shapley score alone, while Only Magnitude uses the magnitude Shapley score alone. 
Both variants underperform the Soft-OR fusion strategy, showing that the two distortion modes capture complementary failure patterns. 
Specifically, on \textsc{LLaMA-3-8B}, HiSA reduces PPL from 24.59 and 23.33 to 22.18 compared with Only Directional and Only Magnitude, respectively. 
On \textsc{Qwen3-8B}, HiSA similarly reduces PPL from 18.15 and 17.59 to 16.26. 
These results confirm that the strong 1-bit performance of AF1 comes not only from hierarchical Shapley allocation, but also from explicitly preserving modules that are vulnerable in either directional or magnitude distortion.

\subsubsection{Additional Ablation on HiSA Hyperparameters}
\label{app:hisa_more_ablation}

\begin{figure*}[t]
    \centering
    \includegraphics[width=\linewidth]{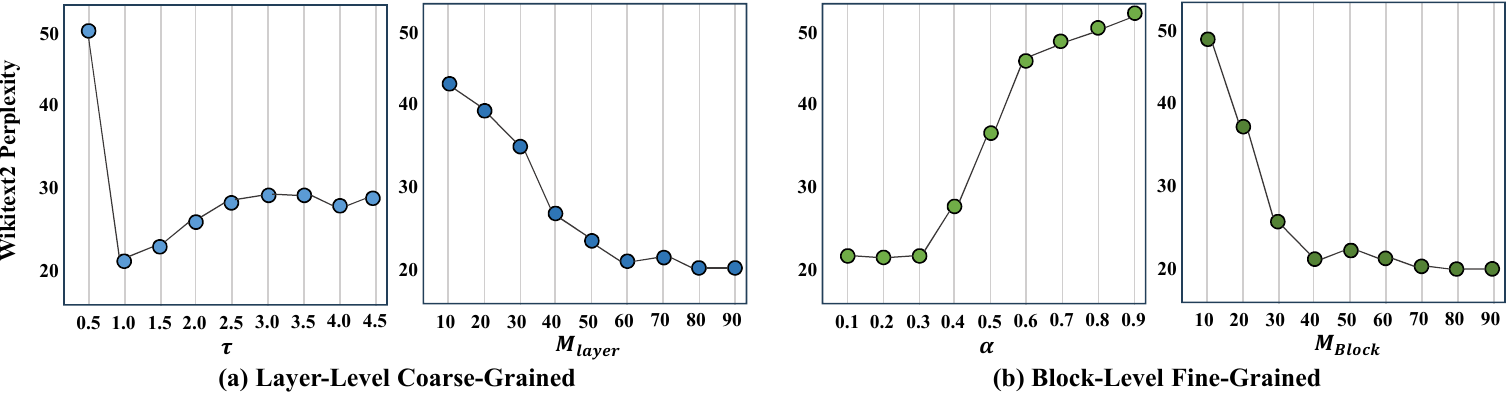}
    \caption{Additional hyperparameter sensitivity of HiSA on \textsc{LLaMA-3-8B}. We report WikiText2 perplexity under different values of the layer-level temperature coefficient $\tau$, the layer-level sampling number $M_{\mathrm{layer}}$, the block-level smoothing exponent $\alpha$, and the block-level sampling number $M_{\mathrm{block}}$.}
    \label{fig:hisa_hyperparam_llama3}
\end{figure*}

\begin{figure*}[t]
    \centering
    \includegraphics[width=\linewidth]{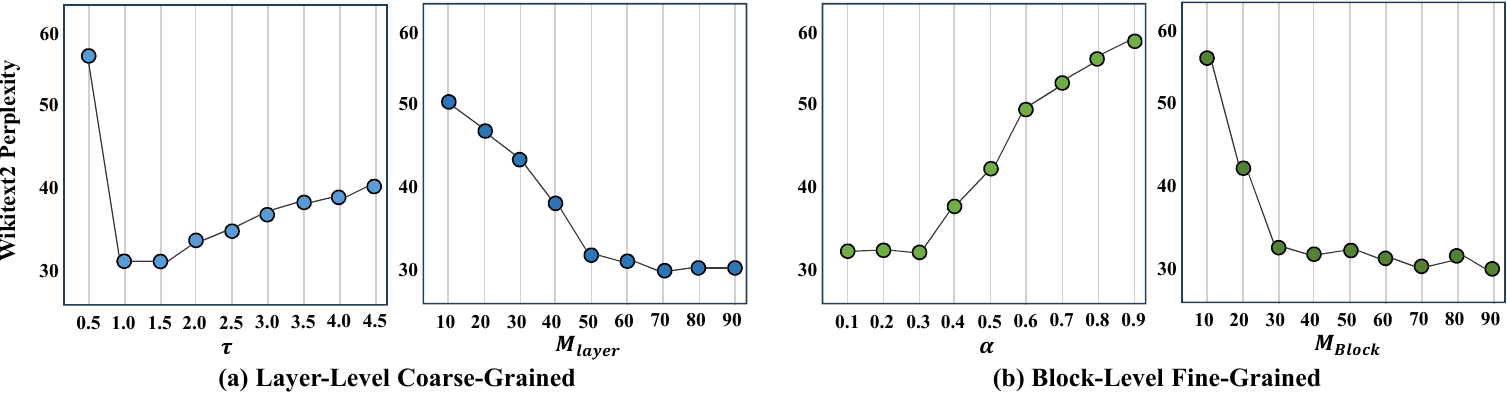}
    \caption{Additional hyperparameter sensitivity of HiSA on \textsc{Gemma-3-4B-it}. We report WikiText2 perplexity under different values of the layer-level temperature coefficient $\tau$, the layer-level sampling number $M_{\mathrm{layer}}$, the block-level smoothing exponent $\alpha$, and the block-level sampling number $M_{\mathrm{block}}$.}
    \label{fig:hisa_hyperparam_gemma3}
\end{figure*}

We provide additional ablation results for HiSA on \textsc{LLaMA-3-8B} and \textsc{Gemma-3-4B-it} to further verify the robustness of the default hyperparameter configuration. 
Together with the Qwen3-8B results in Fig.~\ref{fig:hisa_hyperparam}, these additional curves show that the behavior of HiSA is consistent across different model families and scales. 
We analyze four key hyperparameters: the layer-level temperature coefficient $\tau$, the layer-level sampling number $M_{\mathrm{layer}}$, the block-level smoothing exponent $\alpha$, and the block-level sampling number $M_{\mathrm{block}}$.

For the layer-level allocation, $\tau$ controls how concentrated the structural budget is across layers. 
When $\tau$ is too small, the allocation becomes overly sharp and assigns excessive capacity to a small number of layers, which leaves many other layers under-allocated and leads to a large increase in perplexity. 
When $\tau$ becomes too large, the allocation gradually approaches a uniform distribution and weakens the effect of layer-wise sensitivity estimation. 
As shown in Fig.~\ref{fig:hisa_hyperparam_llama3} and Fig.~\ref{fig:hisa_hyperparam_gemma3}, both models achieve strong performance around $\tau=1.0$, which is consistent with the trend observed on Qwen3-8B. 
This indicates that a moderately concentrated layer-wise allocation is preferable: it preserves the benefit of sensitivity-aware budget assignment while avoiding excessive concentration on only a few layers.

The sampling number $M_{\mathrm{layer}}$ determines the accuracy of the Monte Carlo Shapley estimation at the layer level. 
When only a small number of permutations is used, the estimated layer contribution is noisy, and the resulting allocation is unstable. 
Increasing $M_{\mathrm{layer}}$ consistently reduces perplexity on both \textsc{LLaMA-3-8B} and \textsc{Gemma-3-4B-it}, and the improvement gradually saturates after a moderate number of samples. 
This trend confirms that the layer-level Shapley score benefits from sufficient permutation sampling, but it does not require an excessively large sampling budget to obtain a reliable allocation.

For the block-level allocation, $\alpha$ controls the strength of sensitivity-density redistribution among submodules inside each layer. 
A small-to-moderate value of $\alpha$ leads to stable performance, because it allows sensitive submodules to receive more capacity while keeping the allocation sufficiently smooth. 
In contrast, a large $\alpha$ over-amplifies the difference between submodule sensitivities and causes the budget to be concentrated on a small subset of modules. 
This behavior substantially degrades perplexity on both models, especially when $\alpha$ is larger than $0.4$. 
Therefore, the default setting $\alpha=0.2$ provides a robust balance between sensitivity awareness and allocation smoothness.

The block-level sampling number $M_{\mathrm{block}}$ shows a similar pattern to $M_{\mathrm{layer}}$. 
Using too few samples results in unreliable submodule-level contribution estimates and noticeably worse perplexity. 
As $M_{\mathrm{block}}$ increases, the estimated block-level importance becomes more stable, and the perplexity quickly decreases before reaching a near-saturated region. 
Although using more samples can sometimes bring marginal additional improvement, the gain becomes much smaller after the estimation stabilizes. 
Therefore, we adopt $M_{\mathrm{layer}}/M_{\mathrm{block}}=60/40$ as the default configuration, which provides a good trade-off between quantization quality and allocation overhead.

Overall, these additional ablations support the hyperparameter choices used in the main experiments. 
The trends on \textsc{LLaMA-3-8B}, \textsc{Qwen3-8B}, and \textsc{Gemma-3-4B-it} consistently show that HiSA is not sensitive to a narrow, model-specific setting. 
Instead, it maintains stable performance under a reasonable range of hyperparameters, while clearly degrading only under extreme allocation regimes such as overly concentrated layer budgets or overly aggressive block-level sensitivity amplification. 
These results further demonstrate that the effectiveness of HiSA comes from its hierarchical sensitivity modeling rather than from delicate hyperparameter tuning.

\subsubsection{Ablation on Calibration Data}
\label{app:ablation_calibration_data}

\begin{figure*}[t]
    \centering
    \includegraphics[width=0.92\linewidth]{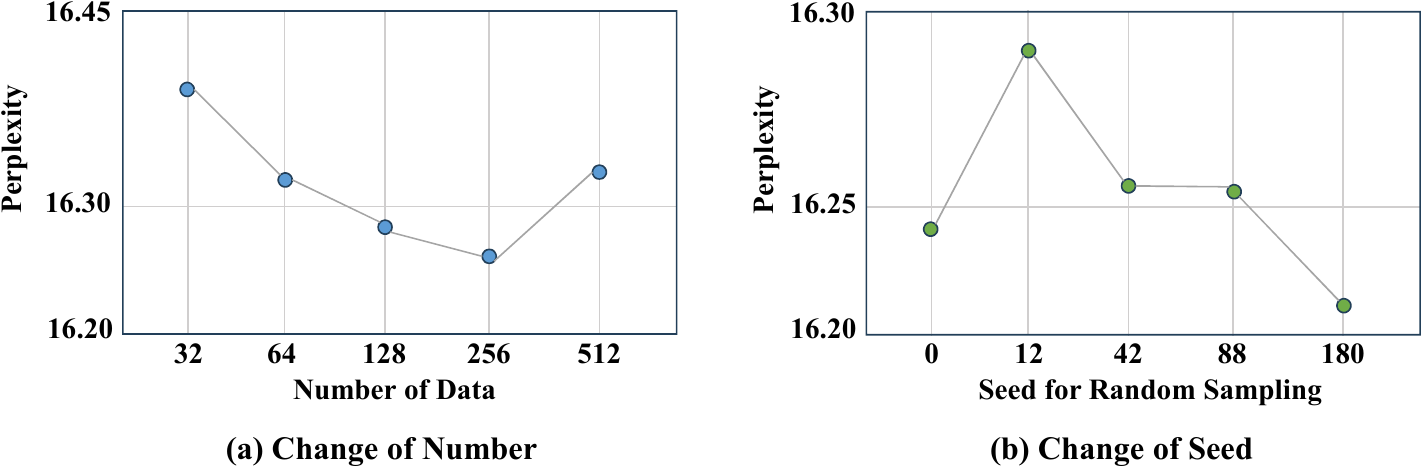}
    \caption{Ablation on calibration data. We use C4 as the calibration corpus and report WikiText2 perplexity after quantization. (a) varying the number of calibration samples. (b) varying the random seed for calibration-data sampling.}
    \label{fig:calibration_ablation}
\end{figure*}

We further investigate the influence of calibration data on AF1. 
Specifically, we use C4 as the calibration corpus and evaluate the quantized model by WikiText2 perplexity, while keeping all quantization hyperparameters and reconstruction settings unchanged. 
We consider two factors: the number of calibration samples and the random seed used for calibration-data sampling. 
For the calibration-size study, we vary the number of C4 samples from 32 to 512. 
For the seed-sensitivity study, we fix the calibration size and change the random seed, which affects both sample selection and token order.

As shown in Fig.~\ref{fig:calibration_ablation}, AF1 exhibits stable performance under different calibration-data configurations. 
When the number of calibration samples increases from 32 to 512, the WikiText2 perplexity remains within a narrow range, with the best result obtained around 256 samples and only a slight degradation when using more samples. 
This indicates that AF1 does not require a large calibration set to obtain reliable reconstruction statistics. 
Similarly, changing the random seed leads to only minor fluctuations in perplexity, and all tested seeds produce comparable results. 
These observations suggest that AF1 is insensitive to both calibration-set size and calibration-sampling randomness. 
Therefore, a small representative subset of C4 is sufficient for stable post-training binarization, which reduces calibration overhead and improves the practicality of AF1 in deployment scenarios.

\subsection{Dialog Examples}
\label{sec:dialog}

To further illustrate the qualitative impact of different binarization schemes, Table~\ref{tab:qwen3_qualitative_examples} presents representative generations from Qwen3-14B-Instruct under full precision and different binary PTQ methods, including BiLLM, ARB-LLM, HBLLM, and our proposed AF1. 
Different from open-ended dialogue prompts, we use short questions with objective answers, covering factual recall, elementary arithmetic, and basic commonsense knowledge. 
This design allows us to directly examine whether the compressed models preserve the key answer content and whether their generations remain stable under extreme weight compression.

In addition to checking the ground-truth answers, we further use GPT-5.5 as an external LLM judge to provide a consistent qualitative assessment of each response. 
Specifically, the judge is asked to evaluate whether a generation contains the correct key answer, whether it remains semantically coherent, and whether it suffers from repetition, symbol collapse, or irrelevant continuation. 
The LLM judge is used only for qualitative annotation rather than as a replacement for standard benchmark evaluation, and its judgments are reported in the assessment column of Table~\ref{tab:qwen3_qualitative_examples}. 
This provides a more systematic way to distinguish factually correct but unstable outputs from truly coherent generations.

As shown in Table~\ref{tab:qwen3_qualitative_examples}, existing binary PTQ methods exhibit different types of generation degradation. 
BiLLM frequently collapses into repeated tokens, symbols, or self-reinforcing phrases, and in some cases produces factually incorrect answers despite the simplicity of the prompts. 
ARB-LLM preserves more factual information than BiLLM on several questions, but it can still produce redundant or unstable continuations, especially when the generation length increases. 
HBLLM achieves relatively stable generations in these examples, but this comes at a substantially larger storage cost, with an effective weight budget exceeding 3 BPW in our setting. 
Therefore, its qualitative advantage should be interpreted together with its much higher bit-width rather than as evidence of genuine 1-bit compression.

In contrast, AF1 targets a genuinely low-bit regime, using only around 1 BPW for weights, while prior binary PTQ baselines typically require approximately 2.5--3.5 BPW due to additional full-precision scaling factors, grouping metadata, or structured compensation terms. 
Despite this much stricter storage budget, AF1 preserves the key answer content across most selected examples and avoids the severe symbolic collapse observed in prior binary PTQ methods. 
For instance, AF1 correctly answers the arithmetic, capital-city, average-speed, red-planet, and triangle-angle questions while maintaining readable sentence structure. 
Although long-form repetitions may still occur in some cases, these qualitative results indicate that AF1 achieves a more favorable trade-off between storage efficiency and generation quality, demonstrating that near-1-BPW LLM compression can retain basic language generation ability under extreme quantization.

\begin{table*}[h!]
\centering
\tablestyle{2.5pt}{1.4}
\begin{tabular}{
|>{\centering\arraybackslash}m{0.19\linewidth}
|>{\centering\arraybackslash}m{0.10\linewidth}
|>{\centering\arraybackslash}m{0.52\linewidth}
|>{\centering\arraybackslash}m{0.15\linewidth}|
}
\hline
\textbf{Prompt} & \textbf{Method} & \textbf{Response} & \textbf{Assessment} \\
\hline

\multirow{5}{=}{\centering 
\textbf{Q1:} I have 3 apples, buy 5 more, and give away 2. How many apples are left?\\
\textbf{Answer:} 6}
& BF16 
& You start with 3 apples, buy 5 more, and give away 2. Final answer: \textcolor{darkgreen}{6 apples}. 
& \textcolor{darkgreen}{Correct} \\
\cdashline{2-4}

& BiLLM 
& \textcolor{darkred}{The ** ** ** ** ** ** ** ** ** ** $\cdots$}
& \textcolor{darkred}{Degenerated} \\
\cdashline{2-4}

& ARB-LLM 
& $3+5=8$, and $8-2=6$. \textcolor{darkgreen}{6 apples} are left.
& \textcolor{darkgreen}{Correct} \\
\cdashline{2-4}

& HBLLM 
& $3+5=8$ apples, then $8-2=6$ apples.
& \textcolor{darkgreen}{Correct} \\
\cdashline{2-4}

& \textbf{AF1} 
& You start with 3 apples, buy 5 more, making 8 apples. Then you give away 2, so you have \textcolor{darkgreen}{6 apples} left.
& \textcolor{darkgreen}{Correct} \\
\hline

\multirow{5}{=}{\centering 
\textbf{Q2:} What are the capitals of Canada, Japan, and Germany?\\
\textbf{Answer:} Ottawa, Tokyo, Berlin}
& BF16 
& Canada: \textcolor{darkgreen}{Ottawa}; Japan: \textcolor{darkgreen}{Tokyo}; Germany: \textcolor{darkgreen}{Berlin}.
& \textcolor{darkgreen}{Correct} \\
\cdashline{2-4}

& BiLLM 
& \textcolor{darkred}{Capital of Canada: Montreal.} \textcolor{darkred}{** ** ** ** $\cdots$}
& \textcolor{darkred}{Wrong + repetitive} \\
\cdashline{2-4}

& ARB-LLM 
& Canada's capital is \textcolor{darkgreen}{Ottawa}, Japan's capital is \textcolor{darkgreen}{Tokyo}, and Germany's capital is \textcolor{darkgreen}{Berlin}.
& \textcolor{darkgreen}{Correct} \\
\cdashline{2-4}

& HBLLM 
& Canada: \textcolor{darkgreen}{Ottawa}; Japan: \textcolor{darkgreen}{Tokyo}; Germany: \textcolor{darkgreen}{Berlin}.
& \textcolor{darkgreen}{Correct} \\
\cdashline{2-4}

& \textbf{AF1} 
& Canada: \textcolor{darkgreen}{Ottawa}; Japan: \textcolor{darkgreen}{Tokyo}; Germany: \textcolor{darkgreen}{Berlin}.
& \textcolor{darkgreen}{Correct} \\
\hline

\multirow{5}{=}{\centering 
\textbf{Q3:} A train travels 120 kilometers in 2 hours. What is its average speed?\\
\textbf{Answer:} 60 km/h}
& BF16 
& Average speed is $120/2=\textcolor{darkgreen}{60}$ km/h.
& \textcolor{darkgreen}{Correct} \\
\cdashline{2-4}

& BiLLM 
& \textcolor{darkgreen}{60 km/h}, followed by \textcolor{darkred}{** ** ** ** ** $\cdots$}
& \textcolor{darkorange}{Correct but repetitive} \\
\cdashline{2-4}

& ARB-LLM 
& The average speed is \textcolor{darkgreen}{60 kilometers per hour}.
& \textcolor{darkgreen}{Correct} \\
\cdashline{2-4}

& HBLLM 
& Average speed is computed as distance divided by time, giving \textcolor{darkgreen}{60 km/h}.
& \textcolor{darkgreen}{Correct} \\
\cdashline{2-4}

& \textbf{AF1} 
& Average speed $=120/2=\textcolor{darkgreen}{60}$ kilometers per hour.
& \textcolor{darkgreen}{Correct} \\
\hline

\multirow{5}{=}{\centering 
\textbf{Q4:} Which planet is known as the Red Planet?\\
\textbf{Answer:} Mars}
& BF16 
& The planet known as the Red Planet is \textcolor{darkgreen}{Mars}.
& \textcolor{darkgreen}{Correct} \\
\cdashline{2-4}

& BiLLM 
& \textcolor{darkred}{The Red Planet is not commonly used in real-world science} \textcolor{darkred}{$\cdots$ repeated}
& \textcolor{darkred}{Wrong + repetitive} \\
\cdashline{2-4}

& ARB-LLM 
& The planet known as the Red Planet is \textcolor{darkgreen}{Mars}, also known as Mars or the Red Planet.
& \textcolor{darkorange}{Correct but redundant} \\
\cdashline{2-4}

& HBLLM 
& The planet known as the Red Planet is \textcolor{darkgreen}{Mars}.
& \textcolor{darkgreen}{Correct} \\
\cdashline{2-4}

& \textbf{AF1} 
& The planet known as the ``Red Planet'' is \textcolor{darkgreen}{Mars}.
& \textcolor{darkgreen}{Correct} \\
\hline

\multirow{5}{=}{\centering 
\textbf{Q5:} What is the sum of the interior angles of a triangle?\\
\textbf{Answer:} 180 degrees}
& BF16 
& The sum of the interior angles of a triangle is \textcolor{darkgreen}{180 degrees}.
& \textcolor{darkgreen}{Correct} \\
\cdashline{2-4}

& BiLLM 
& \textcolor{darkred}{******** ** ** ** ** ** ** ** $\cdots$}
& \textcolor{darkred}{Symbol collapse} \\
\cdashline{2-4}

& ARB-LLM 
& The sum of the interior angles of a triangle is \textcolor{darkgreen}{180 degrees}.
& \textcolor{darkgreen}{Correct} \\
\cdashline{2-4}

& HBLLM 
& The sum of the interior angles of a triangle is \textcolor{darkgreen}{180 degrees}.
& \textcolor{darkgreen}{Correct} \\
\cdashline{2-4}

& \textbf{AF1} 
& The sum of the interior angles of a triangle is \textcolor{darkgreen}{180 degrees}.
& \textcolor{darkgreen}{Correct} \\
\hline

\multirow{5}{=}{\centering 
\textbf{Q6:} What is the chemical formula of water?\\
\textbf{Answer:} H$_2$O}
& BF16 
& The chemical formula of water is \textcolor{darkgreen}{H$_2$O}.
& \textcolor{darkgreen}{Correct} \\
\cdashline{2-4}

& BiLLM 
& The formula is \textcolor{darkgreen}{H$_2$O}, but it repeatedly states that water is composed of two water molecules \textcolor{darkred}{$\cdots$}
& \textcolor{darkred}{Wrong explanation + repetition} \\
\cdashline{2-4}

& ARB-LLM 
& The formula is \textcolor{darkgreen}{H$_2$O}. \textcolor{darkred}{Water is composed of two hydrogen atoms and one oxygen atom $\cdots$ repeated.}
& \textcolor{darkorange}{Correct but repetitive} \\
\cdashline{2-4}

& HBLLM 
& The chemical formula of water is \textcolor{darkgreen}{H$_2$O}.
& \textcolor{darkgreen}{Correct} \\
\cdashline{2-4}

& \textbf{AF1} 
& The formula is \textcolor{darkgreen}{H$_2$O}, but the explanation is repeated afterwards.
& \textcolor{darkorange}{Correct but repetitive} \\
\hline

\end{tabular}
\caption{Qualitative examples on Qwen3-14B-Instruct under full precision and different binary PTQ methods. All prompts have objective answers. Dark green marks correct key information, dark red marks wrong answers or degenerated repetitions, and dark orange indicates partially correct but unstable generations. Repeated continuations are abbreviated for readability.}
\label{tab:qwen3_qualitative_examples}
\end{table*}

\subsection{Use of Large Language Models}
\label{sec:LLMs}
In preparing this manuscript, we employed large language models (LLMs) as writing assistants. Specifically, LLMs were used for grammar correction, wording improvement, and stylistic polishing of the text. In certain cases, LLMs were also leveraged to rephrase or restructure preliminary drafts of specific sections to improve clarity and readability. Importantly, all conceptual contributions, technical methods, experimental designs, and analyses were conceived and developed by the authors without reliance on LLMs. The final responsibility for the accuracy and integrity of the content rests entirely with the authors.

\end{document}